%% file: arxiv.tex
\documentclass{tmlr}

\usepackage[T1]{fontenc}
\usepackage{dsfont}
\usepackage{xspace}
\usepackage{textcomp}
\usepackage{lastpage}
\usepackage{placeins}
\usepackage{needspace}
\usepackage{longtable}
\usepackage{pifont}
\usepackage{tikz}
\usetikzlibrary{arrows.meta,positioning,calc,fit,backgrounds}
\setcitestyle{numbers,square,comma}

\definecolor{citeColor}{RGB}{0,20,115}
\hypersetup{colorlinks,linkcolor={citeColor},citecolor={citeColor},urlcolor={citeColor}}

\definecolor{LightGray}{HTML}{F0F1F2}
\definecolor{AlphaDianaBlue}{HTML}{0A366B}
\definecolor{AlphaDianaLightBlue}{HTML}{DCEAF7}

\definecolor{deepgreen}{RGB}{0,128,0}
\definecolor{myred}{RGB}{200,50,50}
\definecolor{symgray}{RGB}{70,70,70}

\definecolor{py-keyword}{RGB}{0,0,255}
\definecolor{py-string}{RGB}{163,21,21}
\definecolor{py-comment}{RGB}{0,128,0}
\definecolor{py-builtin}{RGB}{0,112,193}
\definecolor{py-bg}{HTML}{FFFFFF}
\definecolor{py-fg}{HTML}{000000}
\definecolor{py-lineno}{HTML}{858585}

\definecolor{action-understanding}{HTML}{E41A1C}
\definecolor{action-planning}{HTML}{377EB8}
\definecolor{action-reasoning}{HTML}{4DAF4A}
\definecolor{action-tooluse}{HTML}{984EA3}
\definecolor{action-verification}{HTML}{FF7F00}
\definecolor{action-finalization}{HTML}{A65628}

\newcommand{\subcap}[2]{\shortstack[c]{#1\\[-2pt](#2)}}

\newcolumntype{Y}{>{\raggedright\arraybackslash}X}

\makeatletter

\renewcommand{\subfigure}[2][]{\subcaptionbox{#1}{#2}}
\newcommand{\modelsubfigure}[2][]{\subcaptionbox{#1}{\makebox[0.16\linewidth][c]{#2}}}
\makeatother

\renewcommand{\emph}[1]{\textit{#1}}
\etocdepthtag.toc{mtchapter}
\etocsettagdepth{mtchapter}{subsection}
\etocsettagdepth{mtappendix}{none}

\newenvironment{nolinenumbers}{}{}

\newcounter{finding}
\NewDocumentCommand{\findingbox}{O{} O{} +m}{%
  \refstepcounter{finding}%
  \begin{tcolorbox}[
    colback=AlphaDianaLightBlue,
    colframe=AlphaDianaBlue,
    boxrule=0.5pt,
    arc=3pt,
    left=3pt, right=3pt, top=2pt, bottom=2pt,
  ]
  \textbf{Finding \thefinding\if\relax\detokenize{#2}\relax\else~(#2)\fi.}%
  \if\relax\detokenize{#1}\relax
  \else
    \label{#1}%
  \fi
  \enspace #3
  \end{tcolorbox}
}

\title{AlphaDiana: A Measurement System \\for Agents in Verifiable Reasoning Tasks}
\renewcommand{\runtitle}{AlphaDiana: A Measurement System for Agents in Verifiable Reasoning Tasks}

\makeatletter
\def\@author{%
\begin{minipage}{0.99\textwidth}
\centering
{\normalsize
Zhanke Zhou\textsuperscript{1$\dagger$},
Zongze Li\textsuperscript{1*},
Weikai Huang\textsuperscript{1*},
Xuan Li\textsuperscript{1*},
Chentao Cao\textsuperscript{1*},
Xiao Feng\textsuperscript{1}\\
Xiangyu Lu\textsuperscript{1},
Jinbo Hu\textsuperscript{1},
Yi Xie\textsuperscript{2},
Nico Pelleriti\textsuperscript{3,4},
Shiyang Liu\textsuperscript{1},
Menghan Lu\textsuperscript{1},
Max Zimmer\textsuperscript{3}\\
Brando Miranda\textsuperscript{5},
Jiangchao Yao\textsuperscript{6},
Bo Liu\textsuperscript{2},
Sanmi Koyejo\textsuperscript{5},
Sebastian Pokutta\textsuperscript{3,4},
Bo Han\textsuperscript{1}
}
\\[1.0em]
{\small\itshape
\textsuperscript{1}TMLR Group, Hong Kong Baptist University;\;\;
\textsuperscript{2}University of Arizona;\\
\textsuperscript{3}Zuse Institute Berlin;\;\;
\textsuperscript{4}Technical University of Berlin;\\
\textsuperscript{5}Stanford University;\;\;
\textsuperscript{6}Shanghai Jiao Tong University;\\[0.25em]
\textsuperscript{$\dagger$}Team lead;\quad
\textsuperscript{*}Equal contribution
}
\end{minipage}%
}
\makeatother
\date{}

\begin{document}
\pagestyle{mynormal}
\maketitle
\thispagestyle{firstpage}

\input{sections/abstract}
\input{sections/main}

\bibliographystyle{plainnat}
\bibliography{bib}

\onecolumn
\appendix
\part{Appendix}
\input{appendix/guide}
\etocdepthtag.toc{mtappendix}
\etocsettagdepth{mtchapter}{none}
\etocsettagdepth{mtappendix}{subsection}
\begingroup
\small
\etocsettocstyle{\small\setlength{\baselineskip}{11pt}\setlength{\parskip}{0pt}}{}
\tableofcontents
\endgroup
\clearpage

\input{appendix/contents}

\end{document}

%% file: sections/abstract.tex
\begin{abstract}

Agents pair reasoning models with harnesses that organize model calls, execute tools, load skill instructions, and retain history in memory. Understanding agent performance requires linking execution conditions to outcomes, behavior, and resource use. We introduce \textbf{AlphaDiana}, a measurement system with modular interfaces for datasets, agents, environments, and scorers. It preserves harness-native control loops while controlling file reuse and memory retention separately. AlphaDiana supports \textit{macro analyses of complete agents} and \textit{micro interventions on agent components} such as tools, skills, and memory under matched conditions. We study three open-source models and three open-source harnesses on five answer-based benchmarks, with two models also tested on two interactive benchmarks. We find that harnesses \textit{do not consistently improve the accuracy of reasoning}: gains on questions unsolved by the model alone can be outweighed by losses on questions it answers correctly without tools. This inconsistency extends to component interventions: neither skill guidance nor retained history reliably improves accuracy. Behavior and output-token use must be assessed with correctness: \textit{low token entropy can accompany wrong answers}, and less output per execution can mean \textit{more output tokens per correct answer} when accuracy falls. AlphaDiana links answer correctness and execution status to traces and resource records for comparing agents and diagnosing failures under shared task and resource policies.\footnote{Project website: \url{https://alphadiana.org/}. Project source code: \url{https://github.com/tmlr-group/AlphaDiana}. We also develop AlphaApollo~\citep{zhou2025alphaapollo}, an open-source reasoning system for problem solving: \url{https://alphaapollo.org/}.}
\end{abstract}

%% file: sections/main.tex
\newcommand{\chordpanel}[3]{\modelsubfigure[\subcap{#1}{\scriptsize #2}]{\includegraphics[width=0.14\linewidth,trim={2pt 2pt 2pt 2pt},clip]{#3}}}

\definecolor{cellGood}{RGB}{215,238,223}
\definecolor{cellGoodL}{RGB}{235,246,239}
\definecolor{cellBadL}{RGB}{247,228,226}
\definecolor{cellBad}{RGB}{239,207,203}
\newcommand{\cellg}[1]{\multicolumn{1}{>{\columncolor{cellGoodL}}c}{#1}}
\newcommand{\cellr}[1]{\multicolumn{1}{>{\columncolor{cellBadL}}c}{#1}}
\newcommand{\cellR}[1]{\multicolumn{1}{>{\columncolor{cellBad}}c}{#1}}

\providecommand{\mainpagebreak}{\FloatBarrier}
\begingroup
\setlength{\intextsep}{2pt}
\setlength{\textfloatsep}{26pt plus 2pt minus 2pt}
\setlength{\floatsep}{26pt plus 2pt minus 2pt}
\captionsetup{skip=7pt}
\tcbset{before skip=6pt, after skip=6pt}
\raggedbottom
\begin{figure}[H]
\captionsetup{hypcap=false}
\centering
\includegraphics[width=\linewidth]{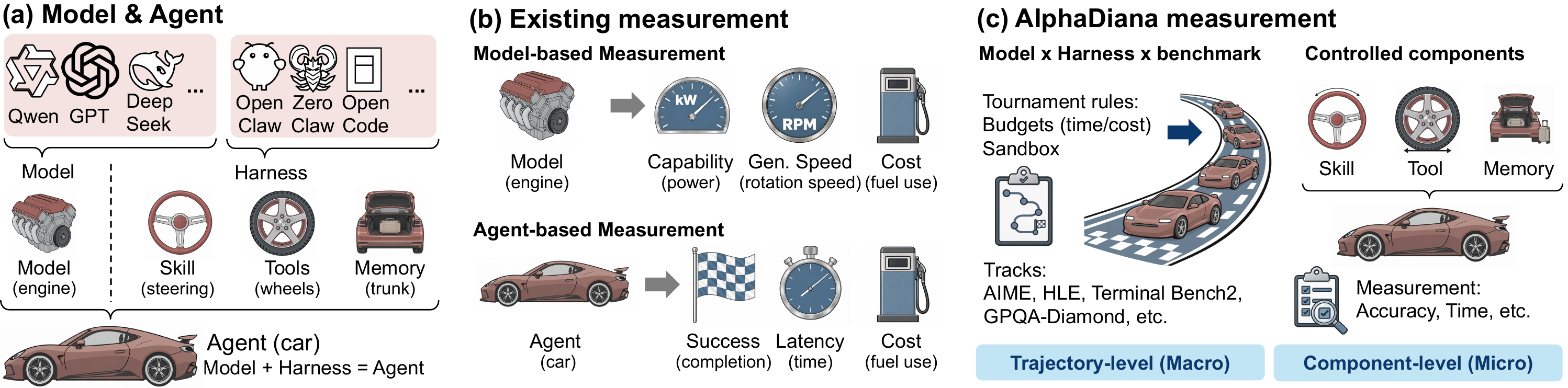}
\captionof{figure}{\textbf{Position of AlphaDiana.} Models and agents are analogous to engines and cars. Model-centered evaluations vary models while fixing the harness. Like a tournament of cars, AlphaDiana compares complete model--harness systems under common rules through trajectory analysis, component interventions, and measurements of task accuracy, output-token use, and wall-clock time.}
\label{fig:alphadiana-motivation}
\end{figure}

\section{Introduction}
\label{sec:intro}

Benchmarks provide verifiable tasks~\citep{rein2024gpqa,jimenez2024swebench}, while evaluation frameworks~\citep{kapoor2025hal,harbor} support agent evaluation. Explaining differences between agents poses three challenges: \textbf{(1)~How should sessions be isolated or reused?} An \emph{attempt} is one execution of an agent on a task. Sessions provide execution environments, while reuse policies control which files and memories later attempts can access. Uncontrolled reuse can expose earlier solutions and make comparisons depend on task order. \textbf{(2)~How can comparisons preserve harness-native behavior?} Adapters need common inputs and outputs while retaining each harness's model calls, tool permissions, and feedback loop. \textbf{(3)~How can outcomes be linked to agent behavior?} Scores must link to the actions, failures, and resource use of the same attempt. Fig.~\ref{fig:alphadiana-motivation} frames these comparisons as a tournament of complete agents.


\begin{figure}[t!]
\captionsetup{hypcap=false}
\centering
\includegraphics[width=\linewidth]{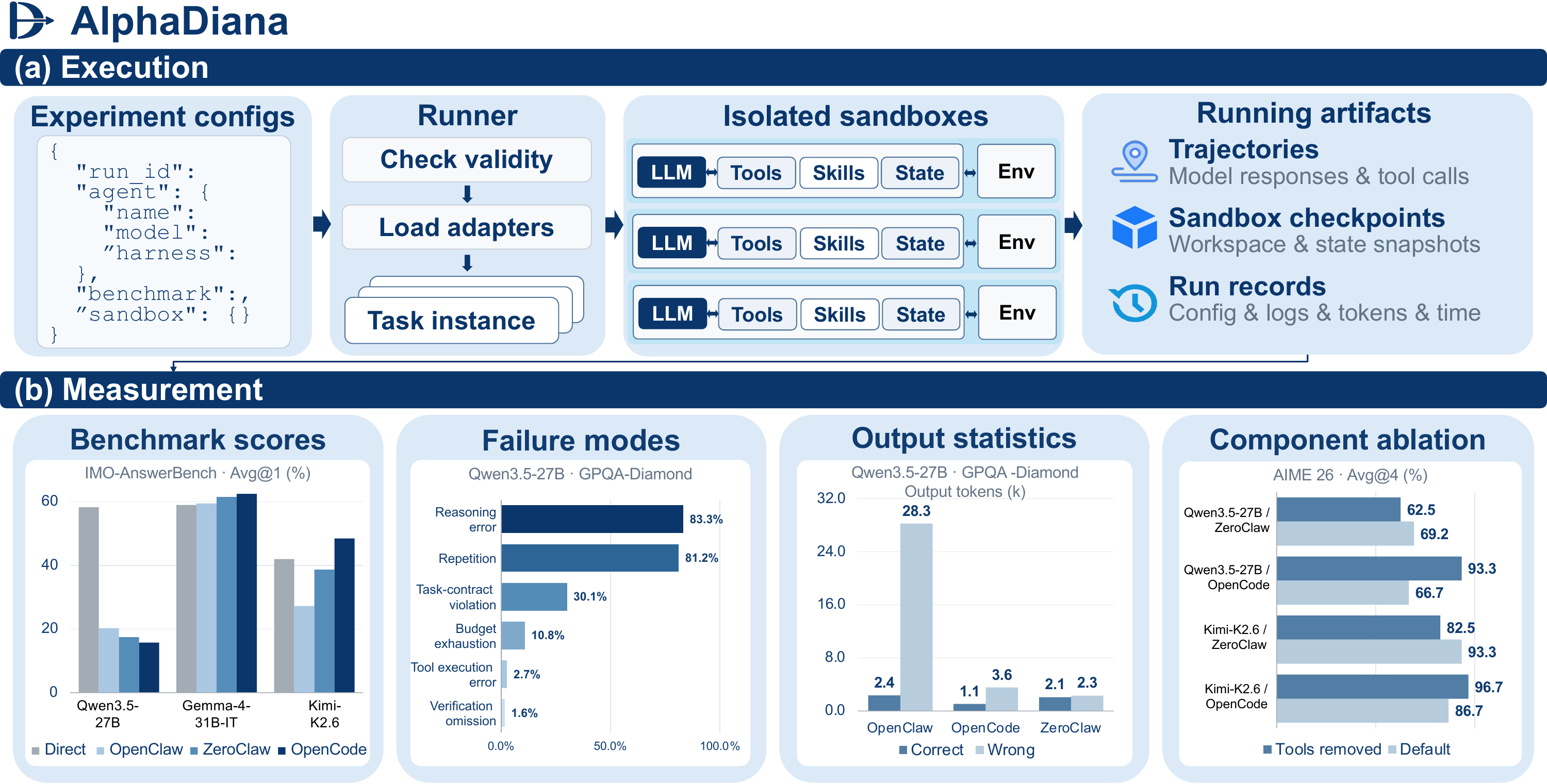}
\captionof{figure}{\textbf{Overview of AlphaDiana.}
AlphaDiana \emph{executes} configured agents in controlled environments and saves their artifacts.
\emph{Measurement} compares performance, reasoning behavior, and resource use, and tests tools, skills, and memory under matched conditions.}
\label{fig:alphadiana-overview}
\end{figure}

We introduce AlphaDiana, a measurement system linking \textit{controlled comparisons} and \textit{component interventions} to outcomes, traces, and resource use. Researchers compare each question's result against a reference and inspect the actions and output tokens behind gains and losses. Matched interventions vary tools, skills, or memory while fixing the model, tasks, scoring, and decoding.

AlphaDiana's \textit{modular interfaces} let users configure, combine, and extend datasets, agents, environments, and scorers without changing the runner. Three contributions address these challenges:
\begin{itemize}[leftmargin=0.5em, itemsep=1pt, topsep=0pt, parsep=0pt]
    \vspace{-2pt}
    \item \textbf{Session isolation and controlled reuse.} Sessions provide isolated or shared execution environments, with file resets and memory retention controlled separately. Attempts updating shared memory run sequentially without automatic retries, because failures can change what later attempts inherit.
    \item \textbf{Adapters preserving harness-native execution.} Adapters standardize task inputs and outputs while preserving each harness's model calls, permissions, feedback, and context updates.
    \item \textbf{Results linked to execution evidence.} Each recorded task execution links its score and execution status to traces and resource use. Reports include failed and unscored results and use the latest record for each requested evaluation, so retries do not count as additional evaluations of that task.
\end{itemize}

We pair the models Qwen, Gemma, and Kimi with the harnesses OpenClaw, ZeroClaw, and OpenCode on five answer-based benchmarks that score final answers. For each model, \emph{Direct} provides the reference: one response per attempt without tools. We also test Qwen and Gemma on two interactive benchmarks, using task verifiers after file edits or commands. We find:
\begin{itemize}[leftmargin=0.5em, itemsep=1pt, topsep=0pt, parsep=0pt]
    \vspace{-2pt}
    \item \textbf{Harnesses can solve extra questions yet lower overall accuracy.} For Qwen on GPQA-Diamond, each harness solves 8--13 questions unsolved by \emph{Direct} but fails on 18--36 that \emph{Direct} solves. Across the four systems, 175/198 questions are solved at least once, versus 159/198 for \emph{Direct} alone.
    \item \textbf{Reasoning patterns alone do not show whether an answer is correct.} Similar tool-use shares, consecutive checks, and low entropy occur with correct and incorrect answers. In annotated incorrect HLE-Verifiable attempts combining the three models, repetition labels occur in 36.8\% for OpenCode and 39.4\% for ZeroClaw, but budget-exhaustion labels occur in 13.0\% versus 0\%.
    \item \textbf{Tools, skills, and memory do not consistently improve answers.} With four attempts per AIME~26 problem, removing tool descriptions and related instructions from Kimi/OpenCode raises correct attempts from 104 to 116 but reduces problems solved at least once from 30 to 29. Skill loading improves accuracy in only 2 of 16 comparisons. For Qwen on AIME~26, memory collected on AIME~25 and kept fixed helps OpenCode but hurts OpenClaw and ZeroClaw relative to defaults.
    \item \textbf{Small token savings can come with far fewer correct answers.} On common Gemma/HLE-Verifiable tasks, OpenCode uses 0.6\% fewer output tokens than \emph{Direct}, but gives 140 correct answers versus 164. We divide output from all attempts by the number of correct answers. This average increases by 16.5\%. OpenClaw's highest-token 10\% of attempts use 62.2\% of output without a correct answer. Traces can reveal repetition, tool failures, or incorrect reasoning.
\end{itemize}

\mainpagebreak
\setlength{\intextsep}{12pt}
\begingroup
\setlength{\intextsep}{2pt}
\begin{figure}[t!]
\captionsetup{hypcap=false}
\centering
\includegraphics[width=\linewidth]{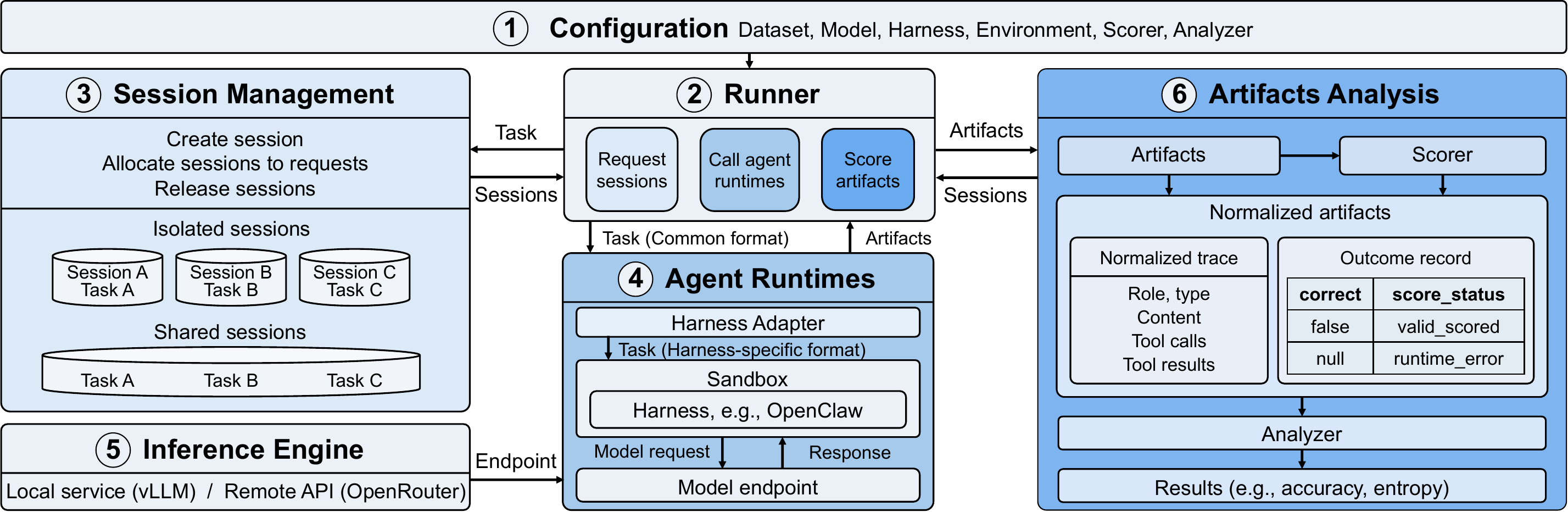}
\captionof{figure}{\textbf{System design of AlphaDiana.} Configuration selects tasks, agents, environments, and scorers. The runner requests sessions and passes tasks to harness adapters. Native runtimes execute permitted tools in the assigned environments and query local or remote model endpoints configured by the inference engine. The runner saves available execution artifacts and scorer judgments before releasing sessions. Normalized traces and outcome records link each attempt's evidence so the analyzer can compare correctness, behavior, and resource use across tasks and inspect failures.}
\label{fig:alphadiana-architecture}
\end{figure}

\section{AlphaDiana: System Design and Interfaces}
\label{sec:alphadiana_main}

The six modules in Fig.~\ref{fig:alphadiana-architecture} connect configuration, harness-native execution, and evidence analysis:

\textbf{(1) Configuration.}
Configurations select tasks, model--harness pairs, environments, scorers, analyses, resource limits, evaluations per task, parallelism, and memory scope. Matched comparisons fix the model, tasks, scoring, and decoding while varying harnesses or components. The adapters in Appendix~\ref{app:alphadiana} add compatible tasks and harnesses while reusing the runner and result format.

\Needspace{4\baselineskip}
\textbf{(2) Runner.}
The runner schedules each requested task evaluation as a \emph{sample}, coordinates its execution and scoring, and saves its results. A retry retains the sample identifier, linking available outputs, errors, and termination details to the same evaluation even when execution or scoring fails.

\textbf{(3) Session Management.}
The session manager creates and releases sessions, allocating them to attempts under two policies. Adapters control file resets and memory retention separately:
\begin{itemize}[leftmargin=0.5em, itemsep=1pt, topsep=0pt, parsep=0pt]
    \item \emph{Isolated sessions.} Where supported, a pool assigns separate sessions to concurrent attempts and resets them before reuse. Strict isolation stops the run if sandbox setup or pool preparation fails.
    \item \emph{Shared sessions.} Attempts may reuse a session within one task or across tasks. Within-task sharing replaces the session and reinitializes the agent before the next task. Shared attempts run sequentially. Attempts updating shared memory are not retried automatically: a failed attempt may already have changed the shared memory that a retry or any later attempt would inherit.
\end{itemize}

\textbf{(4) Agent Runtimes.}
Adapters translate tasks into harness-specific inputs and collect available answers, trajectories, termination reasons, and artifacts. Harnesses control model calls, action parsing, tool permissions, feedback, context updates, skills, and memory. Accepted actions invoke tools, while rejections follow harness-native feedback or termination paths. Runtimes may be local, containerized, or remote. In the matched Qwen/HLE-Verifiable example in Appendix~\ref{app:matched-harness-trace}, OpenClaw recovers from a computation error, ZeroClaw proceeds after a shell-policy rejection, and OpenCode incorporates shell results into its context. They reach the same answer through different feedback paths. Adapters preserve these paths while translating task inputs and normalizing outputs.

\textbf{(5) Inference Engine.}
Harnesses query local services such as vLLM or remote providers such as OpenRouter independently of the environments executing tools. Where supported, AlphaDiana records model requests, responses, and token probabilities for offline analysis of generation.

\textbf{(6) Artifacts Analysis.}
A \emph{normalized trace} stores available actions and feedback in a common format linked to source records. An \emph{outcome record} stores scorer judgments separately from execution status, distinguishing scored wrong answers, execution failures, and missing scores. Reports include failed and unscored results and select the latest record per task and sample, so retries do not add samples. The analyzer compares output-token use and the proportions and ordering of reasoning and tool-use steps across outcomes. Entropy uses renormalized top-$K$ next-token probabilities to measure concentration among the recorded candidate tokens, rather than the full vocabulary.

\mainpagebreak
\begin{figure}[t!]
\captionsetup{hypcap=false}
\centering
\includegraphics[width=\linewidth]{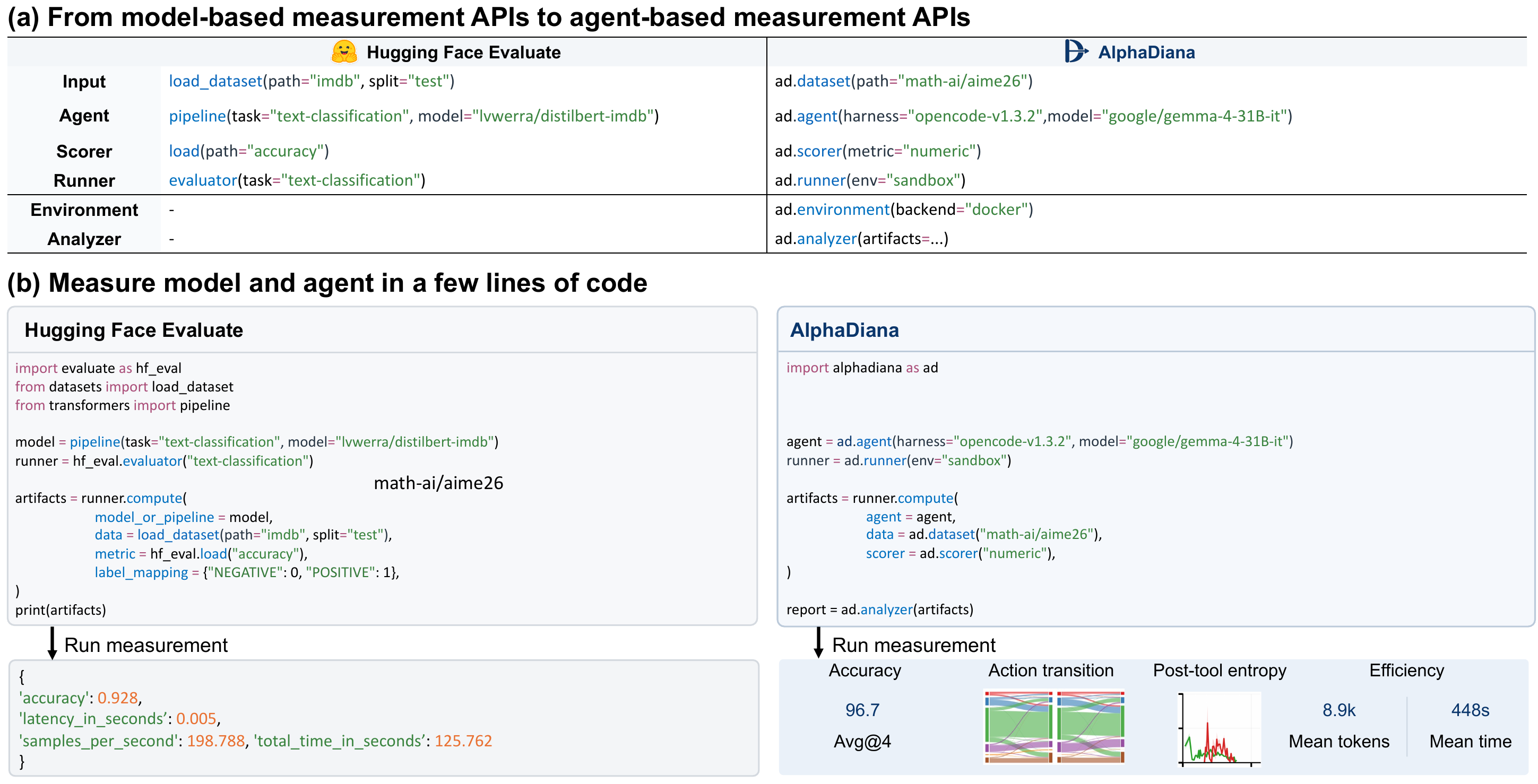}
\caption{\textbf{Functional interface of AlphaDiana.} (a)~Evaluation abstractions in Hugging Face Evaluate and AlphaDiana. (b)~Saved artifacts support performance reporting and offline process analysis.}
\label{fig:alphadiana-workflows}
\end{figure}
\textbf{Unified measurement interface.}
AlphaDiana provides composable execution and measurement interfaces. Its dataset, agent, environment, scorer, runner, and analyzer APIs support macro comparisons and matched micro interventions on tools, skills, and memory. Fig.~\ref{fig:alphadiana-workflows} shows one composition: execution and scoring produce artifacts for analyzing accuracy, behavior, and resource use. New summaries reuse these records. Component changes require matched executions.

\endgroup
\begingroup
\setlength{\intextsep}{2pt}
\setlength{\textfloatsep}{26pt plus 2pt minus 2pt}
\setlength{\floatsep}{26pt plus 2pt minus 2pt}
\captionsetup{skip=7pt}
\tcbset{before skip=7pt, after skip=7pt}
\setcounter{topnumber}{4}
\renewcommand{\topfraction}{0.95}
\renewcommand{\textfraction}{0.05}
\section{Macro View: Open Agents on Verifiable Reasoning Tasks}
\label{sec:macro}\label{sec:macro_process}

We compare which problems complete agents solve, then examine error labels, functional steps, token entropy, and resource use to identify gains and losses hidden by aggregate accuracy.

\looseness=-1
\textbf{Benchmark setup.} We pair \href{https://huggingface.co/Qwen/Qwen3.5-27B}{Qwen3.5-27B}, \href{https://huggingface.co/google/gemma-4-31B-it}{Gemma-4-31B-IT}, and \href{https://huggingface.co/moonshotai/Kimi-K2.6}{Kimi-K2.6} with \href{https://github.com/openclaw/openclaw/releases/tag/v2026.3.7}{OpenClaw}, \href{https://github.com/zeroclaw-labs/zeroclaw/releases/tag/v0.6.9}{ZeroClaw}, and \href{https://github.com/anomalyco/opencode/releases/tag/v1.3.2}{OpenCode}. The answer-based benchmarks score final answers against references: \href{https://maa.org/news/2025-26-aime-thresholds-are-now-available/}{AIME~26}, \href{https://imobench.github.io/}{IMO-AnswerBench}, GPQA-Diamond~\citep{rein2024gpqa}, HLE-Verifiable~\citep{phan2025hle}, and MMMU-Pro~\citep{yue2024mmmupro}. Their baseline is \emph{Direct}. For interactive tasks, we compare Qwen and Gemma with \href{https://github.com/SWE-agent/SWE-agent/releases/tag/v1.1.0}{SWE-Agent} on SWE Mini, a 50-task SWE-bench Verified subset~\citep{jimenez2024swebench}, and with terminus-2 on Terminal-Bench~2~\citep{merrill2026terminalbench}. We use each harness's default configuration, without adding skill libraries or instructions to use memory. We report single-attempt accuracy (Avg@1), except on AIME~26, where Avg@4 averages correctness over four attempts per problem and Pass@4 gives the fraction of problems solved at least once. Interactive accuracy is the fraction of attempts with verified task completion, counting execution or evaluation errors as failures. A shared output limit per model call does not equalize total token use: harnesses make different numbers of calls. Appendices~\ref{app:iclr-setup} and~\ref{app:wall-time-sensitivity} detail decoding and budgets.

\textbf{Agent performance.} In Fig.~\ref{fig:macro-main-results}, all three harnesses lower Qwen's scores on all five answer-based benchmarks, whereas ZeroClaw improves Gemma on all five. Effects also reverse across tasks for the same model: OpenClaw helps Kimi on HLE-Verifiable but hurts it on GPQA-Diamond. No tested harness exceeds the task-native reference on either interactive benchmark. Yet lower scores conceal additional solutions. On Qwen/GPQA-Diamond, each harness solves 8--13 questions missed by \emph{Direct} while losing 18--36 that \emph{Direct} solves (Tab.~\ref{tab:gpqa-accounting}). Together, the four systems solve 175/198 questions, versus \emph{Direct}'s 159/198. Using these complementary solutions requires selecting an agent for each question and testing selection accuracy and output-token use.

\findingbox[find:macro-compatibility][Fig.~\ref{fig:macro-main-results}]{\textbf{Harnesses can add new solutions while reducing overall accuracy.}\label{find:macro-task}\label{find:macro-harness}}

\textbf{Error patterns.} Fig.~\ref{fig:failure-taxonomy} summarizes model-assigned error labels for 2,857 incorrect attempts across three models, four systems, and four benchmarks. Labels overlap. Unscored attempts are excluded. The \emph{repetition rate} is the fraction of these attempts labeled with repetitive cycles (RC-4), counting each once. On HLE-Verifiable, OpenCode and ZeroClaw have similar rates (36.8\% and 39.4\%), yet OpenCode's rate is lower within every model. Different model proportions narrow the gap.

Repetition does not measure resource consumption: budget-exhaustion labels appear in 13.0\% of OpenCode's annotated incorrect HLE attempts and none of ZeroClaw's. Linking the two requires content, resource, and termination evidence. Appendix~\ref{app:error-incidence} defines the labels and procedure. In the study in Appendix~\ref{app:error-human-alignment}, the error-labeling baseline averages 88.0\% agreement with two human annotators across 17 error codes on 42 screened trajectories, versus 86.8\% between the two raters.

\findingbox[find:failure-mode][Fig.~\ref{fig:failure-taxonomy}]{\textbf{Similar overall repetition rates can hide differences within every model.}}

\begin{figure}[t!]
\centering
\includegraphics[width=\linewidth,trim={0pt 0bp 0pt 2.24bp},clip]{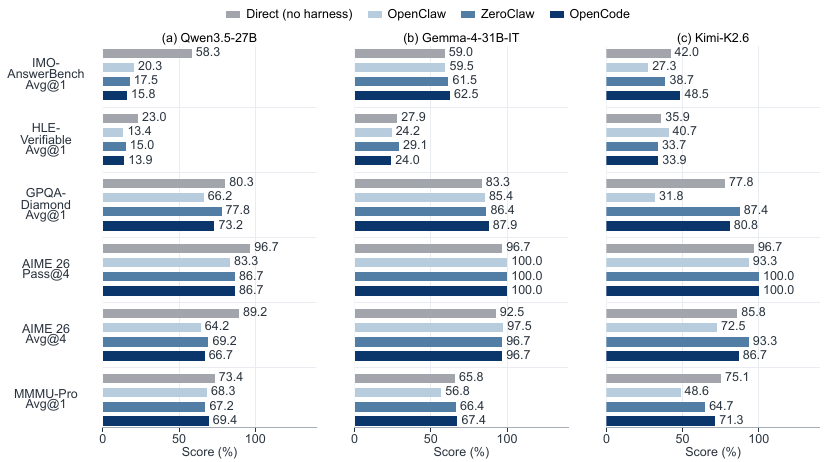}\par
\vspace{2pt}
\includegraphics[width=\linewidth]{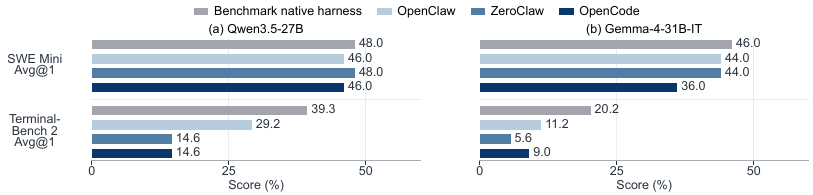}
\caption{\textbf{Harness effects vary across models and tasks} (Finding~\ref{find:macro-compatibility}). References are \emph{Direct} for answer-based tasks (top), and SWE-Agent and terminus-2 for the two interactive benchmarks (bottom).}
\label{fig:macro-main-results}\label{tab:macro-main-results}
\label{fig:macro-interactive-results}\label{tab:macro-interactive-results}
\end{figure}

\textbf{Functional-step patterns.} Fig.~\ref{fig:count-abs} compares 2,751 scored Gemma and Kimi trajectories on IMO-AnswerBench, GPQA-Diamond, and HLE-Verifiable. GPT-5.6-sol labels each \emph{functional step}, a saved text segment or logged call (Appendix~\ref{app:action-labeling}). On the human calibration sample, agreement with majority judgments is 90.9\% (Appendix~\ref{app:action-human-alignment}). Panel~(a) reports step shares within each model--benchmark--system--outcome group, giving longer trajectories more weight. For incorrect Gemma/ZeroClaw trajectories on HLE-Verifiable, the 61.0\% Tool Use share comes entirely from one of 75 trajectories. Computing each trajectory's share first and then averaging gives only 1.3\%. Thus one long trajectory accounts for the large pooled share. Kimi/OpenCode's HLE tool-use shares are similar across outcomes. In Fig.~\ref{fig:count-abs}(b), OpenClaw has consecutive Verification steps in both outcomes. Distinguishing successful error correction from repeated failed checks requires inspecting what each check found and how the model used the result.

\findingbox[find:transition-control][Fig.~\ref{fig:count-abs}]{\textbf{High tool-use shares can come from a single long trajectory.}\label{find:failure-loop}}

\begin{figure}[t!]
\centering
\includegraphics[width=0.95\linewidth]{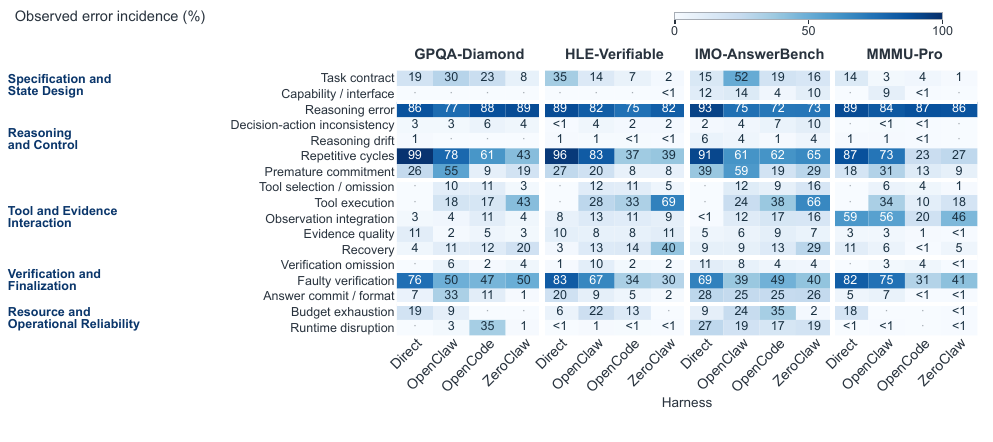}
\vspace{-10pt}
\caption{\textbf{Error patterns across models and harnesses} (Finding~\ref{find:failure-mode}). Columns pool three models; cells give error-label percentages among incorrect attempts (2,857 overall). Labels overlap; dot: zero, $<1$: positive rate below 1\%. Full definitions and boundaries can be found in Appendix~\ref{app:error-incidence}.}
\label{fig:failure-taxonomy}
\end{figure}

\begin{figure}[t!]
\centering
\includegraphics[width=\linewidth]{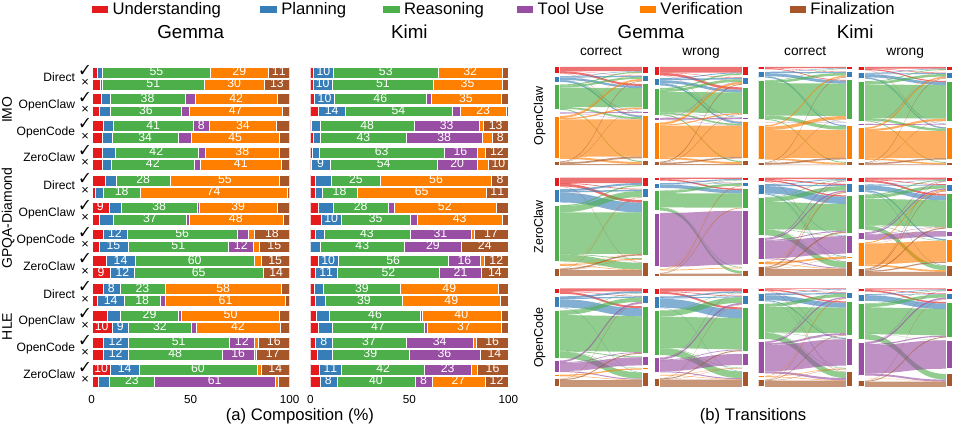}
\caption{\textbf{Functional-step patterns} (Finding~\ref{find:transition-control}). (a)~Step shares for correct ($\checkmark$) and incorrect ($\times$) trajectories. (b)~HLE-Verifiable adjacent-label transitions, normalized per panel. Label definitions: Tab.~\ref{tab:functional-step-definitions}. IMO/HLE denote IMO-AnswerBench/HLE-Verifiable.}
\label{fig:count-abs}\label{fig:action_transition}
\end{figure}

\begin{figure}[t!]
\centering
\includegraphics[width=\linewidth,trim={0pt 1.39bp 0pt 2.17bp},clip]{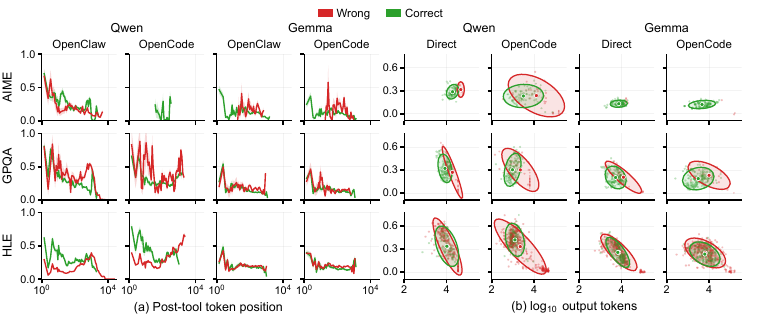}
\caption{\textbf{Token entropy at two scales} (Finding~\ref{find:post-tool-entropy-dynamics}). (a)~Mean entropy by post-tool token position on a logarithmic x-axis, with standard-error bands. (b)~Mean entropy versus $\log_{10}$ output tokens for scored trajectories. Points represent individual trajectories. Large markers are centroids and ellipses are two-standard-deviation covariance contours.}
\label{fig:post-tool_entropy_analyze}\label{fig:entropy_length_scatter}
\vspace{+4pt}
\end{figure}

\textbf{Token-entropy patterns.} We compute entropy from recorded top-$K$ next-token probabilities, renormalized to sum to one, for Qwen and Gemma on AIME~26, GPQA-Diamond, and HLE-Verifiable. Lower values mean probability is concentrated on fewer recorded candidates. In Fig.~\ref{fig:post-tool_entropy_analyze}(a), entropy often falls after tool feedback in trajectories with either correct or incorrect final answers. Fig.~\ref{fig:entropy_length_scatter}(b) also shows long incorrect \emph{Direct} and OpenCode trajectories with low mean entropy. Both post-tool and whole-trajectory views show concentrated predictions accompanying incorrect solutions.

\findingbox[find:post-tool-entropy-dynamics][Fig.~\ref{fig:post-tool_entropy_analyze}]{\textbf{Falling entropy after tool feedback can precede a wrong answer.}\label{find:entropy-length-collapse}}

\textbf{Token use and wall-clock time.} Fig.~\ref{fig:macro-efficiency} compares mean output tokens and recorded wall-clock time per attempt, including unsuccessful attempts. On Qwen/AIME~26, OpenCode uses 24.6\% fewer output tokens than \emph{Direct} but takes 49.6\% longer. Gemma/ZeroClaw on GPQA-Diamond shows the reverse: more output accompanies shorter wall-clock time. Token records cover all 120 and 198 attempts per system in these comparisons, respectively. Wall-clock time is the elapsed duration of an attempt, including model serving, tool execution, and scheduling. The opposite rankings show why token and time budgets need separate evaluation. Appendix~\ref{app:cost-measurement} details coverage and timing.

\findingbox[find:macro-efficiency][Fig.~\ref{fig:macro-efficiency}]{\textbf{Token and time measurements can rank the same systems differently.}}

\begin{figure}[t!]
\centering
\includegraphics[width=\linewidth,trim={0pt 0.83bp 0pt 1.39bp},clip]{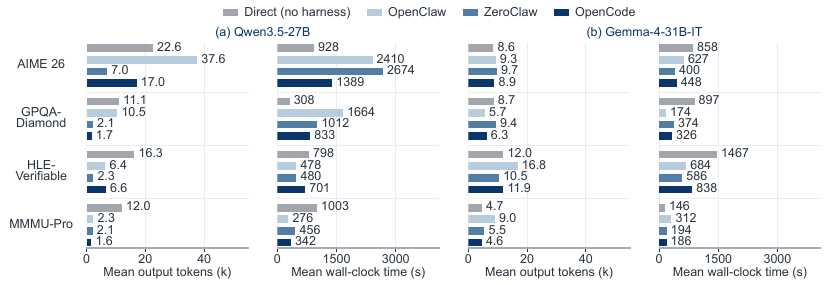}
\caption{\textbf{Output tokens and wall-clock time per attempt} (Finding~\ref{find:macro-efficiency}). (a)~Qwen3.5-27B. (b)~Gemma-4-31B-IT. Bars show mean recorded output tokens (thousands, excluding input) and wall-clock time (seconds), including unsuccessful attempts. Record coverage can differ across systems and metrics.}
\label{fig:macro-efficiency}
\end{figure}
\endgroup

\begingroup
\setlength{\intextsep}{2pt}
\setlength{\textfloatsep}{2pt}
\setlength{\floatsep}{2pt}
\setcounter{topnumber}{3}
\renewcommand{\topfraction}{0.95}
\renewcommand{\textfraction}{0.05}
\captionsetup{skip=7pt}
\tcbset{before skip=3pt, after skip=3pt}

\section{Micro View: How Harness Capabilities Contribute to Reasoning}
\label{sec:micro}

We vary tools, skills, and memory, then measure output-token use of complete agents. Tool and skill interventions use Qwen and Kimi with ZeroClaw and OpenCode on GPQA-Diamond and AIME~26. Each comparison retains the model, tasks, scoring, decoding, budgets, and harness-native loop of \emph{Default Harness}. Appendix~\ref{app:component-intervention-details} specifies the configuration changes and component contents.

\textbf{Tool removal.} \emph{Tool Registry Removed} suppresses model-visible tool descriptions and related instructions while preserving the harness loop. Fig.~\ref{fig:micro-tool} shows average accuracy falling for ZeroClaw in three of four model--task pairs but rising for OpenCode in three. A simpler action interface can thus improve accuracy. This comparison jointly changes prompting and tool availability.

On AIME~26, Kimi/OpenCode solves 29 problems on all four attempts after tool removal: 116 correct attempts, versus 104 for the default. Yet the default solves all 30 problems at least once. Higher Avg@4 thus reflects repeated success on fewer distinct problems, despite a lower Pass@4.

\findingbox[find:micro-tool-sign][Fig.~\ref{fig:micro-tool}]{\textbf{Tool removal can raise average accuracy while solving fewer problems.}\label{find:micro-tool-consistency}}

\looseness=-1
\textbf{Skill loading.} \emph{Skill Loaded} makes a math-specific or general procedural library available through the harness's native interface. Fig.~\ref{fig:micro-skill} shows average-accuracy gains in only two of sixteen comparisons, both with Qwen. Kimi/ZeroClaw opens Math's top-level \texttt{SKILL.md} on every GPQA-Diamond attempt while accuracy falls. Fig.~\ref{fig:skill-use-rate} reports reading rates under the corresponding loading instructions: OpenCode uses matched instructions across libraries, whereas ZeroClaw requires reading Math but makes General optional.

\findingbox[find:micro-skill-preference][Figs.~\ref{fig:micro-skill} and~\ref{fig:skill-use-rate}]{\textbf{Skills can be read on every attempt while accuracy falls.}\label{find:micro-skill-model}}

\begin{figure}[t!]
\centering
\includegraphics[width=\linewidth]{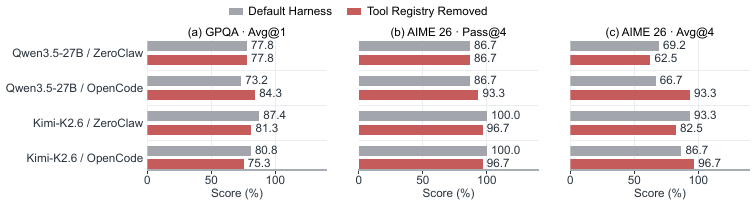}
\caption{\textbf{Tool removal affects harnesses differently} (Finding~\ref{find:micro-tool-sign}). Model-visible tool descriptions and related instructions are removed from \emph{Default Harness}. GPQA denotes GPQA-Diamond.}
\label{fig:micro-tool}\label{tab:micro-tool}
\end{figure}

\begin{figure}[t!]
\centering
\includegraphics[width=\linewidth]{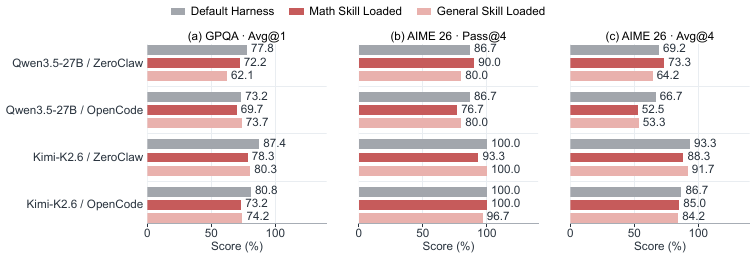}
\caption{\textbf{Skill loading rarely improves average accuracy} (Finding~\ref{find:micro-skill-preference}). Math and General denote the math-specific and general procedural skill libraries used here. GPQA denotes GPQA-Diamond.}
\label{fig:micro-skill}\label{tab:micro-skill}
\end{figure}

\begin{figure}[t!]
\centering
\includegraphics[width=\linewidth,trim={0pt 1.3bp 0pt 1.02bp},clip]{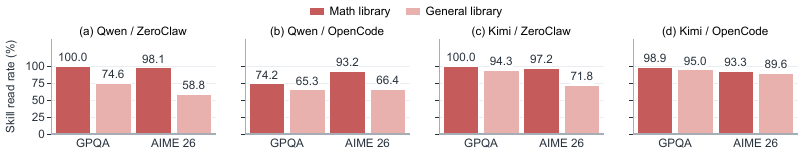}
\caption{\textbf{Skill-reading rates} (Finding~\ref{find:micro-skill-preference}): fraction of trajectories opening the top-level \texttt{SKILL.md}.}
\label{fig:skill-use-rate}
\end{figure}

\begin{figure}[t!]
\centering
\includegraphics[width=\linewidth,trim={0pt 3.02bp 0pt 3.3bp},clip]{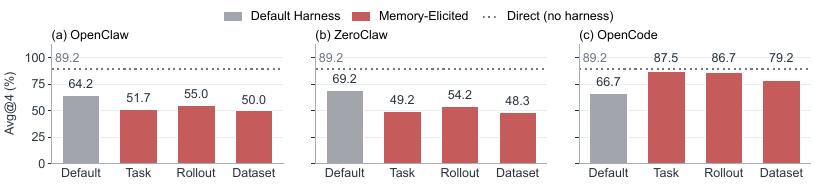}
\caption{\textbf{Memory effects on Qwen/AIME~26} (Finding~\ref{find:micro-memory}). Task/Rollout/Dataset denote the three persistence scopes. The dotted line marks \emph{Direct}'s 89.2\% Avg@4, without a harness or tools.}
\label{fig:micro-memory-all}\label{fig:micro-memory}
\end{figure}

\begin{figure}[t!]
\centering
\includegraphics[width=\linewidth]{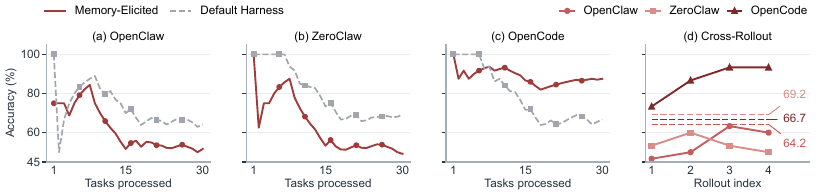}
\caption{\textbf{Accuracy as memory accumulates} (Finding~\ref{find:memory-accumulation}). (a--c)~Cross-Task running accuracy from separate sequential runs, with default scores that may differ from Fig.~\ref{fig:micro-memory}. (d)~Cross-Rollout accuracy.}
\label{fig:memory}
\vspace{29pt}
\end{figure}

\textbf{Memory persistence.} We evaluate Qwen on AIME~26 with fixed model weights. \emph{Cross-Rollout} keeps memory across four attempts on the same task, then clears it. \emph{Cross-Task} keeps memory across tasks within each of four separate runs. \emph{Cross-Dataset} uses frozen AIME~25 memory without updates. The default clears memory between attempts. We vary retention and add memory-use instructions. Raw histories and task files reset between attempts. Appendix~\ref{app:memory-mechanisms} details storage and retrieval.

Fig.~\ref{fig:micro-memory} shows memory helping OpenCode and hurting OpenClaw and ZeroClaw in all three scopes. These directions persist with frozen AIME~25 memory, so gains and losses do not require updates on the target tasks. Yet all nine configurations remain below the model-only \emph{Direct} baseline.

OpenCode uses prior solutions in prompts, whereas OpenClaw and ZeroClaw retrieve summaries. Retrieval, prompt length, and control loops vary, so memory format is not isolated. Fig.~\ref{fig:memory}(d) shows accuracy plateauing or falling across later attempts, which can reuse prior solutions and are not independent. Cross-Task accuracy averages all tasks seen, so it also varies with task difficulty.

\findingbox[find:micro-memory][Figs.~\ref{fig:micro-memory} and~\ref{fig:memory}]{\textbf{Frozen memory can help one harness and hurt another.}\label{find:memory-accumulation}}

\textbf{Output-token use.} We compare complete Gemma agents on 198 GPQA-Diamond and 587 HLE-Verifiable tasks with token records for all systems, following Appendix~\ref{app:cost-measurement}. We divide total output tokens by the number of attempts (\emph{output tokens per attempt}) or correct answers (\emph{output tokens per correct answer}). The total includes output from incorrect answers, unscored attempts, and execution failures, and excludes input tokens. Output tokens per correct answer equal output tokens per attempt divided by accuracy as a fraction.

In Fig.~\ref{fig:micro-solution-cost}, OpenCode saves 0.6\% of output tokens per HLE-Verifiable attempt relative to \emph{Direct}, yet its output tokens per correct answer are 16.5\% higher. Nearly unchanged output is spread across fewer correct attempts. When accuracy falls, token savings reduce output tokens per correct answer only if their relative reduction exceeds the relative accuracy loss. A ranking based only on output length reverses this output-token comparison.

\vspace{15pt}
\findingbox[find:micro-solved-cost][Fig.~\ref{fig:micro-solution-cost}]{\textbf{Fewer output tokens per attempt can mean more per correct answer.}}

\begin{figure}[t!]
\centering
\includegraphics[width=\linewidth]{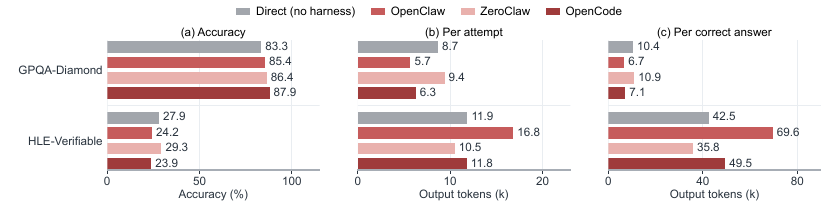}
\caption{\textbf{Output-token use} (Finding~\ref{find:micro-solved-cost}). Each total includes output from unsuccessful attempts.}
\label{fig:micro-solution-cost}
\vspace{36pt}
\end{figure}

\begin{figure}[t!]
\centering
\includegraphics[width=\linewidth]{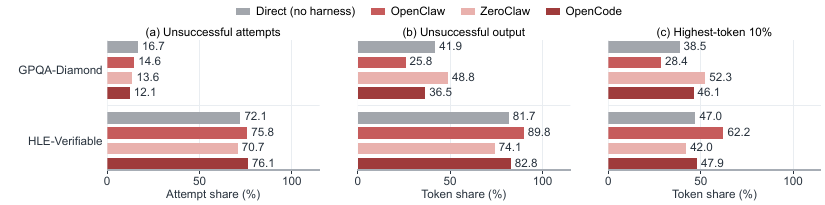}
\caption{\textbf{Where output tokens are spent} (Finding~\ref{find:micro-cost-concentration}). Same tasks as Fig.~\ref{fig:micro-solution-cost}. Panels show the unsuccessful-attempt fraction, its output share, and the highest-token 10\%'s output share.}
\label{fig:micro-token-allocation}
\vspace{36pt}
\end{figure}

\textbf{Output concentration.} Figs.~\ref{fig:micro-solution-cost} and~\ref{fig:micro-token-allocation} show why averaging only successful attempts omits output spent on failures. On HLE-Verifiable, OpenClaw's correct attempts average fewer tokens than \emph{Direct}'s, yet its output tokens per correct answer are 63.9\% higher because unsuccessful attempts consume 89.8\% of its output. Its highest-token 10\% consume 62.2\% without a correct answer. These totals do not reveal when failure becomes predictable. Stopping rules must be tested for both accuracy and token use on matched tasks. The highest-token and unsuccessful groups overlap, so their output shares cannot be added.

\vspace{15pt}
\findingbox[find:micro-cost-concentration][Fig.~\ref{fig:micro-token-allocation}]{\textbf{10\% of attempts can consume most output without a correct answer.}}

\FloatBarrier
\section{Conclusion}
\label{sec:conclusion}

AlphaDiana links harness-native execution and controlled session reuse to outcomes, traces, and resource records. Harnesses can solve questions missed by the model alone yet reduce accuracy. Fewer output tokens per attempt can mean more per correct answer. Component changes should be judged by answer correctness and total output per correct answer. Matched interventions and saved traces let researchers test these changes and inspect individual failures.

\clearpage
\endgroup
\endgroup

\setcounter{topnumber}{3}

%% file: appendix/guide.tex
\noindent\textbf{Appendix guide.}
Appendix~\ref{app:alphadiana} documents system interfaces; Appendix~\ref{app:iclr-main-details} specifies experimental settings and measurement coverage. Appendix~\ref{app:macro} combines analysis methods and supplementary macro results in main-text order: performance, errors, functional steps, token entropy, and resource use. Appendix~\ref{app:component-intervention-details} gives the micro-study protocols. Related work appears in Appendix~\ref{app:related work}.
\medskip

%% file: appendix/contents.tex
\input{appendix/3-alphadiana}

\clearpage
\FloatBarrier
\input{appendix/0-iclr-main-details}

\FloatBarrier
\Needspace{10\baselineskip}
\input{appendix/4-further-macro-exp}

\FloatBarrier
\Needspace{10\baselineskip}
\input{appendix/5-component-intervention-details}

\FloatBarrier
\Needspace{10\baselineskip}
\input{appendix/2-related-work}

%% file: appendix/3-alphadiana.tex
\section{System Design and Implementation Details}
\label{app:alphadiana}

We present \textbf{AlphaDiana}, a modular measurement system for \emph{system-level evaluation} of LLM-based reasoning systems. Performance depends on the base model and its runtime stack: harnesses, tool interfaces, execution environments, scoring protocols, and run configurations. AlphaDiana separates evaluation into independent modules with explicit interfaces, enabling controlled comparison across direct models and harnessed-agent systems.

\noindent\textbf{Relation to the main-text architecture.}
Sec.~\ref{sec:alphadiana_main} describes six runtime responsibilities; this appendix documents the reusable interfaces that implement them. Configuration selects the benchmark, agent, environment, scorer, and run policy. The runner coordinates task execution and result storage; environment sessions support isolation and controlled reuse; agent adapters preserve native harness execution and access the configured inference endpoint. Outcome scorers and offline trajectory analyzers consume the resulting records. The interfaces below complement that runtime view. Experimental settings are specified in Appendix~\ref{app:iclr-main-details}, analysis procedures in Appendix~\ref{app:macro}, and intervention policies in Appendix~\ref{app:component-intervention-details}.

\subsection{Overview}
\label{sec:alphadiana_overview}

\paragraph{Experiment tuple.}
An experiment executed by AlphaDiana is formalized as
\[
\mathcal{R} = \mathcal{U}(\mathcal{B}, f_\theta, \mathcal{H}, \mathcal{E}, \mathcal{S}, \mathcal{C}),
\]
where $\mathcal{R}$ denotes the set of result records of the experiment, $\mathcal{U}$ is the runner that orchestrates the experiment, $\mathcal{B}$ is the benchmark, $f_\theta$ is the base model, $\mathcal{H}$ is the harness, $\mathcal{E}$ is an optional execution environment, $\mathcal{S}$ is the scorer, and $\mathcal{C}$ is the run configuration. The agent system is denoted by
\[
\mathcal{A}_{\theta,\mathcal{H}} = \mathcal{H} \odot f_\theta,
\]
where $\odot$ indicates that the harness wraps the base model by providing prompts, tools, state management, control logic, and answer extraction. The direct-model baseline is the special case in which $\mathcal{H}$ reduces to minimal task formatting without an external tool loop.

\paragraph{Modular decomposition.}
AlphaDiana separates evaluation into four user-specified components (benchmark, agent, environment, scorer) and one orchestration layer (runner). Specifically, the benchmark poses tasks, the agent solves them, the environment executes tool-call requests when tool use is required, the scorer judges the resulting prediction, and the runner coordinates the full protocol. After execution, trajectory-analysis utilities derive process measurements from the saved result records. The runner executes the pipeline
\[
\texttt{load task}
\;\longrightarrow\;
\texttt{solve}
\;\longrightarrow\;
\texttt{score}
\;\longrightarrow\;
\texttt{save results}.
\]
This separation is central to the system: benchmark-specific logic is not embedded in the harness, scorer-specific normalization is not delegated to the agent, and environment-specific execution details are isolated behind a sandbox interface. As a result, the same benchmark and scorer can be reused across multiple model-harness systems, making observed differences more attributable to the evaluated system rather than to hidden evaluation-side implementation choices.

\begin{table}[t]
\centering
\small
\setlength{\tabcolsep}{2pt}
\renewcommand{\arraystretch}{1.08}
\begin{tabular}{p{0.18\linewidth}p{0.78\linewidth}}
\toprule
\textbf{Notation} & \textbf{Meaning} \\
\midrule
$\mathcal{B}$ & Benchmark, represented as a finite collection of task instances. \\
$(x,y)$ & Task input $x$ and reference object $y$. \\
$f_\theta$ & Base language model with parameters $\theta$. \\
$\mathcal{H}$ & Harness that prompts, controls, and operationalizes $f_\theta$. \\
$\mathcal{A}_{\theta,\mathcal{H}}$ & Agent system obtained by pairing $f_\theta$ with $\mathcal{H}$. \\
$\mathcal{E}$ & Optional environment backend or sandbox for tool execution. \\
$\mathcal{S}$ & Post-execution evaluation and analysis, comprising outcome scoring and trajectory-level process analysis. \\
$\mathcal{C}$ & Run configuration specifying budgets, policies, and logging options. \\
$\mathcal{U}$ & Runner that orchestrates the evaluation. \\
$\mathcal{R}$ & Result records of the experiment. \\
$p_t,o_t,k_t,\tilde a_t,$\newline$a_t,q_t,r_t,m_t$ & Model-facing context, model output, native output kind, action candidate, admissible action, parse/validation status, environment or harness feedback, and runtime state at turn $t$. \\
$\hat{y}$ & Final predicted answer. \\
$\tau$ & Complete interaction trajectory. \\
$s$ & Task-level score or verdict. \\
\bottomrule
\end{tabular}
\caption{Core notation used in the AlphaDiana evaluation pipeline.}
\label{tab:alphadiana_notation}
\vspace{-20pt}
\end{table}

\paragraph{Interaction loop.}
For a task instance $(x,y) \in \mathcal{B}$, the harness constructs the initial model context
\[
p_1 = \mathrm{Prompt}_{\mathcal{H}}(x),
\]
and initializes any runtime state
\[
m_1 = \mathrm{InitState}_{\mathcal{H},\mathcal{E}}(x,\mathcal{C}).
\]
At turn $t$, the model produces
\[
o_t = f_\theta(p_t).
\]
The harness first classifies $o_t$ under its native output-precedence rules. If it is classified as a terminal answer, the harness extracts the model-predicted answer $\hat{y}$ and terminates.

Otherwise, the harness attempts to parse an action candidate
\[
(\tilde a_t,q_t) = \mathrm{TryParseAction}_{\mathcal{H}}(o_t).
\]
Here $\tilde a_t$ is the candidate (or $\varnothing$ when none is parseable), while $q_t$ retains the native parse or validation status.
Only if parsing succeeds and the requested operation is exposed and permitted does the harness accept $a_t=\tilde a_t$, dispatch it, and receive environment feedback
\[
r_t = \mathcal{E}(a_t).
\]
If parsing or validation fails, $a_t=\varnothing$ and the native runtime instead returns invalid-action feedback or records a terminal error without dispatching to $\mathcal{E}$. Whenever feedback is returned, the harness updates the runtime state and next model-facing context:
\[
m_{t+1}
=
\mathrm{StateUpdate}_{\mathcal{H}}(m_t,o_t,r_t),
\qquad
p_{t+1}
=
\mathrm{PromptUpdate}_{\mathcal{H}}(p_t,o_t,r_t,m_{t+1}).
\]
Thus, one admissible-action interaction step can be summarized as
\[
p_t
\xrightarrow[]{f_\theta}
o_t
\xrightarrow[]{\mathcal{H}}
a_t
\xrightarrow[]{\mathcal{E}}
r_t
\xrightarrow[]{\mathrm{PromptUpdate}_{\mathcal{H}}}
p_{t+1}.
\]
\paragraph{Trajectory and scoring.}
The full trajectory is
\[
\tau =
\bigl(
(p_1,o_1,a_1,r_1),
\ldots,
(p_T,o_T,a_T,r_T),
\hat{y}
\bigr),
\]
where $T$ is the number of turns before termination and $a_t=\varnothing$ on terminal or non-action turns. Termination occurs when the model emits a final answer, a budget in $\mathcal{C}$ is exhausted, or the harness or environment reaches a terminal error state. After termination, the scorer evaluates the prediction independently of the generation process:
\[
s = \mathcal{S}(x,y,\hat{y},\tau).
\]
The runner then stores the prediction, score, trajectory, and metadata for later analysis.

\paragraph{Full procedure.}
Algorithm~\ref{alg:alphadiana_pipeline} summarizes the full evaluation procedure.

\begin{algorithm}[H]
\footnotesize
\caption{The Evaluation Pipeline of AlphaDiana}
\label{alg:alphadiana_pipeline}
\begin{algorithmic}[1]
\REQUIRE Benchmark $\mathcal{B}$, base model $f_\theta$, harness $\mathcal{H}$, environment $\mathcal{E}$, scorer $\mathcal{S}$, configuration $\mathcal{C}$
\ENSURE Result records $\mathcal{R}$ containing predictions, scores, trajectories, and metadata
\STATE Instantiate the agent system $\mathcal{A}_{\theta,\mathcal{H}} \leftarrow \mathcal{H} \odot f_\theta$
\STATE Initialize result store $\mathcal{R} \leftarrow \emptyset$
\FOR{each task instance $(x,y) \sim \mathcal{B}$}
    \STATE Construct initial prompt $p_1 \leftarrow \mathrm{Prompt}_{\mathcal{H}}(x)$
    \STATE Initialize harness state $m_1 \leftarrow \mathrm{InitState}_{\mathcal{H},\mathcal{E}}(x,\mathcal{C})$
    \STATE Initialize trajectory $\tau \leftarrow \emptyset$, turn index $t \leftarrow 1$, final prediction $\hat{y} \leftarrow \varnothing$
    \WHILE{not terminated under $\mathcal{C}$}
        \STATE Generate model output $o_t \leftarrow f_\theta(p_t)$
        \STATE Classify output kind $k_t \leftarrow \mathrm{OutputKind}_{\mathcal{H}}(o_t)$
        \IF{$k_t$ is a terminal answer}
            \STATE Extract prediction $\hat{y} \leftarrow \mathrm{ExtractAnswer}_{\mathcal{H}}(o_t)$
            \STATE Append terminal record $(p_t,o_t,\varnothing,\varnothing)$ to $\tau$
            \STATE \textbf{break}
        \ENDIF
        \STATE Attempt action parsing $(\tilde a_t,q_t) \leftarrow \mathrm{TryParseAction}_{\mathcal{H}}(o_t)$
        \IF{$q_t$ is parsed and $\tilde a_t$ invokes an exposed, policy-permitted operation}
            \STATE Accept action $a_t \leftarrow \tilde a_t$
            \STATE Execute action and obtain feedback $r_t \leftarrow \mathcal{E}(a_t)$
        \ELSE
            \STATE Reject candidate $a_t \leftarrow \varnothing$
            \STATE Set native error record $r_t \leftarrow \mathrm{HandleInvalidAction}_{\mathcal{H}}(o_t,\tilde a_t,q_t)$
        \ENDIF
        \STATE Append turn record $(p_t,o_t,a_t,r_t)$ to $\tau$
        \IF{$r_t$ marks a terminal error}
            \STATE Record the native finish reason
            \STATE \textbf{break}
        \ENDIF
        \STATE Update harness state $m_{t+1} \leftarrow \mathrm{StateUpdate}_{\mathcal{H}}(m_t,o_t,r_t)$
        \STATE Update prompt/context $p_{t+1} \leftarrow \mathrm{PromptUpdate}_{\mathcal{H}}(p_t,o_t,r_t,m_{t+1})$
        \STATE Increment turn index $t \leftarrow t+1$
    \ENDWHILE
    \IF{$\hat{y} = \varnothing$}
        \STATE Extract fallback prediction $\hat{y} \leftarrow \mathrm{FallbackExtract}_{\mathcal{H}}(\tau)$
    \ENDIF
    \STATE Score prediction $s \leftarrow \mathcal{S}(x,y,\hat{y},\tau)$
    \STATE Collect metadata $z \leftarrow \mathrm{Metadata}(\mathcal{B},f_\theta,\mathcal{H},\mathcal{E},\mathcal{S},\mathcal{C},\tau)$
    \STATE Save result $\mathcal{R} \leftarrow \mathcal{R} \cup \{(x,y,\hat{y},s,\tau,z)\}$
\ENDFOR
\STATE \RETURN $\mathcal{R}$
\end{algorithmic}
\end{algorithm}

\subsection{Dataset APIs}
\label{sec:alphadiana_benchmark}

\noindent\textbf{Role.}
The dataset module loads a dataset and materializes it into standardized task instances. Formally, a benchmark is a finite collection
\[
\mathcal{B} = \{(x_i,y_i)\}_{i=1}^{N},
\]
where $x_i$ is the task input and $y_i$ is the reference object used for scoring. The reference object may be a string, numeric answer, symbolic expression, unit-test specification, verifier target, rubric, set of acceptable answers, or any other task-specific supervision object. AlphaDiana treats $y_i$ abstractly: the benchmark provides the reference, while the scorer determines how it should be interpreted.

\noindent\textbf{Task record.}
In implementation, each task record contains a unique task identifier, input $x$, reference field $y$, metadata, and optional attachments. Metadata may include difficulty, category, source, split, or expected output type. Attachments may include files, images, repositories, PDFs, or other artifacts required by multimodal or environment-based tasks. These fields may be utilized by the runner, agent, environment, or scorer, but they do not alter the core agent-facing interface.

\noindent\textbf{Module boundary.}
The benchmark specifies \emph{what} problem is to be solved, not \emph{how} it should be solved. It therefore does not contain harness-specific prompts, tool policies, memory rules, recovery strategies, or scoring logic. Prompt construction belongs to the harness, execution belongs to the environment, and correctness checking belongs to the scorer. This separation prevents benchmark definitions from implicitly favoring a particular harness design.

\noindent\textbf{Interface.}
The dataset API uses the abstract \texttt{Benchmark} interface, which exposes one required method:
\[
\texttt{load\_tasks(config)} \rightarrow \texttt{list[BenchmarkTask]}.
\]
This method converts a raw dataset, file collection, dataset API, or generated task set into the standardized task sequence utilized by the runner. The configuration may specify split names, subset sizes, seeds, task filters, attachment paths, or benchmark-specific preprocessing options. Adding a new benchmark requires implementing this method and registering the benchmark under a configuration-visible name; no agent, environment, scorer, or runner code needs to change.

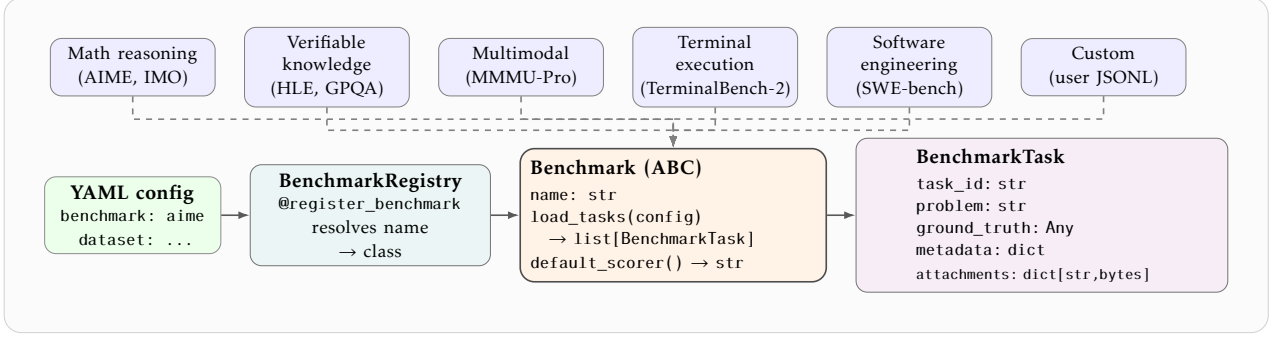
\begin{figure}[t]
\centering
\resizebox{\linewidth}{!}{%
\input{draw-figures/benchmark-module-inline}%
}
\caption{\textbf{Dataset APIs.} Concrete benchmark implementations (left) subclass the abstract \texttt{Benchmark} interface, which exposes \texttt{load\_tasks()} to convert a raw dataset into a sequence of standardized \texttt{BenchmarkTask} tuples (right). The registry resolves a benchmark name from the YAML configuration to the corresponding implementation at runtime.}
\label{fig:benchmark-module}
\end{figure}

\begin{nolinenumbers}
\begin{lstlisting}[
    language=Python,
    label=lst:benchmark,
    caption={Dataset API implemented by \texttt{Benchmark} (\texttt{base.py}, verbatim) and a minimal custom implementation.
             The \texttt{@register\_benchmark} decorator registers the class under a string key
             that matches the \texttt{benchmark:} field in the YAML configuration.},
    basicstyle=\ttfamily\footnotesize\color{py-fg},
    backgroundcolor=\color{py-bg},
    keywordstyle=\color{py-keyword}\bfseries,
    keywordstyle={[2]\color{py-builtin}},
    stringstyle=\color{py-string},
    commentstyle=\color{py-comment}\itshape,
    morekeywords={[2]field,dataclass,abstractmethod,Any,dict,list,str,bytes,True,False,None,self,default_factory,register_benchmark},
    frame=single, framerule=0.4pt,
    rulecolor=\color{black!25},
    numbers=none,
    xleftmargin=1em, framexleftmargin=0.8em,
    breaklines=true, showstringspaces=false
]
# --- base.py (verbatim) -----------------------------------
@dataclass
class BenchmarkTask:
    task_id: str
    problem: str
    ground_truth: Any
    metadata: dict = field(default_factory=dict)
    attachments: dict[str, bytes] = field(default_factory=dict)

class Benchmark(ABC):
    name: str = ""

    @abstractmethod
    def load_tasks(self, config: dict) -> list[BenchmarkTask]: ...

    def default_scorer(self) -> str:
        return "exact_match"

# --- custom implementation (minimal example) --------------
@register_benchmark("my_benchmark")
class MyBenchmark(Benchmark):
    name = "my_benchmark"

    def load_tasks(self, config: dict) -> list[BenchmarkTask]:
        raw = load_my_dataset(config["dataset"])
        return [
            BenchmarkTask(
                task_id=item["id"],
                problem=item["question"],
                ground_truth=item["answer"],
            )
            for item in raw
        ]
\end{lstlisting}
\end{nolinenumbers}

\subsection{Agent Module}
\label{sec:alphadiana_agent}

\noindent\textbf{Role.}
The agent module provides a uniform interface over heterogeneous reasoning systems. It hides implementation differences among direct model calls, lightweight scaffolds, and full agent runtimes behind a common \texttt{solve} interface. Given a task instance, the agent returns an \texttt{AgentResponse} containing the final prediction, interaction trajectory, termination reason, and execution metadata.

\noindent\textbf{Model-harness system.}
The evaluated system is the model-harness pair $\mathcal{A}_{\theta,\mathcal{H}}$. The harness is not merely a prompt template: it determines the initial context, admissible actions, state initialization, parsing rules, error handling, tool exposure, and update logic. It also mediates the distinction between information placed in the model context and information maintained in the external runtime. For example, a prompt may describe available tools, while the actual tool registry, workspace state, and execution constraints are managed outside the model.

\noindent\textbf{Direct and harnessed modes.}
AlphaDiana supports both direct and harnessed modes through the same external interface.
In direct mode, the system performs a single model call with minimal task formatting, extracts an answer, and returns a degenerate one-turn trajectory. In harnessed mode, the system executes the multi-turn model-harness-environment loop defined in Sec.~\ref{sec:alphadiana_overview}. The harness may expose tools, manage persistent state, recover from invalid actions, incorporate environment feedback, and decide how observations enter later model contexts.

\noindent\textbf{AgentResponse.}
The \texttt{AgentResponse} object records both the final prediction and the provenance of the interaction that produced it. At minimum, it contains the predicted answer $\hat{y}$, trajectory $\tau$, and finish reason. It may
also include token usage, wall-clock time, number of turns, number of tool calls, raw model responses, exposed tools, error states, workspace artifacts, and environment snapshots. These fields make it possible to analyze not only whether the final answer was correct, but also how it was produced.

\noindent\textbf{Inference endpoint and recording.}
The native harness queries the configured local or remote model service independently of the execution backend. Request and response recording, including token probabilities when available, supplies the evidence for the offline analysis in Appendix~\ref{app:macro}. Missing probability or usage fields limit the corresponding measurements but do not exclude attempts from outcome accounting.

\noindent\textbf{Module boundary.}
The agent is separated from both the benchmark and the scorer. The benchmark provides the task input and reference object, but does not prescribe the agent's reasoning strategy. The scorer evaluates the prediction after generation, but does not participate in prompting, tool use, or recovery. This boundary is necessary for controlled cross-harness comparison: agents may differ in how they solve tasks, but they are evaluated against the same task and scoring interface.

\noindent\textbf{Interface.}
Adding a new agent requires implementing:
\[
\texttt{setup(config)}
\]
for initializing the model client, harness configuration, prompt templates, tool registry, and runtime state; and
\[
\texttt{solve(task, environment)} \rightarrow \texttt{AgentResponse}
\]
for executing a task and returning the prediction, trajectory, and metadata. Once registered, the agent can be selected directly from the YAML experiment specification.

\begin{nolinenumbers}
\begin{lstlisting}[
    language=Python,
    label=lst:agent,
    caption={Agent interface (\texttt{base.py}, verbatim) and a minimal custom implementation.},
    basicstyle=\ttfamily\footnotesize\color{py-fg},
    backgroundcolor=\color{py-bg},
    keywordstyle=\color{py-keyword}\bfseries,
    keywordstyle={[2]\color{py-builtin}},
    stringstyle=\color{py-string},
    commentstyle=\color{py-comment}\itshape,
    morekeywords={[2]abstractmethod,Any,Optional,dataclass,field,list,dict,str,float,register_agent,AgentResponse,BenchmarkTask},
    frame=single, framerule=0.4pt,
    rulecolor=\color{black!25},
    numbers=none,
    xleftmargin=1em, framexleftmargin=0.8em,
    breaklines=true, showstringspaces=false
]
# --- base.py (verbatim) -----------------------------------
@dataclass
class AgentResponse:
    answer: Any                          # final prediction
    trajectory: list[dict] = ...        # action steps
    reasoning_trajectory: list[dict] = ... # reasoning trace
    token_usage: dict = ...             # token counts
    wall_time_sec: float = 0.0
    artifact_manifest: dict = ...       # workspace artifacts
    system_prompt: str = ""             # harness observability
    # ... (further provenance fields)

class Agent(ABC):
    @abstractmethod
    def setup(self, config: dict) -> None: ...

    @abstractmethod
    def solve(self, task: BenchmarkTask,
              sandbox: Optional[Any] = None) -> AgentResponse: ...

# --- custom implementation (minimal example) --------------
@register_agent("my_agent")
class MyAgent(Agent):
    def setup(self, config: dict) -> None:
        self.model = config["model"]

    def solve(self, task, sandbox=None) -> AgentResponse:
        answer = call_my_model(self.model, task.problem)
        return AgentResponse(answer=answer)
\end{lstlisting}
\end{nolinenumbers}

\subsubsection{Matched Harness Trace}
\label{app:matched-harness-trace}

The common loop aligns corresponding execution stages while preserving each harness's native behavior. Tab.~\ref{tab:matched-harness-trace} follows one matched example, Qwen3.5-27B on HLE item \texttt{hle\_1792}. All three harnesses use local computation, obtain $\min b(t)\approx1.0346$, and select answer E, but they reach that answer through different parsing, validation, feedback, and context-update paths.

\begin{table}[H]
\centering
\fontsize{8}{9.5}\selectfont
\renewcommand{\arraystretch}{1.18}
\setlength{\tabcolsep}{4pt}
\caption{\textbf{Matched native execution trace for one model and task.} The columns align the three harnesses to the common loop without replacing their native events or state representations.}
\label{tab:matched-harness-trace}
\begin{tabularx}{\linewidth}{@{}lYYY@{}}
\toprule
Harness & Model-facing context and proposal & Validation and feedback & Update and outcome \\
\midrule
OpenClaw & Gateway conversation produces a native local-computation request. & The controller dispatches the request; an \texttt{odeint} error returns as structured feedback. & The controller preserves the feedback, recovers with a corrected computation, and the model selects E. \\
ZeroClaw & CLI history produces a native action proposal. & Shell policy rejects one proposal before later file and shell actions are accepted. & The rejection and accepted-action results enter the expanded history, after which the model selects E. \\
OpenCode & Session messages contain a typed \texttt{bash} proposal. & The CLI permission path accepts the tool request and records its typed result. & The completed result remains in the rendered session context, after which the model selects E. \\
\bottomrule
\end{tabularx}
\end{table}

\begin{figure}[t]
\centering
\resizebox{\linewidth}{!}{%
\input{draw-figures/agent-module-inline}%
}
\caption{\textbf{Agent Module.} Concrete agent implementations (top) subclass the abstract \texttt{Agent} interface, which exposes \texttt{solve(task, sandbox) $\to$ AgentResponse}. The registry resolves an agent name from the YAML configuration at runtime. \texttt{AgentResponse} captures prediction, action and reasoning traces, resource metrics, workspace artifacts, and harness observability fields.}
\label{fig:agent-module}
\end{figure}
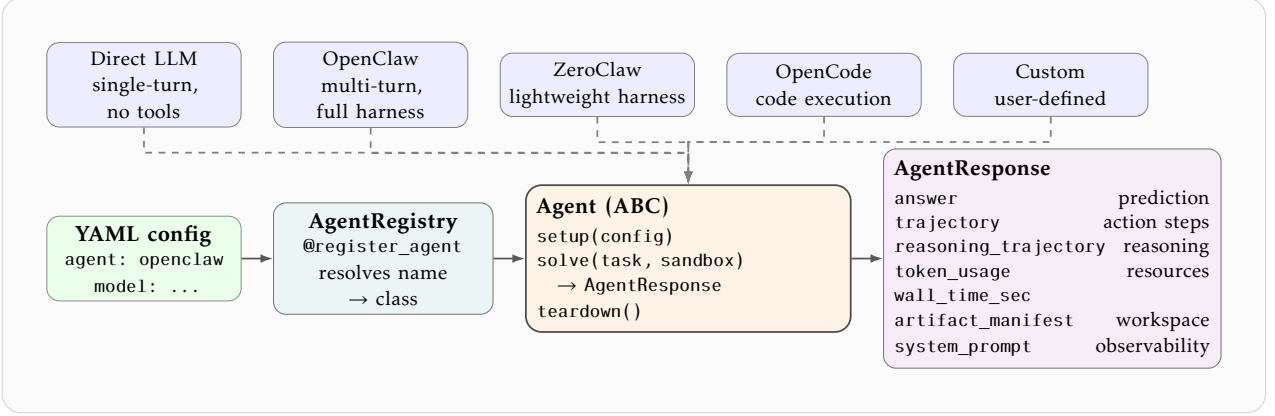

The example separates the shared evaluation contract from the system under study. AlphaDiana fixes the task, model, scorer, expected work, and saved fields, while each harness retains its native action grammar, approval path, error handling, and context representation. The standardized trajectory makes these stages comparable without requiring the three harnesses to execute through the same internal protocol.

\subsubsection{Measured Extension Footprints}
\label{app:extension-footprints}

The module boundary also localizes the work required to add a domain or harness. Tab.~\ref{tab:extension-footprints} reports measured implementation footprints for three completed extensions. The counts describe the code added for these integrations, not universal person-hour estimates; the required work depends on whether the extension can reuse an existing environment and verifier.

\begin{table}[H]
\centering
\fontsize{8.5}{10}\selectfont
\renewcommand{\arraystretch}{1.18}
\setlength{\tabcolsep}{5pt}
\caption{\textbf{Measured implementation footprints for three extension scenarios.} Shared runner and result-storage code is reused in every case.}
\label{tab:extension-footprints}
\begin{tabularx}{\linewidth}{@{}YYl@{}}
\toprule
Extension scenario & Components changed & Measured footprint \\
\midrule
Add MMMU-Pro as a non-interactive domain & Benchmark & 202 lines \\
Extend OpenClaw to Terminal-Bench 2 & Benchmark, environment, and OpenClaw integration & $\sim$2,150 shared + $\sim$850 harness-specific lines \\
Add Pi to the supported AIME 2024 domain & Agent & $\sim$520 lines + $\sim$45-line configuration \\
\bottomrule
\end{tabularx}
\end{table}

A compatible non-interactive domain may require only a benchmark adapter, as in MMMU-Pro. An interactive domain additionally requires task-environment and verifier support, as in Terminal-Bench 2. A new harness on a supported domain primarily requires an agent adapter, as in Pi on AIME 2024; an additional environment integration is needed only when the harness cannot reuse an existing backend. In all three cases, the Runner and standardized result format remain unchanged.

\subsection{Environment Module}
\label{sec:alphadiana_sandbox}

\begin{figure}[t]
\centering
\resizebox{\linewidth}{!}{%
\input{draw-figures/environment-module-inline}%
}
\caption{\textbf{Environment Module.} Backend implementations (top) subclass the abstract \texttt{SandboxSession} interface. The sandbox is optional: direct LLM evaluation requires no execution backend; harnessed agents (\textit{e.g.}, OpenClaw) use a ROCK-based containerized sandbox. The interface contract — \texttt{execute}, \texttt{upload}/\texttt{download}, \texttt{reset}, \texttt{close}, \texttt{metadata} — is the portability boundary that makes backends interchangeable.}
\label{fig:environment-module}
\end{figure}
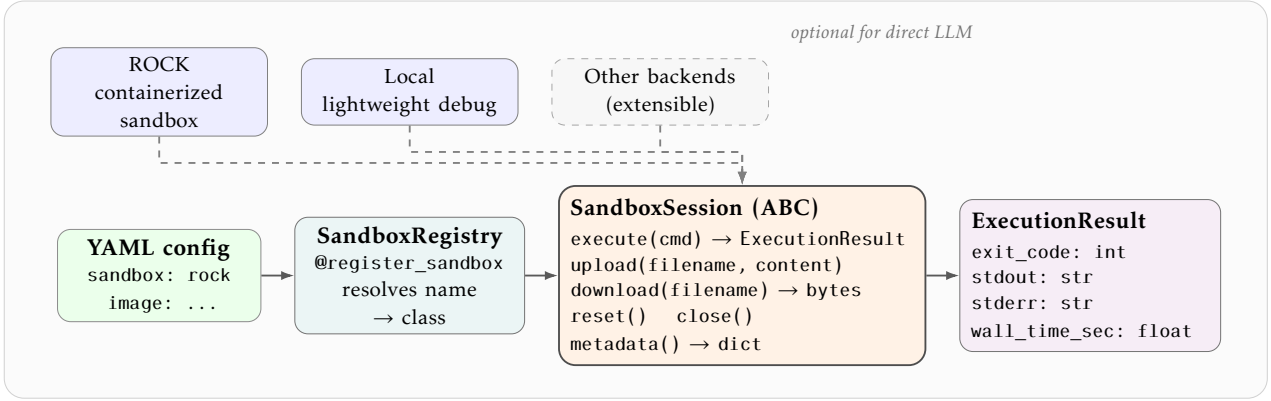

\noindent\textbf{Role.}
The environment module provides the external execution substrate for agents that need to run code, access files, invoke tools, or interact with task-specific artifacts. It is optional for direct-model evaluation but becomes a first-class component in harnessed settings because it determines where actions are executed, what resources are available, what limits apply, and what isolation boundary separates agent behavior from the evaluation host.

\noindent\textbf{Module boundary.}
\looseness=-1
The runner creates and manages the environment session, then passes it to the agent during task execution. The benchmark and scorer do not interact with the environment directly. This preserves the division of responsibilities: the benchmark specifies the task, the agent selects actions, the environment executes admissible actions, and the scorer evaluates the final prediction.

\Needspace{8\baselineskip}
\noindent\textbf{Session interface.}
An environment backend is exposed through a \texttt{SandboxSession} interface with the following core operations:
\[
\texttt{execute(command)} \rightarrow \texttt{ExecutionResult},
\]
\[
\texttt{upload(path, content)}, \qquad
\texttt{download(path)},
\]
\[
\texttt{reset()}, \qquad
\texttt{close()}, \qquad
\texttt{metadata()} \rightarrow \texttt{dict}.
\]
These operations support command execution, file transfer, task-level cleanup, session split, and provenance reporting. Any backend implementing this interface is interchangeable in AlphaDiana.

\noindent\textbf{State reset and reproducibility.}
The \texttt{reset()} operation is important for reproducibility. Agentic tasks often modify the workspace by creating files, installing packages, writing logs, editing repositories, or caching intermediate outputs. Without explicit reset semantics, later tasks may inherit hidden state from earlier tasks, making results order-dependent. AlphaDiana, therefore, treats reset behavior as part of the environment contract. A backend may implement reset by restoring a filesystem snapshot, clearing temporary directories, restarting services, or recreating task-local state.

\noindent\textbf{Session allocation and controlled reuse.}
\looseness=-1
As described in Sec.~\ref{sec:alphadiana_main}, parallel attempts acquire separate sessions when the backend supports pooling, and sessions are reset before reassignment. Under strict isolation, failed sandbox setup or incomplete pool preparation stops the run. Sequential attempts may reuse a session, while the adapter separately controls working files and native memory. Sharing within a task reinitializes the agent and replaces the shared session before the next task. Memory-sharing modes require sequential execution and disable automatic retries because a failed attempt may already have changed memory. The experimental persistence scopes are \mbox{specified in Appendix~\ref{app:memory-mechanisms}}.

\noindent\textbf{ROCK backend.}
The default implementation uses a ROCK-based containerized backend. ROCK isolates agent actions from the evaluation host, confines command execution and file edits to a sandbox boundary, records resource profiles and image provenance, and supports retrieval of workspace artifacts after completion. This standardizes the execution substrate across runs and reduces variation due to local machine state, dependency drift, or uncontrolled filesystem differences. These properties are especially important for evaluating agents on tasks involving code execution, terminal interaction, repository modification, or artifact generation.

\begin{nolinenumbers}
\begin{lstlisting}[
    language=Python,
    label=lst:sandbox,
    caption={Sandbox interface (\texttt{base.py}, verbatim) and a minimal custom backend.},
    basicstyle=\ttfamily\footnotesize\color{py-fg},
    backgroundcolor=\color{py-bg},
    keywordstyle=\color{py-keyword}\bfseries,
    keywordstyle={[2]\color{py-builtin}},
    stringstyle=\color{py-string},
    commentstyle=\color{py-comment}\itshape,
    morekeywords={[2]abstractmethod,dataclass,ABC,int,str,float,bytes,dict,Optional,register_sandbox,ExecutionResult,SandboxSession},
    frame=single, framerule=0.4pt,
    rulecolor=\color{black!25},
    numbers=none,
    xleftmargin=1em, framexleftmargin=0.8em,
    breaklines=true, showstringspaces=false
]
# --- base.py (verbatim) -----------------------------------
@dataclass
class ExecutionResult:
    exit_code: int
    stdout: str
    stderr: str
    wall_time_sec: float

class SandboxSession(ABC):
    @abstractmethod
    def execute(self, command: str) -> ExecutionResult: ...
    @abstractmethod
    def upload(self, filename: str, content: bytes) -> None: ...
    @abstractmethod
    def download(self, filename: str) -> bytes: ...
    @abstractmethod
    def close(self) -> None: ...
    def reset(self) -> None: ...         # clear state between tasks
    def metadata(self) -> dict: ...     # backend provenance

# --- custom backend (minimal example) --------------------
@register_sandbox("my_sandbox")
class MySandboxSession(SandboxSession):
    def execute(self, command: str) -> ExecutionResult:
        result = subprocess.run(command, shell=True, capture_output=True)
        return ExecutionResult(
            exit_code=result.returncode,
            stdout=result.stdout.decode(),
            stderr=result.stderr.decode(),
            wall_time_sec=0.0,
        )
    def upload(self, filename, content): ...
    def download(self, filename): ...
    def close(self): ...
\end{lstlisting}
\end{nolinenumbers}

\subsection{Scorer Module}
\label{sec:alphadiana_scorer}

\noindent\textbf{Role.}
At the conceptual level, $\mathcal{S}$ denotes two post-execution layers. The \emph{outcome-scoring layer} determines whether the final prediction solves the task and returns the task-level judgment $s$. The \emph{process-analysis layer} reads the saved trajectory $\tau$ and run metadata from $\mathcal{R}$ to measure action composition, token entropy, token and time use, completion, and failures. Both layers are external to generation: neither controls prompts, tool use, environment interaction, recovery behavior, or answer generation.

In the implementation, the abstract \texttt{Scorer} interface realizes the outcome-scoring layer. Given the task input $x$, reference object $y$, and prediction $\hat{y}$, it applies a shared verification rule and returns a \texttt{ScoreResult}. Trajectory analyzers remain separate utilities and run over persisted records after execution. This separation keeps correctness independent of harness-specific self-judgment while allowing outcome and process measurements to be reported together.

\noindent\textbf{ScoreResult.}
The output is a \texttt{ScoreResult}, which contains a binary \texttt{correct} label, a scalar \texttt{score}, the expected reference value, the predicted value, a rationale, and metadata. The binary label supports accuracy-style metrics, while the scalar score supports partial credit or continuous-valued grading. Metadata may store normalization details, verifier outputs, judge responses, tolerance settings, error messages, or failure labels.

\noindent\textbf{Built-in scoring strategies.}
AlphaDiana supports three built-in scoring strategies. Exact-match scoring compares normalized strings or categorical labels:
\[
\texttt{correct}
=
\mathbf{1}\!\left[
\mathrm{norm}(\hat{y}) =
\mathrm{norm}(y)
\right].
\]
Math-aware scoring first extracts the final answer and then applies symbolic, numeric, interval, unit, or equivalence checks:
\[
\texttt{correct}
=
\mathbf{1}\!\left[
\mathrm{verify}\bigl(\mathrm{extract}(\hat{y}), y\bigr)
\right].
\]
LLM-as-judge scoring delegates the judgment to an independent judge model or judge pipeline,
\[
\texttt{ScoreResult} = J_{\phi}(x,y,\hat{y}),
\]
using a fixed rubric, prompt, model, and decoding configuration recorded in the scorer metadata. This strategy is used when deterministic matching is too brittle, when multiple acceptable responses exist, or when partial-credit rubrics are required.

\noindent\textbf{Correctness versus system-level measurement.}
\looseness=-1
The outcome scorer determines task-level correctness, while the runner records the trajectory and system-level measurements required by the process-analysis layer, including wall-clock time, token usage, number of turns, tool-call counts, timeout rate, crash rate, and completion rate. Post-hoc analyzers additionally compute action and entropy statistics from these records. Keeping the two layers distinct in the implementation allows systems to be compared along multiple axes without conflating accuracy with resource use, reliability, or trajectory behavior.

\noindent\textbf{Interface.}
Adding a new scorer requires implementing:
\[
\texttt{score(task, response)} \rightarrow \texttt{ScoreResult}.
\]
\looseness=-1
The \texttt{task} argument provides the input, reference object, identifiers, metadata, and attachments; the \texttt{response} argument provides the prediction and agent-side metadata together with the stored trajectory when available. Outcome scorers consume the fields required by their verification rule, while process analyzers operate on the resulting records. Once registered, a scorer can be selected from the YAML experiment specification without modifying the benchmark, agent, environment, or runner.

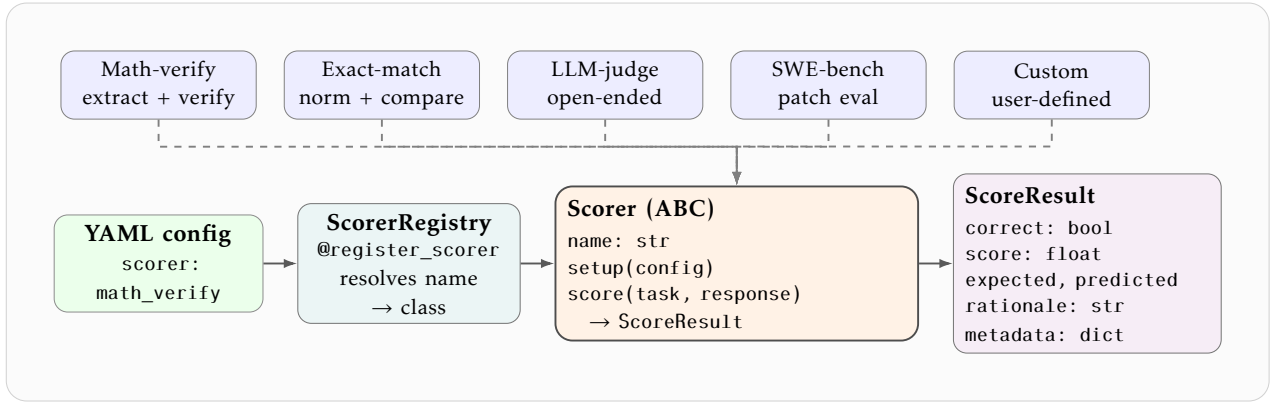
\begin{figure}[t]
\centering
\resizebox{\linewidth}{!}{%
\input{draw-figures/scorer-module-inline}%
}
\caption{\textbf{Outcome-scoring layer of $\mathcal{S}$.} Scorer implementations subclass the abstract \texttt{Scorer} interface, which maps a task and agent response to a \texttt{ScoreResult}. This interface implements task-level outcome judgment; the complementary process-analysis layer reads persisted trajectories and metadata from $\mathcal{R}$ after execution.}
\label{fig:scorer-module}
\end{figure}

\begin{nolinenumbers}
\begin{lstlisting}[
    language=Python,
    label=lst:scorer,
    caption={Scorer interface (\texttt{base.py}, verbatim) and a minimal custom implementation.},
    basicstyle=\ttfamily\footnotesize\color{py-fg},
    backgroundcolor=\color{py-bg},
    keywordstyle=\color{py-keyword}\bfseries,
    keywordstyle={[2]\color{py-builtin}},
    stringstyle=\color{py-string},
    commentstyle=\color{py-comment}\itshape,
    morekeywords={[2]abstractmethod,dataclass,field,ABC,bool,float,str,dict,Any,register_scorer,ScoreResult,BenchmarkTask,AgentResponse},
    frame=single, framerule=0.4pt,
    rulecolor=\color{black!25},
    numbers=none,
    xleftmargin=1em, framexleftmargin=0.8em,
    breaklines=true, showstringspaces=false
]
# --- base.py (verbatim) -----------------------------------
@dataclass
class ScoreResult:
    correct: bool
    score: float
    expected: Any
    predicted: Any
    rationale: str = ""
    metadata: dict = field(default_factory=dict)

class Scorer(ABC):
    @property
    @abstractmethod
    def name(self) -> str: ...

    def setup(self, config: dict) -> None: ...  # optional

    @abstractmethod
    def score(self, task: BenchmarkTask,
              response: AgentResponse) -> ScoreResult: ...

# --- custom implementation (minimal example) --------------
@register_scorer("my_scorer")
class MyScorer(Scorer):

    @property
    def name(self) -> str:
        return "my_scorer"

    def score(self, task, response) -> ScoreResult:
        correct = response.answer == task.ground_truth
        return ScoreResult(
            correct=correct, score=float(correct),
            expected=task.ground_truth,
            predicted=response.answer,
        )
\end{lstlisting}
\end{nolinenumbers}

\subsection{Runner and Execution Records}
\label{sec:alphadiana_runner}

\noindent\textbf{Role.}
\looseness=-1
The runner is the orchestration layer of AlphaDiana. It instantiates the benchmark, agent, environment, scorer, and configuration; manages the environment lifecycle; dispatches tasks to the agent; invokes the scorer after generation; and persists all information needed for later analysis. The runner does not solve tasks or judge correctness directly. It enforces the evaluation protocol $\mathcal{C}$.

\noindent\textbf{Result record.}
For each task attempt, the runner writes an append-only result record. Abstractly, the record for attempt $j$ on task $i$ is
\[
\rho_{i,j}
=
\bigl(
x_i,\,
y_i,\,
\hat{y}_{i,j},\,
s_{i,j},\,
\tau_{i,j},\,
z_{i,j}
\bigr),
\]
where $\tau_{i,j}$ is the trajectory produced by that attempt and $z_{i,j}$ denotes metadata collected from the benchmark, agent, environment, scorer, and run configuration. In single-sample evaluation, each task has one attempt, so we write $\rho_i$ as shorthand for $\rho_{i,1}$. In implementation, records are stored as JSONL entries containing task identity, attempt identity, prediction, score, trajectory, resource and efficiency metrics, environment provenance, harness observability fields, artifacts, raw response payloads, and termination information.

\noindent\textbf{Trajectory-aware analysis.}
This schema supports trajectory-aware evaluation. It preserves not only whether the agent solved the task, but also how the solution was produced, what resources were consumed, which tools were used, what feedback was observed, and why the run terminated. Downstream analyses can therefore compute accuracy, inspect failure modes, compare harness behavior, analyze tool-use patterns, and audit resource consumption without re-running the experiment.

\noindent\textbf{Failure preservation.}
The runner also preserves both system-level and model-level failures, including model API failures, invalid actions, harness crashes, environment timeouts, budget exhaustion, and scorer exceptions. The runner writes records with all available task, trajectory, metadata, error, and finish-reason fields. This prevents failed tasks from being silently discarded and allows completion rate, timeout rate, crash rate, and other reliability statistics to be computed from the same result store as accuracy.

\noindent\textbf{Run-level aggregation.}
Let $\mathcal{R}=\{\rho_{i,j}\}$ denote the saved attempt-level records for a run, where $\rho_{i,j}$ is the record for attempt $j$ on task $i$. In single-sample evaluations, each task has one attempt, so $\rho_i$ may be used as shorthand for $\rho_{i,1}$. Standard aggregate metrics include accuracy, mean score, completion rate, timeout rate, crash rate, mean wall-clock time, token usage, number of turns, and number of tool calls. These summaries are derived from the saved records rather than from transient in-memory state, so reports can be regenerated after the run finishes.
For multi-sample evaluation, AlphaDiana also supports common sampling-based metrics such as pass-rate and average correctness over repeated attempts for the same task. These metrics are computed by grouping attempt-level records by task identifier and aggregating the corresponding predictions, scores, trajectories, and metadata. This allows AlphaDiana to support both single-sample and multi-sample evaluation regimes without changing the underlying result schema.

\noindent\textbf{Retries and completion accounting.}
The append-only store can retain multiple records for one task and sample. Reports select the latest record for that pair; a retry does not create an additional sample. Correctness and execution status remain separate, and failures and unscored outputs are retained. Measurement-specific filters, such as requiring a recorded token count or a common task intersection, are reported in Appendix~\ref{app:cost-measurement}.

\noindent\textbf{Fine-grained provenance.}
Some analyses require information below the task level. For example, uncertainty estimation, calibration analysis, and token-level perplexity analysis require access to model log-probabilities when available. AlphaDiana therefore preserves raw model response payloads, including log-probability fields exposed by the model API, inside the response metadata. This enables downstream computation of decision uncertainty, confidence-correctness gaps, trajectory-level volatility, and related diagnostics.

\noindent\textbf{Interface boundary.}
\looseness=-1
Overall, the runner coordinates modules without collapsing their responsibilities. It does not embed benchmark-specific prompting, agent-specific scoring, or scorer-specific execution logic. This boundary makes AlphaDiana experiments reproducible and comparable: different model-harness systems can share the benchmark, model, scorer, and resource policy, while any harness-coupled environment backend is recorded explicitly and all trajectories and outcomes use a common schema.

\begin{figure}[t]
\centering
\resizebox{\linewidth}{!}{%
\input{draw-figures/runner-module-inline}%
}
\caption{\textbf{Runner.} The runner instantiates all components from a YAML configuration and executes the four-stage pipeline: load tasks $\to$ solve $\to$ score $\to$ persist. Each per-task result record captures correctness, execution traces, resource metrics, sandbox provenance, and harness observability fields in a single append-only JSONL entry.}
\label{fig:runner-module}
\end{figure}
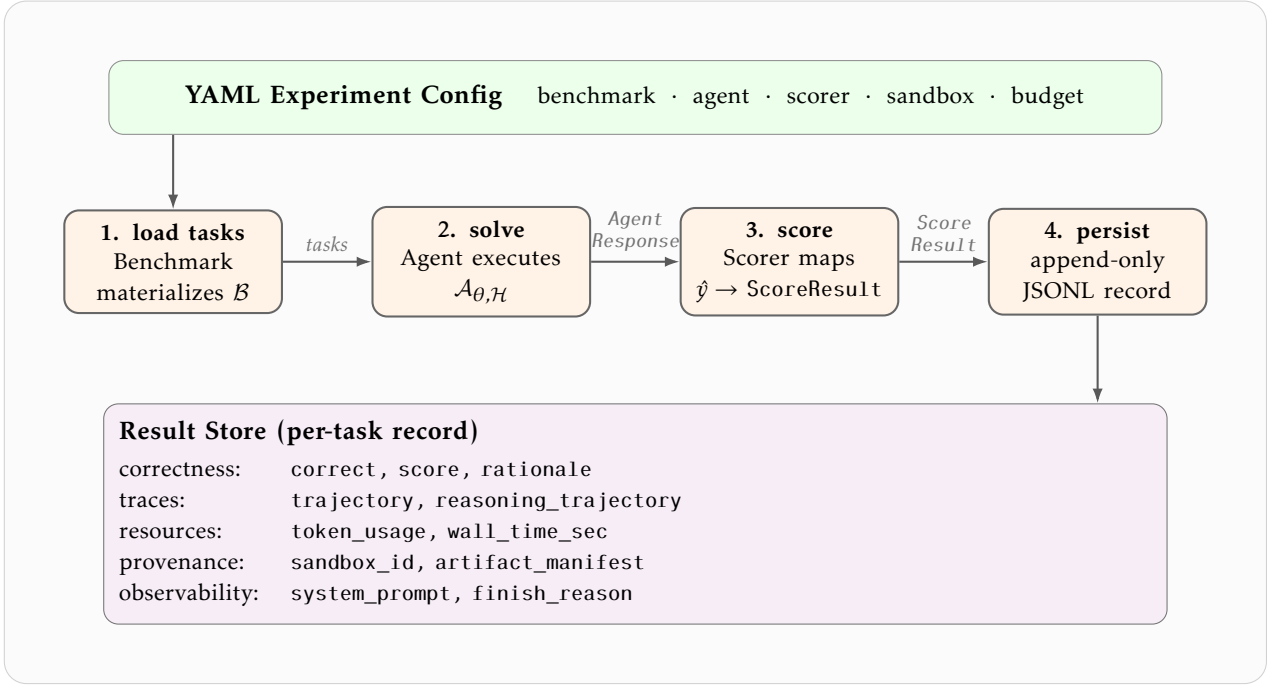

\begin{nolinenumbers}
\begin{lstlisting}[
    language=Python,
    label=lst:runner,
    caption={Minimal runner loop showing the four-stage pipeline. The actual \texttt{Runner} class handles concurrency, error recovery, and sandbox lifecycle; this excerpt shows the core per-task logic.},
    basicstyle=\ttfamily\footnotesize\color{py-fg},
    backgroundcolor=\color{py-bg},
    keywordstyle=\color{py-keyword}\bfseries,
    keywordstyle={[2]\color{py-builtin}},
    stringstyle=\color{py-string},
    commentstyle=\color{py-comment}\itshape,
    morekeywords={[2]ExperimentConfig,ResultStore,BenchmarkRegistry,AgentRegistry,ScorerRegistry,SandboxRegistry},
    frame=single, framerule=0.4pt,
    rulecolor=\color{black!25},
    numbers=none,
    xleftmargin=1em, framexleftmargin=0.8em,
    breaklines=true, showstringspaces=false
]
def run(config: ExperimentConfig, output_dir: str) -> None:
    # instantiate components from config
    benchmark = BenchmarkRegistry.get(config.benchmark_name)()
    agent     = AgentRegistry.get(config.agent_name)()
    scorer    = ScorerRegistry.get(config.scorer_name)()
    sandbox   = SandboxRegistry.get(config.sandbox_name)() \
                if config.sandbox_name else None

    benchmark.setup(config.benchmark_config)
    agent.setup(config.agent_config)
    scorer.setup(config.scorer_config)
    if sandbox: sandbox.setup(config.sandbox_config)

    store = ResultStore(output_dir, run_id=config.run_id)

    # four-stage pipeline
    for task in benchmark.load_tasks(config.benchmark_config): # 1. load
        session = sandbox.create_session() if sandbox else None
        response = agent.solve(task, session)                   # 2. solve
        result   = scorer.score(task, response)                 # 3. score
        store.append(task, response, result)                    # 4. persist
        if session: session.close()
\end{lstlisting}
\end{nolinenumbers}

\subsection{Run Configuration and Inspection}
\label{app: alphaDiana reproduction}

\begin{figure}[ht]
    \centering
    \subfigure[{Create a new evaluation}]{\includegraphics[width=0.489\linewidth, trim=0pt 20pt 0pt 20pt, clip]{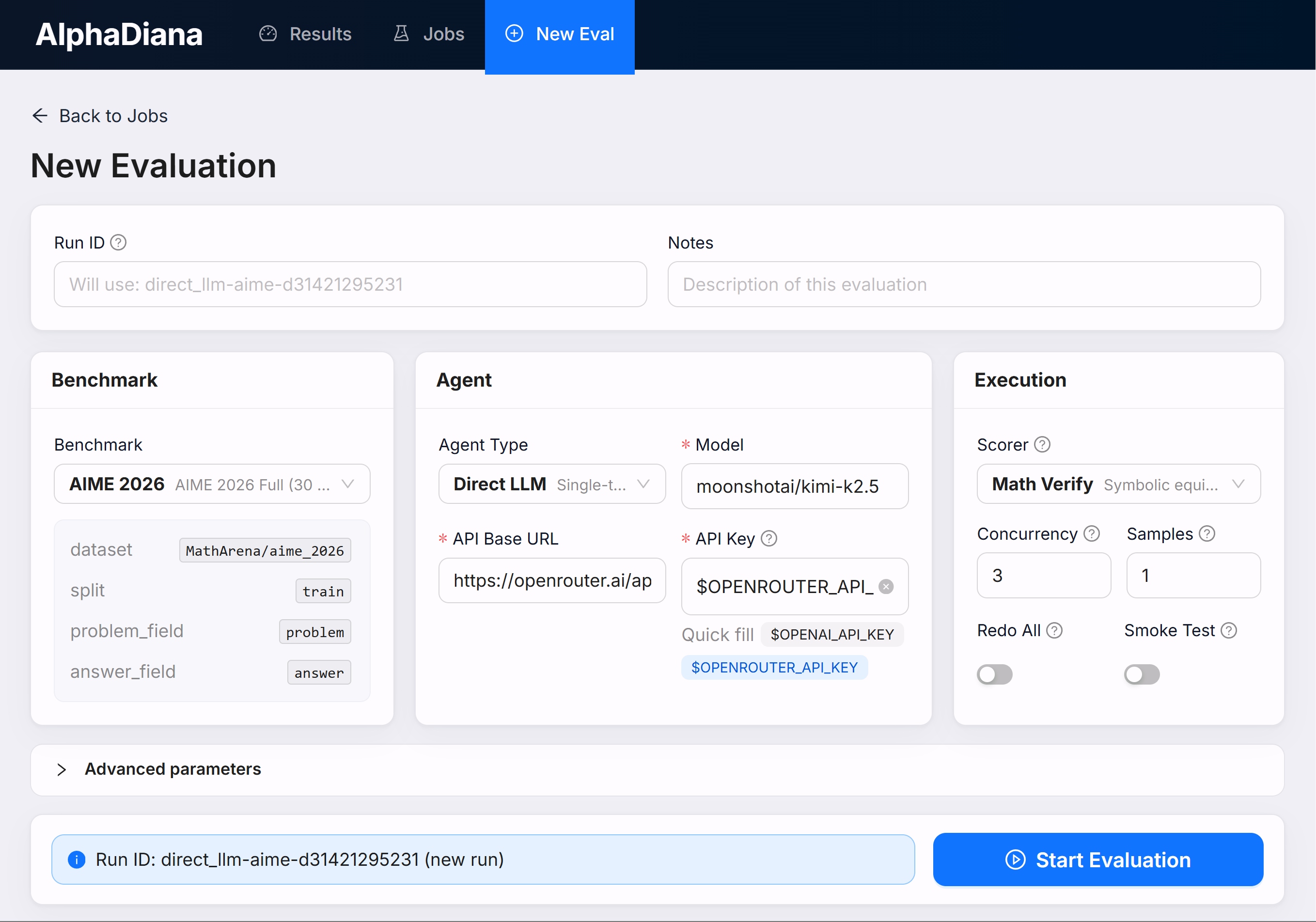}}
    \hfill
    \subfigure[{Obtain the evaluation results}]{\includegraphics[width=0.461\linewidth, trim=0pt 40pt 0pt 0pt, clip]{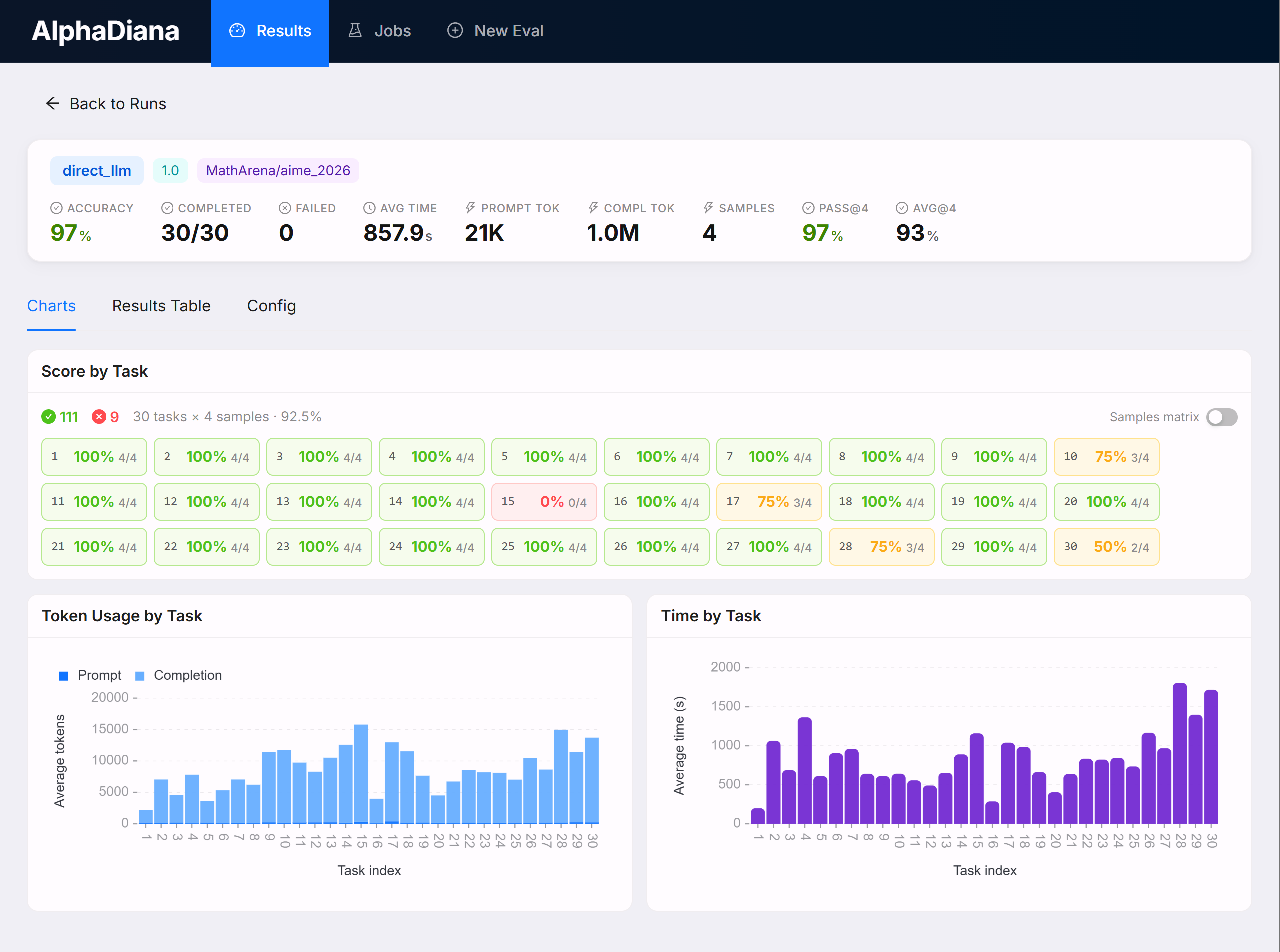}}
    \caption{\looseness=-1\textbf{AlphaDiana dashboard demo.} \textbf{(a)} The run-creation interface instantiates an evaluation from explicit configuration fields. In this example, the benchmark is AIME 2026, the agent is Direct LLM (the \emph{Direct} baseline), and the remaining fields set the model and scorer, which correspond to the run configuration $\mathcal{C}$. \textbf{(b)} The dashboard of results $\mathcal{R}$ visualizes an executed AIME 2026 run with token and time statistics. The summary reports $30/30$ completed tasks, zero failed runs, $97\%$ Pass@4, and $93\%$ Avg@4.}
    \label{fig:alphadiana-dashboard}
\end{figure}

Fig.~\ref{fig:alphadiana-dashboard} illustrates configuration and result inspection. Its displayed run is a dashboard example; the paper's experimental settings and computing resources are documented in Appendix~\ref{app:iclr-main-details}. Saved task and sample identifiers link outcomes to traces for offline analysis without rerunning inference.

%% file: draw-figures/benchmark-module-inline.tex
\begin{tikzpicture}[
    font=\small,
    >=Latex,
    implbox/.style={
        draw=black!50, rounded corners=2mm,
        align=center, inner sep=4pt,
        fill=blue!7, text width=26mm,
        minimum height=10mm
    },
    abcbox/.style={
        draw=black!70, rounded corners=2.5mm, thick,
        align=left, inner sep=5pt,
        fill=orange!10, text width=50mm
    },
    fieldbox/.style={
        draw=black!55, rounded corners=2mm,
        align=left, inner sep=5pt,
        fill=violet!7, text width=44mm
    },
    regbox/.style={
        draw=black!55, rounded corners=2mm,
        align=center, inner sep=4pt,
        fill=teal!8, text width=32mm,
        minimum height=10mm
    },
    configbox/.style={
        draw=black!45, rounded corners=2mm,
        align=center, inner sep=4pt,
        fill=green!8, text width=28mm,
        minimum height=10mm
    },
    labelnode/.style={font=\scriptsize\itshape, text=black!55},
    arr/.style={-{Latex[length=2mm,width=1.4mm]}, thick, draw=black!65},
    darr/.style={-{Latex[length=2mm,width=1.4mm]}, thick, draw=black!50, dashed}
]

\node[implbox] (math)  at (0,0)
    {\footnotesize Math reasoning\\\footnotesize (AIME, IMO)};
\node[implbox, right=5mm of math] (know)
    {\footnotesize Verifiable knowledge\\\footnotesize (HLE, GPQA)};
\node[implbox, right=5mm of know] (mm)
    {\footnotesize Multimodal\\\footnotesize (MMMU-Pro)};
\node[implbox, right=5mm of mm] (term)
    {\footnotesize Terminal execution\\\footnotesize (TerminalBench-2)};
\node[implbox, right=5mm of term] (swe)
    {\footnotesize Software engineering\\\footnotesize (SWE-bench)};
\node[implbox, right=5mm of swe] (cust)
    {\footnotesize Custom\\\footnotesize (user JSONL)};

\node[configbox, below=14mm of math] (config)
    {\textbf{YAML config}\\
     \footnotesize\texttt{benchmark: aime}\\
     \footnotesize\texttt{dataset: ...}};

\node[regbox, right=5mm of config, minimum width=42mm] (registry)
    {\textbf{BenchmarkRegistry}\\
     \footnotesize\texttt{@register\_benchmark}\\
     \footnotesize resolves name $\to$ class};

\node[abcbox, right=5mm of registry] (abc)
    {\textbf{Benchmark (ABC)}\\[2pt]
     \footnotesize\texttt{name: str}\\
     \footnotesize\texttt{load\_tasks(config)}\\
     \footnotesize\quad\texttt{$\to$ list[BenchmarkTask]}\\
     \footnotesize\texttt{default\_scorer() $\to$ str}};

\node[fieldbox, right=5mm of abc, minimum width=65mm] (task)
    {\textbf{BenchmarkTask}\\[2pt]
     \footnotesize\texttt{task\_id:\;str}\\
     \footnotesize\texttt{problem:\;str}\\
     \footnotesize\texttt{ground\_truth:\;Any}\\
     \footnotesize\texttt{metadata:\;dict}\\
     \scriptsize\texttt{attachments:\;dict[str,bytes]}};

\foreach \src in {math, know, mm, term, swe, cust} {
    \draw[darr] (\src.south) -- ++(0,-4mm) -| (abc.north);
}

\draw[arr] (config.east) -- (registry.west);
\draw[arr] (registry.east) -- (abc.west);
\draw[arr] (abc.east) -- (task.west);

\begin{scope}[on background layer]
\node[
    draw=black!18, rounded corners=3mm,
    inner sep=7mm, fill=gray!3,
    fit=(math)(cust)(config)(task)
] (frame) {};
\end{scope}

\end{tikzpicture}

%% file: draw-figures/agent-module-inline.tex
\begin{tikzpicture}[
    font=\small,
    >=Latex,
    implbox/.style={
        draw=black!50, rounded corners=2mm,
        align=center, inner sep=4pt,
        fill=blue!7, text width=28mm,
        minimum height=10mm
    },
    abcbox/.style={
        draw=black!70, rounded corners=2.5mm, thick,
        align=left, inner sep=5pt,
        fill=orange!10, text width=48mm
    },
    respbox/.style={
        draw=black!55, rounded corners=2mm,
        align=left, inner sep=5pt,
        fill=violet!7, text width=50mm
    },
    configbox/.style={
        draw=black!45, rounded corners=2mm,
        align=center, inner sep=4pt,
        fill=green!8, text width=28mm,
        minimum height=10mm
    },
    regbox/.style={
        draw=black!55, rounded corners=2mm,
        align=center, inner sep=4pt,
        fill=teal!8, text width=32mm,
        minimum height=10mm
    },
    labelnode/.style={font=\scriptsize\itshape, text=black!55},
    arr/.style={-{Latex[length=2mm,width=1.4mm]}, thick, draw=black!65},
    darr/.style={-{Latex[length=2mm,width=1.4mm]}, thick, draw=black!50, dashed}
]

\node[implbox] (direct) at (0,0)
    {\footnotesize Direct LLM\\\footnotesize single-turn, no tools};
\node[implbox, right=5mm of direct] (openclaw)
    {\footnotesize OpenClaw\\\footnotesize multi-turn, full harness};
\node[implbox, right=5mm of openclaw] (zeroclaw)
    {\footnotesize ZeroClaw\\\footnotesize lightweight harness};
\node[implbox, right=5mm of zeroclaw] (opencode)
    {\footnotesize OpenCode\\\footnotesize code execution};
\node[implbox, right=5mm of opencode] (custom)
    {\footnotesize Custom\\\footnotesize user-defined};

\node[configbox, below=14mm of direct] (config)
    {\textbf{YAML config}\\
     \footnotesize\texttt{agent: openclaw}\\
     \footnotesize\texttt{model: ...}};

\node[regbox, right=5mm of config] (registry)
    {\textbf{AgentRegistry}\\
     \footnotesize\texttt{@register\_agent}\\
     \footnotesize resolves name $\to$ class};

\node[abcbox, right=5mm of registry] (abc)
    {\textbf{Agent (ABC)}\\[2pt]
     \footnotesize\texttt{setup(config)}\\
     \footnotesize\texttt{solve(task, sandbox)}\\
     \footnotesize\quad\texttt{$\to$ AgentResponse}\\
     \footnotesize\texttt{teardown()}};

\node[respbox, right=5mm of abc] (resp)
    {\textbf{AgentResponse}\\[2pt]
     \footnotesize\texttt{answer} \hfill prediction\\
     \footnotesize\texttt{trajectory} \hfill action steps\\
     \footnotesize\texttt{reasoning\_trajectory} \hfill reasoning\\
     \footnotesize\texttt{token\_usage} \hfill resources\\
     \footnotesize\texttt{wall\_time\_sec}\\
     \footnotesize\texttt{artifact\_manifest} \hfill workspace\\
     \footnotesize\texttt{system\_prompt} \hfill observability};

\foreach \src in {direct, openclaw, zeroclaw, opencode, custom} {
    \draw[darr] (\src.south) -- ++(0,-4mm) -| (abc.north);
}

\draw[arr] (config.east) -- (registry.west);
\draw[arr] (registry.east) -- (abc.west);
\draw[arr] (abc.east) -- (resp.west);

\begin{scope}[on background layer]
\node[
    draw=black!18, rounded corners=3mm,
    inner sep=7mm, fill=gray!3,
    fit=(direct)(custom)(config)(resp)
] (frame) {};
\end{scope}

\end{tikzpicture}

%% file: draw-figures/environment-module-inline.tex
\begin{tikzpicture}[
    font=\small,
    >=Latex,
    implbox/.style={
        draw=black!50, rounded corners=2mm,
        align=center, inner sep=4pt,
        fill=blue!7, text width=30mm,
        minimum height=10mm
    },
    abcbox/.style={
        draw=black!70, rounded corners=2.5mm, thick,
        align=left, inner sep=5pt,
        fill=orange!10, text width=52mm
    },
    resbox/.style={
        draw=black!55, rounded corners=2mm,
        align=left, inner sep=5pt,
        fill=violet!7, text width=36mm
    },
    configbox/.style={
        draw=black!45, rounded corners=2mm,
        align=center, inner sep=4pt,
        fill=green!8, text width=28mm,
        minimum height=10mm
    },
    regbox/.style={
        draw=black!55, rounded corners=2mm,
        align=center, inner sep=4pt,
        fill=teal!8, text width=32mm,
        minimum height=10mm
    },
    optbox/.style={
        draw=black!35, rounded corners=2mm,
        align=center, inner sep=4pt,
        fill=gray!6, text width=30mm,
        minimum height=10mm,
        dashed
    },
    labelnode/.style={font=\scriptsize\itshape, text=black!55},
    arr/.style={-{Latex[length=2mm,width=1.4mm]}, thick, draw=black!65},
    darr/.style={-{Latex[length=2mm,width=1.4mm]}, thick, draw=black!50, dashed}
]

\node[implbox] (rock) at (0,0)
    {\footnotesize ROCK\\\footnotesize containerized sandbox};
\node[implbox, right=5mm of rock] (local)
    {\footnotesize Local\\\footnotesize lightweight debug};
\node[optbox, right=5mm of local] (other)
    {\footnotesize Other backends\\\footnotesize (extensible)};

\node[configbox, below=14mm of rock] (config)
    {\textbf{YAML config}\\
     \footnotesize\texttt{sandbox: rock}\\
     \footnotesize\texttt{image: ...}};

\node[regbox, right=5mm of config] (registry)
    {\textbf{SandboxRegistry}\\
     \footnotesize\texttt{@register\_sandbox}\\
     \footnotesize resolves name $\to$ class};

\node[abcbox, right=5mm of registry] (abc)
    {\textbf{SandboxSession (ABC)}\\[2pt]
     \footnotesize\texttt{execute(cmd) $\to$ ExecutionResult}\\
     \footnotesize\texttt{upload(filename, content)}\\
     \footnotesize\texttt{download(filename) $\to$ bytes}\\
     \footnotesize\texttt{reset()} \quad\texttt{close()}\\
     \footnotesize\texttt{metadata() $\to$ dict}};

\node[resbox, right=5mm of abc] (res)
    {\textbf{ExecutionResult}\\[2pt]
     \footnotesize\texttt{exit\_code: int}\\
     \footnotesize\texttt{stdout: str}\\
     \footnotesize\texttt{stderr: str}\\
     \footnotesize\texttt{wall\_time\_sec: float}};

\foreach \src in {rock, local, other} {
    \draw[darr] (\src.south) -- ++(0,-4mm) -| (abc.north);
}

\draw[arr] (config.east) -- (registry.west);
\draw[arr] (registry.east) -- (abc.west);
\draw[arr] (abc.east) -- (res.west);

\node[labelnode, above right=1mm and 2mm of other] {optional for direct LLM};

\begin{scope}[on background layer]
\node[
    draw=black!18, rounded corners=3mm,
    inner sep=7mm, fill=gray!3,
    fit=(rock)(other)(config)(res)
] (frame) {};
\end{scope}

\end{tikzpicture}

%% file: draw-figures/scorer-module-inline.tex
\begin{tikzpicture}[
    font=\small,
    >=Latex,
    implbox/.style={
        draw=black!50, rounded corners=2mm,
        align=center, inner sep=4pt,
        fill=blue!7, text width=26mm,
        minimum height=10mm
    },
    abcbox/.style={
        draw=black!70, rounded corners=2.5mm, thick,
        align=left, inner sep=5pt,
        fill=orange!10, text width=50mm
    },
    resbox/.style={
        draw=black!55, rounded corners=2mm,
        align=left, inner sep=5pt,
        fill=violet!7, text width=36mm
    },
    configbox/.style={
        draw=black!45, rounded corners=2mm,
        align=center, inner sep=4pt,
        fill=green!8, text width=28mm,
        minimum height=10mm
    },
    regbox/.style={
        draw=black!55, rounded corners=2mm,
        align=center, inner sep=4pt,
        fill=teal!8, text width=30mm,
        minimum height=10mm
    },
    labelnode/.style={font=\scriptsize\itshape, text=black!55},
    arr/.style={-{Latex[length=2mm,width=1.4mm]}, thick, draw=black!65},
    darr/.style={-{Latex[length=2mm,width=1.4mm]}, thick, draw=black!50, dashed}
]

\node[implbox] (math) at (0,0)
    {\footnotesize Math-verify\\\footnotesize extract + verify};
\node[implbox, right=4mm of math] (exact)
    {\footnotesize Exact-match\\\footnotesize norm + compare};
\node[implbox, right=4mm of exact] (llm)
    {\footnotesize LLM-judge\\\footnotesize open-ended};
\node[implbox, right=4mm of llm] (swe)
    {\footnotesize SWE-bench\\\footnotesize patch eval};
\node[implbox, right=4mm of swe] (custom)
    {\footnotesize Custom\\\footnotesize user-defined};

\node[configbox, below=14mm of math] (config)
    {\textbf{YAML config}\\
     \footnotesize\texttt{scorer: math\_verify}};

\node[regbox, right=5mm of config] (registry)
    {\textbf{ScorerRegistry}\\
     \footnotesize\texttt{@register\_scorer}\\
     \footnotesize resolves name $\to$ class};

\node[abcbox, right=5mm of registry] (abc)
    {\textbf{Scorer (ABC)}\\[2pt]
     \footnotesize\texttt{name: str}\\
     \footnotesize\texttt{setup(config)}\\
     \footnotesize\texttt{score(task, response)}\\
     \footnotesize\quad\texttt{$\to$ ScoreResult}};

\node[resbox, right=5mm of abc] (res)
    {\textbf{ScoreResult}\\[2pt]
     \footnotesize\texttt{correct: bool}\\
     \footnotesize\texttt{score: float}\\
     \footnotesize\texttt{expected, predicted}\\
     \footnotesize\texttt{rationale: str}\\
     \footnotesize\texttt{metadata: dict}};

\foreach \src in {math, exact, llm, swe, custom} {
    \draw[darr] (\src.south) -- ++(0,-4mm) -| (abc.north);
}

\draw[arr] (config.east) -- (registry.west);
\draw[arr] (registry.east) -- (abc.west);
\draw[arr] (abc.east) -- (res.west);

\begin{scope}[on background layer]
\node[
    draw=black!18, rounded corners=3mm,
    inner sep=7mm, fill=gray!3,
    fit=(math)(custom)(config)(res)
] (frame) {};
\end{scope}

\end{tikzpicture}

%% file: draw-figures/runner-module-inline.tex
\begin{tikzpicture}[
    font=\small,
    >=Latex,
    pipenode/.style={
        draw=black!60, rounded corners=2mm, thick,
        align=center, inner sep=5pt,
        fill=orange!10, text width=26mm,
        minimum height=12mm
    },
    storebox/.style={
        draw=black!55, rounded corners=2mm,
        align=left, inner sep=6pt,
        fill=violet!7, text width=140mm
    },
    configbox/.style={
        draw=black!45, rounded corners=2mm,
        align=center, inner sep=4pt,
        fill=green!8, text width=140mm,
        minimum height=10mm
    },
    labelnode/.style={font=\scriptsize\itshape, text=black!55, align=center},
    arr/.style={-{Latex[length=2mm,width=1.4mm]}, thick, draw=black!65}
]

\node[pipenode] (load) at (0,0)
    {\footnotesize\textbf{1. load tasks}\\
     \footnotesize Benchmark\\materializes $\mathcal{B}$};

\node[pipenode, right=12mm of load] (solve)
    {\footnotesize\textbf{2. solve}\\
     \footnotesize Agent executes\\$\mathcal{A}_{\theta, \mathcal{H}}$};

\node[pipenode, right=12mm of solve] (score)
    {\footnotesize\textbf{3. score}\\
     \footnotesize Scorer maps\\$\hat{y} \to$ \texttt{ScoreResult}};

\node[pipenode, right=12mm of score] (persist)
    {\footnotesize\textbf{4. persist}\\
     \footnotesize append-only\\JSONL record};

\node[configbox, above=10mm of $(load.north)!0.5!(persist.north)$] (config)
    {\textbf{YAML Experiment Config} \quad
     \footnotesize benchmark \;·\; agent \;·\; scorer \;·\; sandbox \;·\; budget};

\node[storebox, below=12mm of $(load.south)!0.5!(persist.south)$] (store)
    {\textbf{Result Store (per-task record)}\\[3pt]
     \footnotesize
     \begin{tabular}{@{}ll@{}}
     correctness: & \texttt{correct, score, rationale} \\[1pt]
     traces:      & \texttt{trajectory, reasoning\_trajectory} \\[1pt]
     resources:   & \texttt{token\_usage,\ wall\_time\_sec} \\[1pt]
     provenance:  & \texttt{sandbox\_id,\ artifact\_manifest} \\[1pt]
     observability: & \texttt{system\_prompt,\ finish\_reason} \\
     \end{tabular}};


\draw[arr] (config.south -| load.north) -- (load.north);

\draw[arr] (load.east) -- (solve.west)
    node[midway, above, labelnode] {tasks};
\draw[arr] (solve.east) -- (score.west)
    node[midway, above, labelnode] {\texttt{Agent}\\[-1pt]\texttt{Response}};
\draw[arr] (score.east) -- (persist.west)
    node[midway, above, labelnode] {\texttt{Score}\\[-1pt]\texttt{Result}};

\draw[arr] (persist.south) -- (persist.south |- store.north);

\begin{scope}[on background layer]
\node[
    draw=black!18, rounded corners=3mm,
    inner sep=8mm, fill=gray!3,
    fit=(config)(load)(store)(persist)
] (frame) {};
\end{scope}

\end{tikzpicture}

%% file: appendix/0-iclr-main-details.tex
\section{Experimental Setup and Reproducibility}
\label{app:iclr-main-details}

This appendix specifies the shared experimental settings, resource limits, and measurement coverage for Secs.~\ref{sec:macro} and~\ref{sec:micro}. Analysis methods and supplementary macro results are collected in Appendix~\ref{app:macro}; intervention-specific configurations are collected in Appendix~\ref{app:component-intervention-details}.

\subsection{Evaluation Setup}
\label{app:iclr-setup}

\begin{figure}[ht]
    \centering
    \resizebox{\linewidth}{!}{\input{draw-figures/harness-system-components}}
    \caption{\textbf{Functional anatomy of a native harness.} The interface exchanges prompts and outputs with the fixed model. The control logic advances the interaction while reading or updating runtime state. Direct uses one response, with no tools, control loop, or runtime state.}
    \label{fig:harness-system-components}
\end{figure}
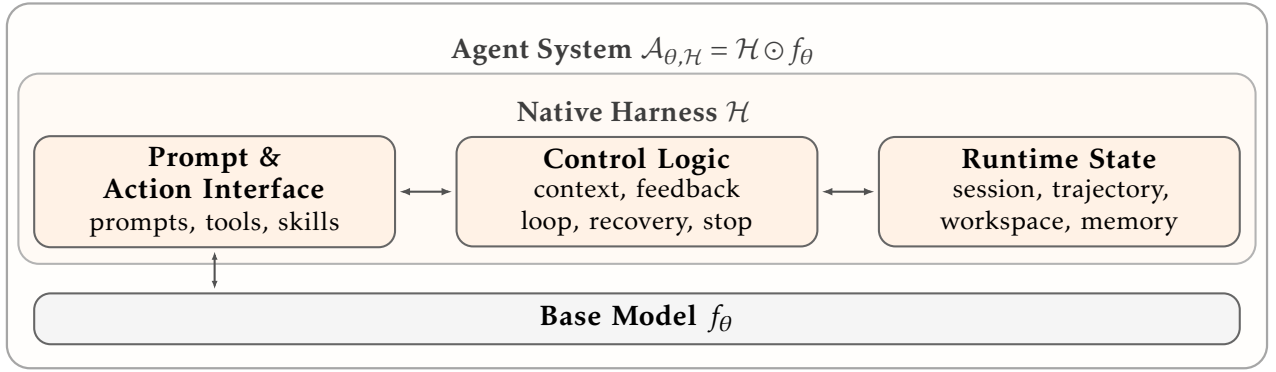

Tab.~\ref{tab:implementation-benchmark-contract} lists the task inputs, outcome scorers, and reported metrics of each benchmark $\mathcal{B}$. Tab.~\ref{tab:implementation-harness-adapters} shows how each evaluated system implements the three functions of a native harness in Fig.~\ref{fig:harness-system-components}: the prompt and action interface, control logic, and runtime state. During execution, each harness $\mathcal{H}$ presents the task input $x$ to the base model $f_\theta$ with its native prompts and available tools; if the model produces an admissible action, the harness sends it to $\mathcal{E}$, adds the returned feedback to the next prompt, and continues while maintaining its session, trajectory, workspace, and any native memory. Fig.~\ref{fig:alphadiana-dashboard} shows the dashboard that instantiates a run from explicit configuration fields and inspects its saved records.

\noindent\textbf{Harness defaults and interactive tasks.} OpenClaw, ZeroClaw, and OpenCode use default configurations, with native memory available but no explicit instruction to use it. \emph{Direct} uses a single model response without a tool loop. Interactive evaluations use Qwen3.5-27B and Gemma-4-31B-IT: SWE Mini contains 50 SWE-bench Verified tasks and uses SWE-Agent as the reference; Terminal-Bench~2 contains 89 tasks and uses terminus-2. Their Avg@1 scores count runtime and evaluator failures as unsuccessful attempts.

\noindent\textbf{Sampling, decoding, and budgets.} Macro results use one sample per problem (Avg@1), except on AIME~26, which has 30 problems and draws four samples each at temperature $0.6$. Pass@4 is the fraction of tasks solved at least once; Avg@4 is the mean accuracy across all attempts. Within a model--task block, systems share the task list and a decoding policy of top-$p$ $0.95$ with a per-request generation limit of \texttt{max\_tokens}$=131{,}072$. Appendix~\ref{app:wall-time-sensitivity} gives the wall-clock time and iteration caps and explains why these matched limits do not equalize realized compute.

\begin{table}[ht]
\centering
\footnotesize
\setlength{\tabcolsep}{2.4pt}
\renewcommand{\arraystretch}{1.05}
\caption{\textbf{Benchmark inputs, scorers, and reported metrics.} Rows follow Fig.~\ref{fig:macro-main-results}.}
\label{tab:implementation-benchmark-contract}
\begin{tabularx}{\linewidth}{@{}lYYl@{}}
\toprule
Benchmark $\mathcal B$ & Task input & Outcome scorer & Reported metric \\
\midrule
AIME 26 & Text & \texttt{numeric} & Pass@4, Avg@4 \\
IMO-AnswerBench & Text & \texttt{imo\_verify} & Avg@1 \\
GPQA-Diamond & Text & \texttt{exact\_match} & Avg@1 \\
HLE-Verifiable & Text & \texttt{exact\_match} & Avg@1 \\
MMMU-Pro & Text + image & \texttt{exact\_match} & Avg@1 \\
SWE Mini & Issue + repository & \texttt{swe\_bench} & Avg@1 \\
Terminal-Bench 2 & Instruction + container & \texttt{terminal\_bench2} & Avg@1 \\
\bottomrule
\end{tabularx}
\end{table}

\begin{table}[ht]
\centering
\scriptsize
\setlength{\tabcolsep}{3.8pt}
\renewcommand{\arraystretch}{1.05}
\caption{\textbf{Harness components and settings.} The three component columns follow Fig.~\ref{fig:harness-system-components}; the last two rows are the baseline systems for SWE Mini and Terminal-Bench 2.}
\label{tab:implementation-harness-adapters}
\begin{tabularx}{\linewidth}{@{}lYYY@{}}
\toprule
System and setting & Prompt and action interface & Control logic & Runtime state \\
\midrule
Direct (Base model) & Minimal task prompt; no tools or actions & Single model response; no harness control loop & No harness state \\
OpenClaw (2026.3.7) & Native prompt and tools; no added skills & Prompt updates, interaction loop, recovery, stopping & Session and workspace for each attempt; memory use not elicited \\
OpenCode (v1.3.2) & System and task prompts; native tools; no added skills & Interaction loop, run policies, recovery, stopping & New session and workspace; memory use not elicited \\
ZeroClaw (v0.6.9) & Native prompt and configurable tools; no added skills & Interaction loop, run policies, turn and stopping limits & Session and workspace for each task; memory use not elicited \\
SWE-Agent (v1.1.0; SWE Mini baseline) & SWE Mini task prompt and file editing tools & Interaction loop, inspection, editing, patch submission & Task container and trajectory; no state shared across tasks \\
terminus-2 (Terminal-Bench 2 baseline) & Terminal-Bench 2 task prompt and native tool calls & Interaction loop, execution, feedback, stopping & Task container; no state shared across tasks \\
\bottomrule
\end{tabularx}
\end{table}

\subsection{Resource Measurement and Observed Task Sets}
\label{app:cost-measurement}

We report recorded output tokens in thousands, excluding input tokens, and wall-clock time in seconds. Both measures include work spent on unsuccessful attempts.

\noindent\textbf{Macro comparisons.} Fig.~\ref{fig:macro-efficiency} uses each system's available cohort without enforcing a common task intersection. Token means include only attempts with recorded counts. Token capture is incomplete for OpenClaw and ZeroClaw on Qwen/AIME~26, all three harnesses on Qwen/GPQA-Diamond, ZeroClaw on both models/HLE-Verifiable, and OpenClaw and ZeroClaw on both models/MMMU-Pro. Gemma/MMMU-Pro/OpenClaw has 1,570 attempts, compared with 1,730 for the other systems. The two comparisons discussed in the main text have complete token coverage: 120 attempts each for Direct and OpenCode on Qwen/AIME~26, and 198 tasks each for Direct and ZeroClaw on Gemma/GPQA-Diamond. Recorded wall-clock time includes model serving, tool execution, and scheduling. The figure summarizes observed runs rather than a standardized wall-clock time experiment. In particular, Gemma/AIME~26 source runs differ in execution date and serving endpoint. Serving configuration, concurrency, load, and timing boundaries are not fully matched across the source runs. These summaries therefore describe observed wall-clock time, without isolating harness overhead.

\noindent\textbf{Output tokens per correct answer.} Figs.~\ref{fig:micro-solution-cost} and~\ref{fig:micro-token-allocation} use common Gemma task sets of 198 GPQA-Diamond and 587 HLE-Verifiable tasks. Four HLE tasks with missing ZeroClaw token captures are excluded from every system. All four excluded ZeroClaw attempts were unsuccessful, so results describe the observed subset. Output tokens per attempt divide total recorded output by the number of attempts. Output tokens per correct answer divide that same total, including output from unsuccessful attempts, by the number of correct answers.

\noindent\textbf{Token allocation.} Unsuccessful attempts include incorrect answers, invalid or unscored outputs, and operational failures. The highest-token group contains 20 of 198 GPQA attempts and 59 of 587 HLE attempts, rounding 10\% upward. Its token share overlaps with the unsuccessful-attempt share, so the two shares should not be added.

\subsection{Reproduction Resources and Configuration Scope}
\label{app:reproduction-resources}

\noindent\textbf{Configuration scope.}
A reproduced comparison must retain the model endpoint, harness version, task list, scorer, decoding settings, sample identifiers, and resource limits of its model--task block. The system interfaces and saved records are described in Appendix~\ref{app:alphadiana}. The sampling settings above and the runtime caps in Appendix~\ref{app:wall-time-sensitivity} define the macro operating point; Appendix~\ref{app:component-intervention-details} specifies the additional tool, skill, and memory settings for the micro study. Resource summaries must use the coverage and denominator conventions in Appendix~\ref{app:cost-measurement}.

\noindent\textbf{Compute resources.}
All AlphaDiana experiments were conducted on a local GPU cluster with 16 NVIDIA A100 GPUs, each with 80GB memory. The GPUs were used for model serving, trajectory generation, and harness-based evaluation. Since our study focuses on evaluation-time behavior rather than large-scale model training, the compute budget was mainly spent on repeated benchmark execution across models, harnesses, and sampling configurations.

%% file: draw-figures/harness-system-components.tex
\begin{tikzpicture}[
    font=\small,
    >=Latex,
    line cap=round,
    line join=round,
    flow/.style={
        -{Latex[length=1.55mm,width=1.08mm]},
        line width=0.78pt,
        draw=black!70,
        shorten <=1pt,
        shorten >=1pt
    },
    stateflow/.style={
        {Latex[length=1.35mm,width=0.95mm]}-{Latex[length=1.35mm,width=0.95mm]},
        line width=0.72pt,
        draw=black!68,
        shorten <=1pt,
        shorten >=1pt
    },
    modelcall/.style={
        {Latex[length=1.05mm,width=0.75mm]}-{Latex[length=1.05mm,width=0.75mm]},
        line width=0.72pt,
        draw=black!68,
        shorten <=1pt,
        shorten >=1pt
    },
    card/.style={
        draw=black!60,
        rounded corners=2.0mm,
        line width=0.72pt,
        align=center,
        inner sep=3.8pt,
        fill=white
    },
    responsibility/.style={
        card,
        fill=orange!10,
        text width=4.10cm,
        minimum width=4.40cm,
        minimum height=1.28cm
    },
    modelbox/.style={
        card,
        fill=gray!8,
        text width=14.40cm,
        minimum width=14.70cm,
        minimum height=0.62cm
    }
]

\node[responsibility] (interface) at (-5.15,0) {
    \textbf{Prompt \&}\\[-1pt]
    \textbf{Action Interface}\\[-1pt]
    \footnotesize prompts, tools, skills
};

\node[responsibility] (control) at (0,0) {
    \textbf{Control Logic}\\[-1pt]
    \footnotesize context, feedback\\[-1pt]
    \footnotesize loop, recovery, stop
};

\node[responsibility] (state) at (5.15,0) {
    \textbf{Runtime State}\\[-1pt]
    \footnotesize session, trajectory,\\[-1pt]
    \footnotesize workspace, memory
};

\node[modelbox] (model) at (0,-1.55) {
    \textbf{Base Model $f_\theta$}
};

\node[align=center, text=black!76] (harnesstitle) at (0,0.98) {
    \textbf{Native Harness $\mathcal H$}
};

\node[align=center, text=black!76] (agenttitle) at (0,1.70) {
    \textbf{Agent System}
    $\mathcal A_{\theta,\mathcal H}=\mathcal H\odot f_\theta$
};

\begin{scope}[on background layer]
\node[
    draw=black!35,
    rounded corners=2.8mm,
    line width=0.78pt,
    inner xsep=3.2mm,
    inner ysep=2.8mm,
    fill=orange!1,
    fit=(agenttitle)(harnesstitle)(interface)(state)(model)
] (agentframe) {};

\node[
    draw=black!30,
    rounded corners=2.5mm,
    line width=0.72pt,
    inner xsep=1.8mm,
    inner ysep=1.9mm,
    fill=orange!4,
    fit=(harnesstitle)(interface)(state)
] (harnessframe) {};
\end{scope}

\draw[stateflow] (interface.east) -- (control.west);
\draw[stateflow] (control.east) -- (state.west);
\draw[modelcall] (interface.south) -- (interface.south |- model.north);

\end{tikzpicture}

%% file: appendix/4-further-macro-exp.tex
\section{Macro Study: Analysis Methods and Supplementary Results}
\label{app:macro}
\label{app:trajectory-annotation}

This appendix follows the order of Sec.~\ref{sec:macro}: benchmark performance, error patterns, functional-step patterns, token entropy, and token use and wall-clock time. Each analysis is accompanied by its measurement procedure and supporting evidence. Evaluation settings and resource-coverage conventions appear in Appendix~\ref{app:iclr-main-details}; component interventions appear in Appendix~\ref{app:component-intervention-details}.

\subsection{Benchmark Performance and Task Coverage}
\label{app:macro-performance}

The benchmark scores in Fig.~\ref{fig:macro-main-results} summarize outcomes. The paired comparisons identify tasks that change status; the extraction ablation tests sensitivity to post-execution answer processing.

\subsubsection{Paired Coverage and Token Use on GPQA-Diamond}
\label{app:gpqa-paired-coverage}

Task-paired accounting on Qwen3.5-27B/GPQA-Diamond separates three quantities that aggregate accuracy combines (Tab.~\ref{tab:gpqa-accounting}): tasks added beyond \emph{Direct}, \emph{Direct}-correct tasks lost by each harness, and output-token use on wrong versus correct trajectories. The four-system union solves 175 of 198 tasks (88.4\%), compared with 159 of 198 (80.3\%) for \emph{Direct}; each individual harness nevertheless loses more \emph{Direct}-correct tasks than it adds from the \emph{Direct}-wrong set.

\begin{table}[H]
\centering
\fontsize{9}{11}\selectfont
\renewcommand{\arraystretch}{1.12}
\setlength{\tabcolsep}{7pt}
\caption{\textbf{Paired coverage and token use on Qwen3.5-27B/GPQA-Diamond ($n=198$).} Added/lost count tasks that change between correct and non-correct outcomes relative to \emph{Direct}, including operational failures among non-correct outcomes; W/C is mean output tokens on wrong trajectories divided by mean output tokens on correct trajectories. Union credits a task when any of the four systems is correct.}
\label{tab:gpqa-accounting}
\vspace{-4pt}
\begin{tabular}{@{}lrrrr@{}}
\toprule
System & Accuracy (\%) & Added & Lost & W/C \\
\midrule
\emph{Direct} & 80.3 & -- & -- & $2.2\times$ \\
OpenClaw & 66.2 & 8 & 36 & $11.9\times$ \\
OpenCode & 73.2 & 10 & 24 & $3.3\times$ \\
ZeroClaw & 77.8 & 13 & 18 & $1.1\times$ \\
\midrule
Four-system union & \textbf{88.4} & 16 & -- & -- \\
\bottomrule
\end{tabular}
\end{table}

\subsubsection{Paired Outcomes on Interactive Tasks}
\label{app:interactive-outcomes}

\looseness=-1
On Qwen/Terminal-Bench 2, terminus-2 solves 35/89 tasks, OpenClaw 26/89, and ZeroClaw and OpenCode 13/89 each. Across 88 paired tasks, OpenClaw gains seven solutions and loses 16 relative to terminus-2. The Gemma four-system union solves 25/89 tasks. Exact McNemar tests ($n=86$) against terminus-2 give $p=0.021$, $0.001$, and $0.115$ for OpenCode, ZeroClaw, and OpenClaw.

\subsubsection{Answer-Extraction Ablation}
\label{app:iclr-extraction}

\looseness=-1
Tab.~\ref{tab:macro-extraction-ablation} re-scores the saved answer-based evaluations after removing the permissive extractor's final-line and final-integer fallbacks. Native execution is unchanged.

\begin{table}[ht]
\centering
\fontsize{8}{8}\selectfont
\renewcommand{\arraystretch}{1.12}
\setlength{\tabcolsep}{4pt}
\caption{\textbf{Framework-level ablation of answer extraction on the saved answer-based evaluations.} Each entry is the percentage-point change from the reported score after removing the permissive extractor's final-line and final-integer fallbacks and retaining only boxed or explicitly marked answers. Model outputs, task references, and scorers are unchanged; zero means that strict and reported extraction produce the same aggregate score. Green/red cells denote positive/negative changes under strict extraction.}
\label{tab:macro-extraction-ablation}
\begin{tabular}{@{}llcccccc@{}}
\toprule
\multirow{2}{*}{Model} & \multirow{2}{*}{System}
& IMO-AnswerBench
& HLE-Verifiable
& GPQA-Diamond
& \multicolumn{2}{c}{AIME~26}
& MMMU-Pro \\
\cmidrule(lr){6-7}
& & Avg@1 & Avg@1 & Avg@1 & Pass@4 & Avg@4 & Avg@1 \\
\midrule
\multirow{4}{*}{Qwen3.5}
& Direct   & 0 & 0 & 0 & 0 & 0 & 0 \\
& OpenClaw & \cellr{$-1.25$} & 0 & 0 & 0 & 0 & 0 \\
& ZeroClaw & \cellr{$-2.50$} & 0 & 0 & 0 & 0 & 0 \\
& OpenCode & \cellr{$-4.00$} & 0 & 0 & 0 & \cellr{$-0.83$} & 0 \\
\midrule
\multirow{4}{*}{Gemma-4}
& Direct   & 0 & 0 & 0 & 0 & 0 & 0 \\
& OpenClaw & \cellr{$-0.25$} & 0 & 0 & 0 & 0 & 0 \\
& ZeroClaw & 0 & 0 & 0 & 0 & 0 & 0 \\
& OpenCode & 0 & 0 & 0 & 0 & 0 & 0 \\
\midrule
\multirow{4}{*}{Kimi-K2.6}
& Direct   & \cellr{$-1.00$} & 0 & 0 & 0 & 0 & 0 \\
& OpenClaw & \cellr{$-1.25$} & 0 & 0 & 0 & \cellr{$-1.67$} & 0 \\
& ZeroClaw & 0 & 0 & 0 & 0 & 0 & 0 \\
& OpenCode & \cellg{$+0.25$} & 0 & 0 & 0 & \cellr{$-0.83$} & \cellr{$-0.12$} \\
\bottomrule
\end{tabular}
\end{table}

\FloatBarrier
\subsection{Error Patterns in Incorrect Attempts}
\input{appendix/error-incidence}

\FloatBarrier
\subsection{Functional-Step Annotation and Aggregation}
\label{app:functional-step-patterns}
\input{appendix/trajectory-annotation}

\FloatBarrier
\subsection{Token-Entropy Patterns}
\label{app:macro-entropy}

Sec.~\ref{sec:macro} compares token entropy after tool calls and across full trajectories. Entropy is computed from the returned top-$K$ token probabilities after renormalization and measures concentration within the recorded candidate set.

\subsubsection{Entropy and Tool-Call Depth}
\begin{figure*}[t]
    \centering
    \subfigure[OpenClaw]{
    \includegraphics[width=0.48\linewidth,]{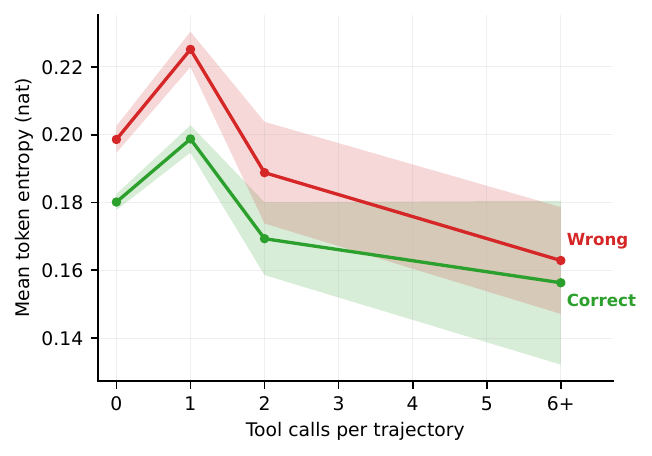}
    }
    \hfill
    \subfigure[OpenCode]{
    \includegraphics[width=0.48\linewidth,]{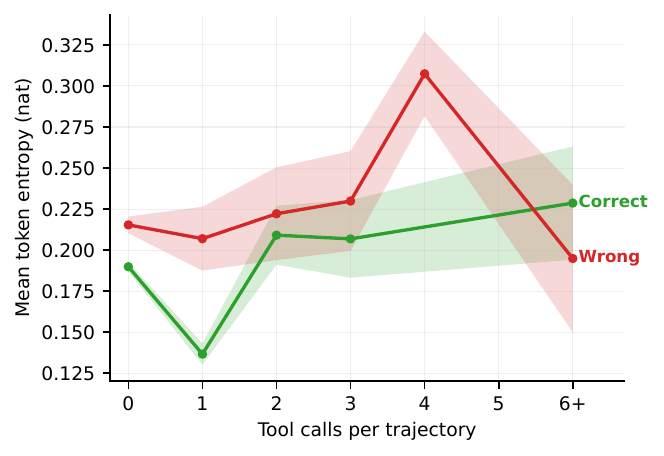}
    }
    \vspace{-4pt}
    \caption{\textbf{Mean token entropy versus tool-call depth on MMMU-Pro (Gemma-4-31B-IT).} Each panel is one tool-using harness; the x-axis is the number of tool calls per trajectory and the y-axis is mean token entropy (nat). Green: correct; red: wrong; shaded bands are confidence intervals.}
    \label{fig:mmmu-entropy-tool}
    \vspace{-8pt}
\end{figure*}

\findingbox[find:mmmu-tool-entropy][Fig.~\ref{fig:mmmu-entropy-tool}]{\looseness=-1\textbf{Entropy trends with tool-call depth depend on the harness.} Mean token entropy declines with depth under OpenClaw but peaks at intermediate depths for wrong OpenCode trajectories.}

\begingroup
\clubpenalties 3 10000 10000 0
\widowpenalties 3 10000 10000 0
Fig.~\ref{fig:mmmu-entropy-tool} compares groups of trajectories with different tool-call counts. Under OpenClaw, entropy peaks at the first tool call and then falls for both outcomes, whose curves remain close. Under OpenCode, wrong-trajectory entropy rises with tool depth and peaks near four calls; the correct curve moves above the wrong one at $6+$ calls.
\endgroup

\Needspace{6\baselineskip}
\subsubsection{Entropy and Output Length}

\begin{figure*}[t]
    \centering
    \includegraphics[width=\linewidth]{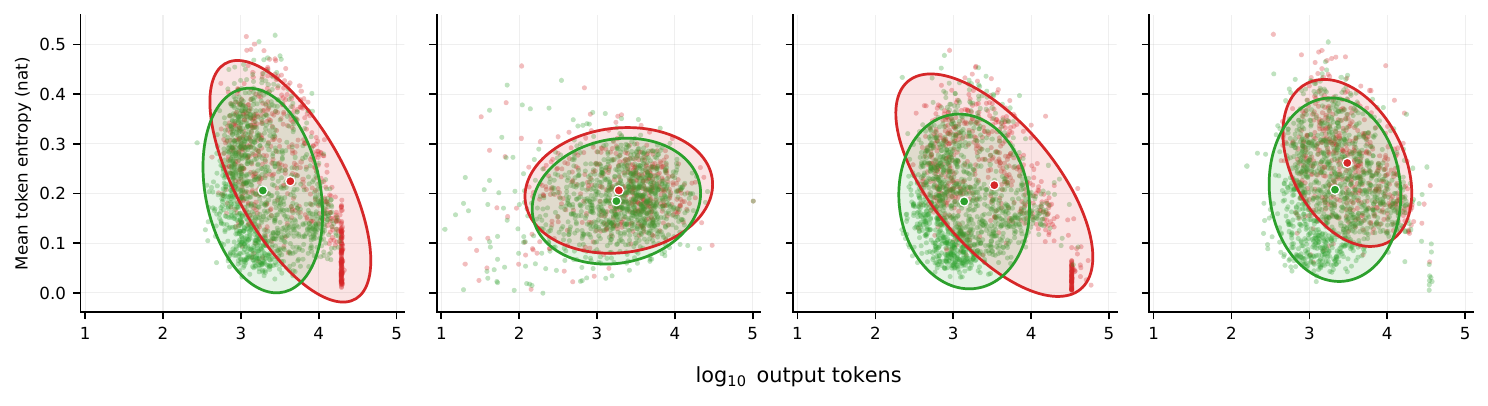}
    \vspace{-12pt}
    \caption{\textbf{Trajectory-level mean token entropy versus output length on MMMU-Pro (Gemma-4-31B-IT).} Each point is one trajectory; green: correct; red: wrong, with per-outcome covariance ellipses and centroids. Length is on a $\log_{10}$ scale. From left to right, panels show \emph{Direct}, OpenClaw, OpenCode, and ZeroClaw.}
    \label{fig:mmmu-entropy-scatter}
    \vspace{-10pt}
\end{figure*}

\findingbox[find:mmmu-confident-tail][Fig.~\ref{fig:mmmu-entropy-scatter}]{\textbf{Long-tail failure persists here, but its entropy direction depends on the harness.} Wrong trajectories form a long, low-entropy tail under \emph{Direct} and OpenCode, yet carry \emph{higher} entropy than correct ones under ZeroClaw, which shows little tool use in this analysis.}

Fig.~\ref{fig:mmmu-entropy-scatter} shows long, low-entropy incorrect trajectories under \emph{Direct} and OpenCode, consistent with Finding~\ref{find:entropy-length-collapse}. OpenClaw's correct and incorrect groups overlap substantially. Under ZeroClaw, the incorrect-trajectory centroid lies \emph{above} the correct one. Low mean top-$K$ entropy can therefore accompany incorrect answers, while the entropy--outcome association varies across harnesses.

\FloatBarrier
\subsection{Token Use and Wall-Clock Time Sensitivity}
\label{app:wall-time-sensitivity}

\looseness=-1
Appendix~\ref{app:cost-measurement} specifies output-token accounting and the observed task sets used in Fig.~\ref{fig:macro-efficiency}. The analysis below uses a separate matched-attempt intersection to examine realized time and timeout sensitivity.

The macro evaluation compares complete systems at one predeclared operating point rather than tuning a separate resource optimum for each harness. Matched conditions share the model context configuration and a per-request generation limit of \texttt{max\_tokens}$=131{,}072$. The controller or request cap is 9,300 seconds; OpenClaw additionally uses a 9,000-second stream-total cap, while ZeroClaw limits each run to 100 tool iterations and 200 actions per hour. These safeguards were fixed before evaluation. Because multi-turn systems can terminate early or issue different numbers of requests, fixed limits bound the comparison but do not equalize realized compute.

We measure wall-clock time in nine validated model--benchmark blocks spanning all three models and four answer-based benchmarks. Within each block, we intersect attempt IDs across \emph{Direct}, OpenClaw, OpenCode, and ZeroClaw, retain failed and timed-out attempts, compute each system's mean, median, and 90th-percentile wall-clock time, and then macro-average each statistic across blocks. This yields 6,848 matched attempts per system. Tab.~\ref{tab:wall-time-summary} shows substantial long-tail wall-clock times, but no monotonic relation between longer realized time and higher accuracy.

\begin{table}[H]
\centering
\fontsize{9}{11}\selectfont
\renewcommand{\arraystretch}{1.15}
\setlength{\tabcolsep}{14pt}
\caption{\textbf{Macro-averaged realized wall-clock time over matched attempts.} Statistics are computed separately in nine model--benchmark blocks and then macro-averaged. Each system contributes 6,848 matched attempts, including failures and timeouts.}
\label{tab:wall-time-summary}
\vspace{-4pt}
\begin{tabular}{@{}lrrr@{}}
\toprule
System & Mean (s) & Median (s) & p90 (s) \\
\midrule
\emph{Direct} & 682.7 & 429.2 & 1,124.2 \\
OpenClaw & 576.8 & 265.4 & 1,635.4 \\
OpenCode & 530.9 & 217.1 & 1,450.2 \\
ZeroClaw & 513.3 & 259.3 & 1,189.5 \\
\bottomrule
\end{tabular}
\end{table}

We next re-score the same attempts at 17 lower wall-clock time caps up to 9,300 seconds, counting an otherwise correct attempt as unsuccessful when it completes after a cap. In 35 of 36 model--system--benchmark cells, some smaller cap retains at least 99\% of the original accuracy. The exception is Qwen3.5-27B/GPQA-Diamond/OpenCode, which remains right-censored at 9,300 seconds. In each of the eight blocks with at least two cap values above the 99\% accuracy-retention threshold for all four systems, the ranking remains unchanged across those values.

This analysis measures ranking sensitivity to the timeout policy for the recorded runs. Wall-clock time includes model-serving latency, harness overhead, tool execution, and early termination.

%% file: appendix/error-incidence.tex
\label{app:error-incidence}

\noindent\textbf{Analysis sample.} Fig.~\ref{fig:failure-taxonomy} summarizes 2,857 recorded incorrect attempts: 1,108 from Gemma, 839 from Kimi, and 910 from Qwen. All 48 model--benchmark--system combinations are represented, with at least 30 attempts in 40 combinations. Sampling supplements sparse combinations independently of their error labels. Correct and unscored attempts, empty or corrupted records, and known scoring or task-matching defects are excluded. Rates describe this sample, not full-archive prevalence.

\noindent\textbf{Annotation procedure.} GPT-5.6-sol with high reasoning effort assigns binary error labels in a single annotation stage, using the problem, saved trajectory and output, reference answer, and recorded grading result. Positive labels require supporting source references; accepted annotations pass schema and source-reference checks. Incomplete annotations are excluded from the analysis. For 255 MMMU-Pro annotation packets, benchmark images are attached from a pinned snapshot after matching task IDs, ordered options, and reference answers.

\noindent\textbf{Error codes and counting.} We report 17 labels in five families: Specification and State Design (SD), Reasoning and Control (RC), Tool and Evidence Interaction (TE), Verification and Finalization (VF), and Resource and Operational Reliability (OR). Tabs.~\ref{tab:error-definitions-reasoning}--\ref{tab:error-definitions-finalization} give each label's definition, required evidence, and decision boundaries. Unlike the six functional step types in Tab.~\ref{tab:functional-step-definitions}, these labels identify errors supported by the recorded evidence. Each code receives \emph{YES} for a supported incident and \emph{NO} otherwise. Multiple labels may apply to one attempt; they are not mutually exclusive stages.

For group $g$ and code $c$, incidence is $100\,n_{g,c}^{\mathrm{yes}}/N_g^{\mathrm{incorrect}}$. Each attempt counts at most once per code, including an error later recovered within that incorrect attempt. Codes overlap, so percentages need not sum to 100\%. Fig.~\ref{fig:failure-taxonomy} pools attempts across the three models within each benchmark--system group. Repetition incidence follows the RC-4 definition: repeated reasoning, queries, or actions without new evidence, meaningful input changes, or progress. Each positive attempt contributes one count regardless of cycle frequency or duration. Budget exhaustion is assessed separately from termination evidence.

\noindent\textbf{Repetition across models.} On HLE-Verifiable, OpenCode has a lower repetition rate than ZeroClaw within each model: 19.0\% versus 33.3\% for Gemma (15/79 versus 31/93), 16.2\% versus 24.4\% for Kimi (12/74 versus 19/78), and 66.0\% versus 76.6\% for Qwen (66/100 versus 36/47). Combining the models gives 93/253 (36.8\%) for OpenCode and 86/218 (39.4\%) for ZeroClaw. Different model proportions narrow the combined difference, which is smaller than the difference within any of the three models.

\noindent\textbf{Sensitivity checks.} Adding 49 annotations without source references gives 2,906 attempts and changes every overall code rate by less than 0.54 percentage points. Excluding the 255 image-supplemented attempts gives RC-1/RC-4/VF-2 rates of 82.9\%/\allowbreak 71.0\%/\allowbreak 59.5\%, compared with 83.2\%/\allowbreak 67.1\%/\allowbreak 57.5\% in the main analysis; this comparison also removes four model--benchmark--system combinations. Equal weighting of all 48 combinations gives 82.4\%/\allowbreak 63.1\%/\allowbreak 53.0\%. The ordering of these three leading labels is unchanged under both checks. Execution-level labels, source hashes, group counts, and plotting code accompany the analysis.

\input{appendix/error-code-definitions}

\input{appendix/error-human-alignment}

%% file: appendix/error-code-definitions.tex
\paragraph{Definitions and decision boundaries.}
\label{app:error-definitions}
Tabs.~\ref{tab:error-definitions-reasoning}--\ref{tab:error-definitions-finalization} define the 17 labels in Fig.~\ref{fig:failure-taxonomy}. Codes identify labels in the saved annotations; the figure uses readable names. Family names organize observable error indicators, not root-cause attributions. A positive judgment requires the stated evidence. An incorrect final answer alone is insufficient to infer a particular process error, and missing evidence is not proof that the behavior was absent.

\begingroup
\small
\renewcommand{\arraystretch}{1.12}
\setlength{\tabcolsep}{4pt}
\setlength{\LTleft}{0pt}
\setlength{\LTright}{0pt plus 1fill}
\begin{longtable}{@{}p{0.27\linewidth}p{\dimexpr0.73\linewidth-8pt\relax}@{}}
\caption{\textbf{Error definitions: specification and reasoning.} Row names match Fig.~\ref{fig:failure-taxonomy}.}\label{tab:error-definitions-reasoning}\\
\toprule
Code and label & Definition and required evidence \\
\midrule
\endfirsthead
\multicolumn{2}{@{}l}{\small\tablename~\thetable{} (continued)} \\
\toprule
Code and label & Definition and required evidence \\
\midrule
\endhead
\bottomrule
\endfoot
\endlastfoot
\multicolumn{2}{@{}l}{\textbf{Specification and State Design (SD)}} \\*
SD-1: Task contract & Violates an explicit task requirement or output contract. Identify the requirement and conflicting action or output; a wrong answer or reasoning error alone does not qualify. \\
SD-2: Capability / interface & The exposed or withheld capability is materially mismatched to the task or model-facing contract. Require the actual configuration and necessary capability; absence claims require a complete relevant inventory and no equivalent route. Choosing not to use an available suitable interface is not this error. \\
\midrule
\multicolumn{2}{@{}l}{\textbf{Reasoning and Control (RC)}} \\*
RC-1: Reasoning error & A visible factual, mathematical, logical, perceptual, or inferential step conflicts with task evidence or a verified reference. Localize the incorrect proposition or inference; a wrong final answer without an evidenced erroneous step is insufficient. Later correction does not erase the incident. \\
RC-2: Decision--action inconsistency & A stated decision, plan, or conclusion conflicts with a later action or submitted answer. Identify both events and establish that no intervening explicit revision explains the change. \\
RC-3: Reasoning drift & Shifts to an unrelated objective or sustained off-goal activity without a justified path to task progress. Show the task goal and the departure; relevant exploration that is later abandoned does not qualify. \\
RC-4: Repetitive cycles & Repeats substantially the same reasoning, query, or action without new evidence, meaningful input changes, or progress. Compare at least two cycles. Justified retries, appropriate polling, and independent verification are excluded. \\
RC-5: Premature commitment & Commits or stops while an essential obligation remains unresolved. Show the agent-controlled decision and the obligation; omission claims require the complete relevant window. Forced budget/runtime stops and a wrong answer alone do not qualify. \\
\bottomrule
\end{longtable}
\endgroup

\begingroup
\small
\renewcommand{\arraystretch}{1.12}
\setlength{\tabcolsep}{4pt}
\setlength{\LTleft}{0pt}
\setlength{\LTright}{0pt plus 1fill}
\begin{longtable}{@{}p{0.27\linewidth}p{\dimexpr0.73\linewidth-8pt\relax}@{}}
\caption{\textbf{Error definitions: Tool and Evidence Interaction (TE).}}\label{tab:error-definitions-tools}\\
\toprule
Code and label & Definition and required evidence \\
\midrule
\endfirsthead
\multicolumn{2}{@{}l}{\small\tablename~\thetable{} (continued)} \\
\toprule
Code and label & Definition and required evidence \\
\midrule
\endhead
\bottomrule
\endfoot
\endlastfoot
TE-1: Tool selection / omission & Selects a materially unsuitable operation, or omits an explicitly required, available tool. Show the mismatch; omission requires the tool obligation, complete inventory and observation window, and no equivalent action. A permitted direct solution is not an omission. \\
TE-2: Tool execution & A concrete invocation fails through malformed arguments, unavailable capability, permission denial, timeout, process failure, or environment incompatibility. Require a failed call/result or terminal error state; pending asynchronous work is not failure. \\
TE-3: Observation integration & Available, relevant tool or environment feedback is ignored, misread, contradicted, or not incorporated. Identify the observation and incompatible later reasoning/action. Correctly integrating a defective source is a different issue. \\
TE-4: Evidence quality & Relies without adequate qualification on external evidence with a demonstrable support, relevance, or provenance defect. Identify the source defect and its use in the conclusion; internal knowledge errors or a source's status alone do not qualify. \\
TE-5: Recovery & After recognized failure feedback, a recovery attempt does not address the precursor condition and leaves it unresolved. Identify the feedback, attempted remedy, and unresolved result. No recovery attempt, a successful remedy, or a justified alternative failing for a new reason does not qualify. \\
\bottomrule
\end{longtable}
\endgroup

\begingroup
\small
\renewcommand{\arraystretch}{1.12}
\setlength{\tabcolsep}{4pt}
\setlength{\LTleft}{0pt}
\setlength{\LTright}{0pt plus 1fill}
\begin{longtable}{@{}p{0.27\linewidth}p{\dimexpr0.73\linewidth-8pt\relax}@{}}
\caption{\textbf{Error definitions: verification, finalization, and operational reliability.}}\label{tab:error-definitions-finalization}\\
\toprule
Code and label & Definition and required evidence \\
\midrule
\endfirsthead
\multicolumn{2}{@{}l}{\small\tablename~\thetable{} (continued)} \\
\toprule
Code and label & Definition and required evidence \\
\midrule
\endhead
\bottomrule
\endfoot
\endlastfoot
\multicolumn{2}{@{}l}{\textbf{Verification and Finalization (VF)}} \\*
VF-1: Verification omission & Submits a checkable candidate without a necessary check, or omits a decisive condition. Require a complete relevant trace, a feasible check, and an explicit obligation, unresolved expressed uncertainty/contradiction, or prespecified checklist item. Merely not showing extra checking is insufficient. \\
VF-2: Faulty verification & A check of an existing candidate uses a demonstrably faulty method or verdict. Identify the candidate and subsequent checking step; distinguish checking from first-pass reasoning and from an absent check. Later repair does not erase the incident. \\
VF-3: Answer commit / format & An identifiable intended answer is not committed, formatted, preserved, extracted, or transmitted according to the scorer-facing contract. Localize the boundary defect independently of semantic correctness; a faithfully delivered wrong answer or missing output alone is insufficient. \\
\midrule
\multicolumn{2}{@{}l}{\textbf{Resource and Operational Reliability (OR)}} \\*
OR-1: Budget exhaustion & A run-level token, context, time, step, or cost limit stops or truncates required processing. Require an explicit limit signal, finish reason, timeout, or limit-reaching metric. Length alone and a timeout confined to one tool call do not qualify. \\
OR-2: Runtime disruption & A harness, gateway, sandbox, provider, process, artifact, or infrastructure fault disrupts execution or scoring. Require a concrete run-level status or fault record; an isolated tool error handled within a healthy run is not sufficient. \\
\bottomrule
\end{longtable}
\endgroup

\noindent\textbf{Optional reasoning-error subtypes.} RC-1 may additionally identify task interpretation (misreading the goal or conditions), perception/reading (misreading supplied visual or textual content), knowledge/concepts (using an incorrect fact or domain rule), numerical/symbolic calculation (arithmetic, algebra, units, or symbol manipulation), or logic/modeling (invalid assumptions, formalization, inference, or proof). If the evidence does not support a reliable subtype, retain RC-1 alone. These subtypes do not add independent counts to Fig.~\ref{fig:failure-taxonomy}.

\noindent\textbf{Overlapping incidents.} RC-1 concerns an erroneous reasoning step, TE-3 concerns integrating an available observation, and VF-2 concerns checking an existing candidate. They may coexist when the record supports each judgment; do not assign several labels merely to redescribe the same error. Similarly, premature commitment requires evidence beyond an already identified tool or verification omission. Label the supported behavior and execution level rather than inferring a hidden cause.

%% file: appendix/error-human-alignment.tex
\subsubsection{Human Alignment of Error Labels}
\label{app:error-human-alignment}

\noindent\textbf{Study and sample.} Two human annotators separately labeled saved trajectories using the error taxonomy. We report agreement between the single-pass GPT-5.6-sol baseline and these annotators using the 17 parent codes. The core sample comprises 60 trajectories selected by fixed-hash ranking within model--benchmark--harness strata from 4,566 readable executions recorded as failures. Five additional tool-error cases form a separate targeted supplement.

Screening the labeled trajectories for task validity and outcome status retains 42 of the 60 core trajectories, covering 40 problems. Exclusions comprise six correct or equivalent answers under documented task assumptions, one task--trajectory mismatch, two incomplete tasks, and nine unresolved outcomes.

\noindent\textbf{Comparison.} Comparisons use available paired judgments on the same trajectory and code to assess the presence or absence of a positive error assignment. The primary comparison contains 714 paired decisions per rater pair. Overall agreement and Cohen's $\kappa$ are computed from pooled binary decisions, with each trajectory and code equally weighted. LLM--human results average the metrics computed against each of the two human references.

\begin{table}[H]
\centering
\small
\caption{\textbf{Overall human alignment of error annotations.} Parentheses give 95\% problem-cluster bootstrap intervals. LLM--human denotes the mean over two human references.}
\label{tab:error-human-alignment}
\begin{tabular}{@{}lcc@{}}
\toprule
Comparison & Agreement (\%) & Cohen's $\kappa$ \\
\midrule
Human--human & 86.8 (84.5--89.2) & 0.472 (0.393--0.554) \\
LLM--human & 88.0 (86.0--89.9) & 0.538 (0.470--0.602) \\
\bottomrule
\end{tabular}
\end{table}

The LLM's aggregate agreement is comparable to the observed inter-annotator agreement (Tab.~\ref{tab:error-human-alignment}). Reference-specific agreement is 84.5\% with H1 and 91.5\% with H2. Because negative assignments are frequent, Tab.~\ref{tab:error-human-positive} also reports positive-label precision, recall, and F1 with each human as the reference. Mean positive F1 is 0.608, compared with 0.548 between humans.

\begin{table}[H]
\centering
\small
\caption{\textbf{Positive-error agreement by human reference.} H1 and H2 are anonymized annotators. Parentheses give 95\% problem-cluster bootstrap intervals.}
\label{tab:error-human-positive}
\begin{tabular}{@{}lccc@{}}
\toprule
Reference & Precision & Recall & Positive F1 \\
\midrule
H1 & 0.417 (0.327--0.509) & 0.549 (0.451--0.644) & 0.474 (0.385--0.558) \\
H2 & 0.733 (0.650--0.819) & 0.752 (0.686--0.815) & 0.743 (0.684--0.798) \\
\bottomrule
\end{tabular}
\end{table}

\noindent\textbf{Uncertainty and sensitivity.} We use 10,000 percentile bootstrap resamples (seed 20260913), sampling underlying problems within benchmarks and keeping associated attempts, codes, and raters together. Both LLM--human comparisons use the same resamples before averaging. Resampling preserves dependence among decisions for the same problem. Restricting the analysis to 347 decisions passing the common-evidence screen gives 86.7\% LLM--human and 85.0\% human--human agreement. Including the nine unresolved outcomes gives 87.3\% and 86.2\%, respectively, without treating those outcomes as resolved.

\noindent\textbf{Scope.} The analysis reports agreement on the selected trajectories using the 17 parent codes.

%% file: appendix/trajectory-annotation.tex
\label{app:action-labeling}

Fig.~\ref{fig:count-abs} compares 2,751 scored trajectories across two models, three benchmarks, four systems, and two outcomes (48 cells). Tab.~\ref{tab:action-batch-coverage} reports coverage. Eligible trajectories have substantive saved content and resolvable event order. Supplementation uses a fixed, label-independent rank to improve sparse-cell coverage, while retaining existing larger cells. Correct and incorrect groups follow the recorded scorer; unscored attempts are excluded.

\begin{table}[H]
\centering
\fontsize{9}{11}\selectfont
\renewcommand{\arraystretch}{1.12}
\setlength{\tabcolsep}{8pt}
\caption{\textbf{Trajectory coverage for Fig.~\ref{fig:count-abs}.} Entries give correct/incorrect counts pooled over Direct, OpenClaw, ZeroClaw, and OpenCode.}
\label{tab:action-batch-coverage}
\begin{tabular}{@{}lrrrr@{}}
\toprule
Model & GPQA-Diamond & HLE-Verifiable & IMO-AnswerBench & Total \\
\midrule
Gemma-4-31B-IT & 299/62 & 295/359 & 303/300 & 1,618 \\
Kimi-K2.6 & 219/46 & 185/256 & 219/208 & 1,133 \\
\midrule
Total & 518/108 & 480/615 & 522/508 & 2,751 \\
\bottomrule
\end{tabular}
\end{table}

\paragraph{Step construction and labeling.}
A \emph{functional step} is one saved assistant-text segment or one logged tool call. An adapter removes duplicate storage copies while preserving source-linked assistant text, saved reasoning, and logged tool calls in order. Text is split into natural paragraphs or list steps; each actual call is a separate target. System, user, and tool-result events provide context rather than additional targets. Empty-text ZeroClaw turns require consistent call records and linked results. Only recorded reasoning is included.

\texttt{GPT-5.6-sol} at high reasoning effort assigns one label per fixed step in a single classification stage, using the task and saved textual context (Tab.~\ref{tab:functional-step-definitions}). The classifier receives text only. Long inputs are partitioned at step boundaries or use a long-context deployment with the same prompt. Accepted outputs cover every assigned step exactly once and pass source-linked structural checks; retries do not add observations.

\begin{table}[H]
\centering
\fontsize{9}{11}\selectfont
\renewcommand{\arraystretch}{1.12}
\setlength{\tabcolsep}{5pt}
\caption{\looseness=-1\textbf{Functional-step labels.} Labels describe step function independently of content correctness.}
\label{tab:functional-step-definitions}
\begin{tabularx}{\linewidth}{@{}lX@{}}
\toprule
Label & Definition and boundary \\
\midrule
Understanding & Identifies or restates the task's goals, givens, constraints, or answer form. \\
Planning & Proposes a strategy or next operation without carrying it out. \\
Reasoning & Derives, calculates, applies knowledge, or develops a candidate solution. \\
Tool Use & Issues a logged call or directly incorporates returned information, including a failure; a proposed call alone is Planning. \\
Verification & Checks or revises an existing result or interpretation; first-time derivation is Reasoning, and an unsuccessful check still qualifies. \\
Finalization & Commits to a task-level answer or terminal inability to answer; an intermediate value or boxed expression alone is insufficient. \\
\bottomrule
\end{tabularx}
\end{table}

For genuinely mixed steps, the tie-break is Finalization $>$ Tool Use $>$ Verification $>$ Reasoning $>$ Planning $>$ Understanding, applied only to functions supported by the context. Actual tool-call targets always receive Tool Use, including denied calls. Each step receives exactly one functional label; the labels are categories, not a prescribed stage sequence. A Verification step may perform either a valid or an invalid check, and Tool Use does not imply successful execution. These functional labels therefore differ from the multi-label error judgments in Appendix~\ref{app:error-incidence}.

\paragraph{Aggregation and weighting.}
Panel~(a) pools labeled-step counts within each model--benchmark--system--outcome cell and divides by that cell's total. Panel~(b) counts adjacent labels within HLE trajectories, including self-transitions but never cross-trajectory edges, and normalizes each panel by its transition total. These summaries describe pooled recorded activity, giving longer trajectories greater weight. Under equal-trajectory weighting, Kimi/ZeroClaw's incorrect HLE Verification share is 3.4\%, compared with 26.7\% when pooling steps; one trajectory contributes 211 of 217 Verification steps. This comparison distinguishes concentrated checking within a long trajectory from a pattern shared across trajectories.

\input{appendix/action-human-alignment}

%% file: appendix/action-human-alignment.tex
\subsubsection{Human Alignment of Functional-Step Labels}
\label{app:action-human-alignment}

\noindent\textbf{Protocol.} Three human annotators were assigned the same 120 fixed step spans and the six definitions in Tab.~\ref{tab:functional-step-definitions}. Each span was shown with surrounding context for classification. For 60 spans, machine labels were hidden and annotators selected a category independently. For the other 60, a machine suggestion was visible and annotators could accept it or choose another category. We analyze these phases separately. \mbox{GPT-5.6-sol} with high reasoning effort classified each fixed span in a single call using the corresponding textual context, without receiving human labels. All 120 machine outputs were valid.

\noindent\textbf{Primary comparison.} The machine-hidden phase contains 176 human judgments: two annotators completed all 60 spans and one completed 56. Comparisons use available judgments. A human majority requires at least two matching votes and strictly more than half of the available votes. This yields a reference for 55 spans. Pairwise analyses also include the five spans without a majority. The classifier agrees with the majority on 50 of 55 spans (90.9\%; $\kappa=0.868$).

\begin{table}[H]
\centering
\small
\caption{\textbf{Human alignment on fixed step spans with machine labels hidden.} Pairwise rows average the three rater-pair metrics. The majority-reference row uses only spans with a human majority; its reference and denominator differ from those of the pairwise rows.}
\label{tab:action-human-alignment}
\begin{tabular}{@{}lcc@{}}
\toprule
Comparison & Agreement (\%) & Cohen's $\kappa$ \\
\midrule
Human--human (pairwise mean) & 63.2 & 0.470 \\
LLM--human (pairwise mean) & 76.0 & 0.644 \\
LLM--human majority & 90.9 & 0.868 \\
\bottomrule
\end{tabular}
\end{table}

Agreement with the three individual annotators is 67.9\%, 83.3\%, and 76.7\%, respectively. Restricting all comparisons to the 56 spans completed by all three annotators gives 75.0\% mean LLM--human agreement and 62.5\% mean human--human agreement. The pairwise pattern persists on this complete subset. The five disagreements with the majority concern boundaries among Reasoning, Planning, Verification, and Finalization.

\noindent\textbf{Machine-assisted comparison.} The machine-visible phase contains 174 human judgments. Mean LLM--human agreement is 67.2\%, versus 67.7\% between humans; majority-reference agreement is 73.2\% on 56 spans. This phase uses a separate set of fixed step spans and is summarized separately from the machine-hidden phase.

The annotation instructions were calibrated using this audit. These results summarize six-way classification of fixed step spans in the calibration sample.

%% file: appendix/5-component-intervention-details.tex
\section{Component Intervention Protocols}
\label{app:component-intervention-details}

This appendix details the matched configurations, tool removal, skill loading, memory persistence, and full-harness output-token comparisons in Sec.~\ref{sec:micro}. Appendix~\ref{app:iclr-main-details} gives shared evaluation settings.

\Needspace{22\baselineskip}
\subsection{Component Intervention Matrix}
\label{app:iclr-intervention-matrix}

\begin{table}[H]
\centering
\fontsize{8}{10}\selectfont
\renewcommand{\arraystretch}{1.3}
\setlength{\tabcolsep}{12pt}
\caption{\textbf{Configurations for tool, skill, and memory interventions.} \ding{51}/\ding{55} indicate whether the named interface or intervention is active; shading identifies the component targeted relative to \emph{Default Harness}. Each intervention includes the associated interface and prompt changes specified below. The memory intervention combines an explicit memory-use instruction with the persistence policy in Appendix~\ref{app:memory-mechanisms}. \emph{Default Harness} retains its native memory interface but clears memory between attempts and does not explicitly elicit memory use. Within each matched comparison, the agent loop, model, benchmark, scorer, decoding parameters, and sampling budget are fixed.}
\label{tab:micro-component-interventions}
\begin{tabular*}{0.82\linewidth}{@{\extracolsep{\fill}}lccc}
\toprule
Configuration & Tool registry & Skill library & Memory intervention \\
\midrule
\emph{Default Harness} & \ding{51} & \ding{55} & \ding{55} \\
\midrule
\emph{Tool Registry Removed}        & \cellcolor{black!8}\ding{55} & \ding{55} & \ding{55} \\
\emph{Skill Loaded}        & \ding{51} & \cellcolor{black!8}\ding{51} & \ding{55} \\
\emph{Memory-Elicited} & \ding{51} & \ding{55} & \cellcolor{black!8}\ding{51} \\
\bottomrule
\end{tabular*}
\end{table}

\subsection{Tool Registry Removal}
\label{app:tool-skill-contents}

Tab.~\ref{tab:component-intervention-contents} specifies the native tool registries removed by the tool-registry intervention and the procedural libraries mounted by the skill-loading intervention. These contents are fixed within each matched intervention in Sec.~\ref{sec:micro}. Tool removal suppresses the selected tool descriptions and related prompt instructions while retaining the matched model, task, scorer, decoding, budget, and agent loop. The skill-library rows describe the separate loading intervention in Appendix~\ref{app:skill-loading}.

\begin{table}[H]
\centering
\fontsize{9}{11}\selectfont
\renewcommand{\arraystretch}{1.2}
\setlength{\tabcolsep}{8pt}
\caption{\textbf{Tool surfaces and procedural libraries used in the component interventions.}}
\label{tab:component-intervention-contents}
\vspace{-4pt}
\begin{tabularx}{0.96\linewidth}{@{}lY@{}}
\toprule
Component & Contents \\
\midrule
OpenCode v1.3.2
& Shell, file operations and search, task execution, web retrieval, task lists, and skill loading \\
ZeroClaw v0.6.9
& Shell, file operations and search, tasks, web, scheduling and memory, configuration, version control, calculation, weather, and canvas \\
Math library
& A seven-step quantitative protocol with references for major mathematical domains \\
General library
& 17 skills covering documents, UI, design, communication, APIs, and web development and testing \\
\bottomrule
\end{tabularx}
\end{table}

\subsection{Skill Loading}
\label{app:skill-loading}

Loading adds a pointer to a library mounted through the harness's native interface; native tools remain available. OpenCode uses the same loading-instruction wording for both libraries; ZeroClaw requires Math loading but makes General optional. Fig.~\ref{fig:micro-skill} compares these complete loading configurations. Fig.~\ref{fig:skill-use-rate} reports the fraction of trajectories opening the top-level \texttt{SKILL.md}.

\subsection{Memory Mechanisms and Persistence}
\label{app:memory-mechanisms}

The memory study uses Qwen3.5-27B on AIME~26 with fixed model weights. OpenCode includes prior solutions in the prompt for later attempts. OpenClaw and ZeroClaw store summaries and retrieve notes through semantic search. Raw interaction histories and workspace files are reset between attempts. Only the designated memory is retained, with no separate answer cache.

\Needspace{5\baselineskip}
The three settings differ in how long this memory is kept:
\begin{itemize}[leftmargin=*, itemsep=2pt, topsep=2pt]
\item \textbf{Cross-Rollout.} Four attempts on a task share writable memory, cleared before the next task. These attempts are dependent because later ones can use earlier solutions.
\item \textbf{Cross-Task.} Four separate runs each process Tasks 1--30 in the same order. Each run starts with empty memory and keeps it across tasks. The four runs do not share memory.
\item \textbf{Cross-Dataset.} Memory is built on AIME~25 and frozen before AIME~26. Attempts on AIME~26 read this memory without updating it.
\end{itemize}

Across these three settings, the task prompt, memory instruction, limits, resources, and sampling are matched. \emph{Default Harness} does not explicitly instruct the agent to use memory and clears memory between attempts. Comparing it with \emph{Memory-Elicited} therefore measures the combined effect of memory instructions and persistence. Comparisons across harnesses cannot isolate summary quality, retrieval, prompt length, or control logic as the cause of the score differences.

\subsection{Output-Token Use Across Full Harness Configurations}
\label{app:micro-cost-protocol}

The final output-token comparisons in Sec.~\ref{sec:micro} compare full harness configurations rather than isolating an individual component. Figs.~\ref{fig:micro-solution-cost} and~\ref{fig:micro-token-allocation} use the common observed Gemma task sets and output-token accounting specified in Appendix~\ref{app:cost-measurement}. Output tokens per correct answer include tokens spent on unsuccessful attempts; the high-token and unsuccessful-attempt shares overlap. These comparisons therefore describe the output tokens used to obtain correct answers under each complete configuration, not the causal overhead of a particular tool, skill, or memory mechanism.

%% file: appendix/2-related-work.tex
\section{Related Work}
\label{app:related work}

This work studies \emph{system-level} evaluation of agentic reasoning. 
Our starting point is that an open agent is best viewed not as a model alone, but as a \emph{model-harness system}: the \emph{model} provides raw generative and reasoning capacity, while the \emph{harness} determines how that capacity is realized through control flow, tool access, context management, memory, interaction protocols, runtime policies, and feedback handling. 
This perspective connects several research threads: reasoning-capable foundation models, agent runtimes and tool-use frameworks, memory and context-management mechanisms, benchmark families ranging from model-centric reasoning tasks to interactive agent environments, and evaluation protocols that emphasize trajectories, robustness, and efficiency. 
It also clarifies a key distinction. 
\emph{Model reasoning} studies a standalone model under direct inference, whereas \emph{agentic reasoning} studies a model embedded in a multi-step loop that can act, observe, revise, and continue across turns. 
Once reasoning is coupled with action, environment interaction, and persistent state, the object of evaluation is no longer just the model, but the full system that turns the model into an agent. 
We therefore organize related work around the transition from \emph{model-only} evaluation to \emph{model-harness} evaluation. The system interfaces in Appendix~\ref{app:alphadiana}, macro analyses in Appendix~\ref{app:macro}, and component protocols in Appendix~\ref{app:component-intervention-details} provide the concrete setting for this comparison.

\subsection{Reasoning Foundation Models}

\paragraph{What this line of work evaluates.}
A large body of work evaluates the reasoning ability of language models under direct or lightly structured inference.
Chain-of-thought prompting showed that eliciting intermediate reasoning steps can substantially improve complex reasoning in large models~\citep{wei2022chain}, while benchmarks such as MATH and GSM8K became standard tests of mathematical reasoning~\citep{hendrycks2021math,cobbe2021gsm8k}. 
Broader evaluation efforts such as HELM emphasized the importance of standardized, multi-metric, and scenario-diverse assessment of language models~\citep{liang2022helm}, while challenging reasoning-focused suites such as BIG-Bench Hard highlighted persistent gaps in compositional and symbolic reasoning even for strong language models~\citep{suzgun2022bbh}. 
Subsequent work explored inference-time search and aggregation strategies, including self-consistency and tree-based reasoning, showing that measured reasoning quality depends not only on the base model but also on the inference procedure used to query it~\citep{wang2023selfconsistency,yao2023tree}. 
Open-weight reasoning models such as Qwen3 and DeepSeek-R1 further demonstrate that targeted post-training and reinforcement learning can substantially improve reasoning performance beyond standard instruction tuning~\citep{yang2025qwen3,guo2025deepseekr1}. 
In parallel, API-based reasoning models expose explicit deliberation or ``thinking'' modes and have become common baselines for complex reasoning and agentic workflows~\citep{openai2024learningtoreason,openai2026reasoningguide,google2024geminithinking}.

\paragraph{Why this is still insufficient for harness-aware evaluation.}
These works are primarily evaluated at the \emph{model level}, usually under direct prompting, fixed sampling budgets, and standardized inputs.
That setting is appropriate when the target of study is the standalone model.
However, in realistic use, strong reasoning increasingly depends on how a model is embedded in a runtime with tools, memory, observations, retries, and execution policies.
As a result, this literature tells us a great deal about \emph{raw model reasoning}, but much less about how reasoning changes once the model is operationalized by a harness.
This creates a gap between \emph{model reasoning} as it is usually benchmarked and \emph{agentic reasoning} as it is actually deployed.

\subsection{Agentic Reasoning Systems and Open Agents}

\paragraph{What this line of work evaluates.}
Modern reasoning systems increasingly combine a foundation model with planning, tool use, memory, and environment interaction, producing long-horizon trajectories rather than single completions.
Foundational examples include ReAct~\citep{yao2023react}, which interleaves reasoning and acting, LATS~\citep{zhou2024language}, which performs search over action trajectories, and Toolformer, which trains models to invoke tools in a self-supervised manner~\citep{schick2023toolformer}. 
More broadly, recent work on externalized reasoning emphasizes that capabilities often emerge from structured interaction between a model and external resources rather than from the parametric model alone~\citep{zhou2026externalization}.

Open agent systems vary widely in how they realize this interaction.
\textbf{Gateway- and runtime-centered} systems treat sessions, routing, and persistent state as first-class concerns; for example, OpenClaw embeds a model in a sessionful gateway with tool interfaces, memory, and policy controls. 
\textbf{Tool-oriented} frameworks emphasize modular tool abstractions, planning, and execution, aiming to improve reasoning through better external affordances rather than extra model training~\citep{lu2025octotools}. 
\textbf{Domain- and workflow-oriented} systems, especially in scientific settings, prioritize execution grounding and traceability~\citep{zhang2025scimaster}. 
AlphaApollo similarly orchestrates multiple models and professional tools through explicit iteration and structured tool use~\citep{zhou2025alphaapollo}. 
Recent surveys synthesize this broader shift from model-centric reasoning toward autonomous, lifelong, and self-improving agent systems~\citep{zheng2025lifelongagents,ferrag2025comprehensivereview}.

\paragraph{Why this is still insufficient for harness-aware evaluation.}
This literature establishes that many agent capabilities arise from the surrounding runtime rather than from the base model alone.
But it typically studies complete systems, methods, or design patterns, rather than isolating the harness as an explicit object of evaluation.
As a result, two agents built on the same model may behave very differently because of differences in orchestration, tool routing, state management, or feedback handling, yet those differences are often reported only as properties of the overall system.
AlphaDiana addresses this need with a measurement system that compares complete model--harness systems under matched evaluation policies and tests tool, skill, and memory configurations.

\paragraph{Memory and context management.}
A central source of variation across agent systems is how they manage context windows and persistent state.
Early work on long-context efficiency focused on compression and memory virtualization, for example LongLLMLingua for prompt compression~\citep{jiang2024longllmlingua} and MemGPT for OS-inspired hierarchical memory management~\citep{packer2023memgpt}. 
Subsequent work introduced retrieval- and structure-based long-term memory, including HippoRAG~\citep{gutierrez2024hipporag}, temporal knowledge-graph memory in Zep~\citep{rasmussen2025zep}, object-centric persistent memory~\citep{zahn2026knowledgeobjects}, graph-native versioned memory architectures~\citep{park2026kumiho}, provenance-aware verified memory~\citep{zhu2026tiermem}, and governance-oriented memory control~\citep{kumar2026memarchitect}. 
A second line of work studies working-memory management for long-horizon agents, including hierarchical working memory in HiAgent~\citep{hu2025hiagent}, agentic memory modules such as A-MEM and Mem0~\citep{xu2025amem,chhikara2025mem0}, adaptive memory distillation and memory-structure selection~\citep{ma2025whatdeservesmemory,lu2026choosingmemory}, query-aware budget-tier routing~\citep{zhang2026budgetmem}, and compression or pruning methods such as ACON, Active Context Compression, SWE-Pruner, and trajectory reduction~\citep{kang2025acon,verma2026activecompression,wang2026swepruner,xiao2026trajectoryreduction}. 
Other work emphasizes evolving contexts and summarization-based management, including COMPASS, context folding, and end-to-end summarization for multi-turn RL~\citep{wan2025compass,sun2025contextfolding,lu2025supo}, as well as efficient lifelong and self-evolving memory systems such as Dynamic Cheatsheet, SimpleMem, MemRL, MemSkill, and General Agentic Memory~\citep{suzgun2026dynamic,liu2026simplemem,zhang2026memrl,zhang2026memskill,yan2025gam}. 

\emph{What this line of work evaluates} is how different memory mechanisms improve long-horizon reasoning, retrieval, persistence, or adaptation.
\emph{Why this is still insufficient} is that memory is usually treated as a design improvement inside a particular agent, rather than as a controlled harness factor whose contribution can be measured under matched models, tasks, and budgets.
For our framing, memory is not merely a model property; it is a harness choice that can substantially alter performance even when the base model is fixed.

\paragraph{Routing, orchestration, planning, and budgeted control.}
Another major aspect of the harness concerns how computation is allocated and how decisions are organized across steps, tools, and sub-agents.
Planning-oriented approaches such as Plan-and-Act improve long-horizon behavior through explicit planning before execution~\citep{erdogan2025planandact}. 
Recent work extends this into broader orchestration and adaptive control, including reflective prompt evolution in GEPA~\citep{agrawal2025gepa}, generalized routing across models and agents~\citep{guo2025moma}, evolving contexts for self-improving agents~\citep{zhang2025ace}, orchestration frameworks such as Orchestral AI~\citep{roman2026orchestral}, context-aware multi-agent routing in CASTER~\citep{liu2026caster}, intention-based planning under costly tool budgets~\citep{liu2026intent}, and adaptive routing via active inference in ODAR~\citep{ma2026odar}. 

\emph{What this line of work evaluates} is how better control policies improve long-horizon behavior, budget allocation, or tool-use efficiency.
\emph{Why this is still insufficient} is that these methods usually optimize a specific system rather than support controlled comparison across different harnesses under a shared protocol.
Together, these works show that the harness is not a passive wrapper but an active policy layer that determines planning depth, routing decisions, budget allocation, and fallback behavior.
What remains underexplored is how much these factors contribute to measured reasoning performance when the underlying model is held fixed.

\paragraph{Skills, tool abstraction, and compositional reasoning.}
A related line of work studies \emph{skills} as higher-level abstractions that sit between a model and raw tool calls.
Examples include induced programmatic skills for agentic tasks~\citep{wang2025asi}, compositional skill synthesis for reasoning~\citep{jiao2026agenticproposing}, empirical analyses of deployed skill ecosystems~\citep{ling2026agentskills}, lifelong skill self-evolution~\citep{yang2026autoskill}, and token-efficient skill optimization~\citep{gao2026skillreducer}. 

\emph{What this line of work evaluates} is how structured action abstractions can make agent behavior more effective or efficient.
\emph{Why this is still insufficient} is that skills reshape the effective action space and control flow seen by the model, yet are rarely analyzed as part of a broader harness decomposition.
These abstractions therefore naturally belong to the harness layer rather than to the base model alone, and should be evaluated as such.

\paragraph{Self-improving and self-evolving agents.}
The same is true for systems that adapt at test time through experience, memory updates, tool-policy updates, or self-modification.
Recent work studies EvoAgent~\citep{feng2025evoagent}, self-evolving multi-agent simulation frameworks~\citep{almansoori2025medagentsim}, SafeSieve for experience-driven pruning in multi-agent communication~\citep{zhang2025safesieve}, EvoTool for self-evolving tool-use policies~\citep{yang2026evotool}, and Memento, which improves agents by tuning the surrounding system rather than the base model itself~\citep{zhou2025memento}. 
Test-time learning is also explored through adaptive memory and human guidance, as in Dynamic Cheatsheet and human-in-the-loop self-improvement methods~\citep{suzgun2026dynamic,he2025aria}, while Evo-Memory makes self-evolving memory an explicit evaluation target~\citep{wei2025evomemory}. 

\emph{What this line of work evaluates} is how agents can improve during deployment through adaptation of memory, policy, or surrounding scaffolds.
\emph{Why this is still insufficient} is that these gains are often reported as improvements in the overall agent, without disentangling whether the change comes from better reasoning, better adaptation mechanisms, or more favorable harness policies.
These works reinforce the same point: gains in agent performance often arise through harness adaptation rather than parametric model changes alone.

\paragraph{Harness engineering, runtime infrastructure, and observability.}
A recent and rapidly growing body of work studies the harness layer itself as a systems problem.
Some work focuses on observability, proposing system-level tracing and structured logs for agent execution, such as AgentSight and AgentTrace~\citep{zheng2025agentsight,alsayyad2026agenttrace}. 
Other work studies governed or typed execution, hierarchical retrieval interfaces, or infrastructure for complex codebase interaction, for example POLARIS, A-RAG, and Codified Context~\citep{moslemi2026polaris,du2026arag,vasilopoulos2026codified}. 
There is also growing interest in automatically constructing or optimizing the harness, including AutoHarness, domain-specific harnesses for terminal coding and compilers, and explicit optimization of model-harness combinations in Meta-Harness~\citep{lou2026autoharness,bui2026terminalagents,zheng2026compilerharness,lee2026metaharness}. 
Recent conceptual work makes this systems view explicit by treating harness engineering as a distinct layer of control, runtime, and agency for language agents~\citep{he2026harnessengineering}. 

\emph{What this line of work evaluates} is the harness as a systems artifact: something that can be instrumented, governed, synthesized, and optimized.
\emph{Why this is still insufficient} is that most of this work still does not provide a general evaluation substrate for matched comparison of heterogeneous model-harness systems on shared tasks with shared verifiers and budgets.
Harnesses that can be traced, governed, synthesized, and optimized as modular artifacts should be evaluated accordingly.

\paragraph{Security and safety.}
\looseness=-1
Because agent behavior is mediated by execution environments, tools, memory, and runtime policies, security and safety arise at the harness level rather than purely at the model level.
Recent work on OpenClaw emphasizes security controls in session-based tool execution; broader studies examine agent security under realistic threats and failure modes~\citep{buehler2025securingagentexecution}. 

\emph{What this line of work evaluates} is whether runtime policies, containment, and execution safeguards can make agents safer in practice.
\emph{Why this is still insufficient} is that safety-oriented harness choices can also affect measured task performance, making it difficult to interpret end-to-end scores unless policy enforcement and reasoning ability are considered together.
This perspective reinforces the need for evaluation settings that can separate raw reasoning ability from policy enforcement, containment, and execution safeguards.

\subsection{Reasoning and Interactive Evaluation Benchmarks}

Benchmarking work relevant to our setting falls into two broad categories: \textbf{model-centric reasoning benchmarks} and \textbf{interactive agent benchmarks}. 
This distinction mirrors the transition studied in our paper: from evaluating a standalone model under direct inference to evaluating a model embedded in a harness that can act, observe, and interact with an environment.

\paragraph{Model-centric reasoning benchmarks.}
A large body of work evaluates language models on mathematical, scientific, code, and formal reasoning under a prompt-in/answer-out protocol.
Representative examples include MATH~\citep{hendrycks2021math}, GSM8K~\citep{cobbe2021gsm8k}, GPQA~\citep{rein2024gpqa}, SciBench~\citep{wang2024scibench}, BBH~\citep{suzgun2022bbh}, HumanEval~\citep{chen2021codex}, and LiveCodeBench~\citep{jain2024livecodebench}. 
Broader model-evaluation efforts such as HELM also highlight the importance of standardized evaluation across scenarios and metrics~\citep{liang2022helm}. 
More recent frontier and verifier-backed benchmarks extend this paradigm while retaining relatively clean correctness criteria. 
Humanity's Last Exam is designed to remain difficult at the knowledge frontier while preserving closed-ended grading~\citep{phan2025hle}, and HLE-Verified was introduced to improve reliability and reduce ambiguity through structured verification~\citep{zhai2026hleverified}. 
FrontierMath similarly emphasizes advanced mathematical reasoning with automated verification to reduce ambiguity and contamination~\citep{glazer2024frontiermath}, while FormalMATH targets formal mathematical reasoning in theorem-proving-style settings~\citep{yu2025formalmath}. MMMU-Pro evaluates multi-discipline multimodal understanding and reasoning~\citep{yue2024mmmupro}.

\emph{What this line of work evaluates} is raw reasoning capability under controlled inference and relatively reliable judging.
\emph{Why this is still insufficient} is that most model-centric reasoning benchmarks assume a fixed inference protocol and evaluate the model as the primary object of interest.
They are therefore ideal for measuring \emph{raw model reasoning}, but do not by themselves distinguish model capability from harness capability once reasoning is mediated by tools, memory, execution support, and multi-step interaction.

\paragraph{Interactive agent benchmarks.}
A second line of work evaluates agents through action in realistic environments rather than through answer matching alone.
Early and influential examples include AgentBench~\citep{liu2024agentbench}, Mind2Web~\citep{deng2023mind2web}, WebArena~\citep{zhou2023webarena}, VisualWebArena~\citep{koh2024visualwebarena}, and WebVoyager~\citep{he2024webvoyager}, which study web interaction, browsing, and multimodal web-based tasks. 
Other benchmarks target broader forms of digital and workplace interaction, including OSWorld for open computer-use tasks~\citep{xie2024osworld}, WorkArena for common knowledge-work tasks~\citep{drouin2024workarena}, and GAIA for general AI assistants solving realistic multi-step tasks~\citep{mialon2023gaia}. 
A large coding- and software-oriented line includes SWE-bench for resolving real GitHub issues~\citep{jimenez2024swebench}, Terminal-Bench for shell-based workflows~\citep{merrill2026terminalbench}, ToolSandbox for stateful conversational tool use~\citep{lu2024toolsandbox}, R2E-Gym for procedural software-engineering environments with hybrid verifiers~\citep{jain2025r2egym}, ProdCodeBench for production-derived coding-agent evaluation~\citep{jha2026prodcodebench}, and FeatureBench for complex feature-level agentic coding tasks~\citep{zhou2026featurebench}. 
Additional benchmarks emphasize realistic assistant, research, or collaborative-agent settings, including AssistantBench~\citep{yoran2024assistantbench}, BrowseComp~\citep{wei2025browsecomp}, CORE-Bench for computational reproducibility~\citep{siegel2024corebench}, DeepResearchGym~\citep{coelho2025deepresearchgym}, ResearchGym~\citep{garikaparthi2026researchgym}, MLGym~\citep{nathani2025mlgym}, InnovatorBench for AI research agents~\citep{wu2026innovatorbench}, MultiAgentBench for collaboration and competition among multiple agents~\citep{zhu2025multiagentbench}, and Agentified Benchmarking for Logical Reasoning Agents~\citep{ni2026agentified}. 
Newer multi-turn and workflow-oriented settings such as $\tau$-bench, WebWalker, NESTFUL, and $\tau$-Knowledge place even greater emphasis on structured tool use, conversational state changes, and policy-constrained interaction~\citep{yao2024taubench,wu2025webwalker,basu2024nestful,shi2026tauknowledge}. 
Recent work also argues that sufficiently rich computer environments can elicit broader forms of agentic intelligence than static prompting alone~\citep{cheng2026sandbox}.

\emph{What this line of work evaluates} is end-to-end behavior of agents acting in realistic environments.
\emph{Why this is still insufficient} is that end-to-end performance depends not only on the base model, but also on the surrounding harness, including tool interfaces, control flow, observation formatting, memory and state handling, and runtime policies.
This makes interactive agent benchmarks highly valuable for measuring full-system behavior, while also motivating harness-aware evaluation frameworks that can better separate model effects from harness effects.

Taken together, these two benchmark families motivate the core distinction in our paper.
\emph{Model-centric reasoning benchmarks} evaluate a model in isolation under direct inference, whereas \emph{interactive agent benchmarks} evaluate a model-harness system acting in an environment.
Our empirical design focuses on largely verifiable reasoning settings because they provide a clean testbed for connecting these two traditions: they retain reliable correctness signals from model-centric evaluation while surfacing the system-level effects that arise once a model is embedded in a harness.

\subsection{Evaluation Protocols, Reliability, and Human--Agent Interaction}

\paragraph{What this line of work evaluates.}
A growing body of work argues that agent evaluation cannot be reduced to final accuracy alone.
Agent behavior is shaped by planning quality, tool choice, policy compliance, recoverability, robustness, efficiency, and interaction quality.
AgentBench evaluates LLMs as agents across diverse environments and highlights gaps in long-horizon reasoning and instruction following~\citep{liu2024agentbench}. 
T-Eval evaluates tool use \emph{step by step} rather than only through final outcomes~\citep{chen2024teval}. 
$\tau$-bench proposes reliability-oriented metrics over repeated trials in dynamic tool-agent-user interactions~\citep{yao2024taubench}. 
MCP-Bench emphasizes trajectory-level planning and cross-tool coordination through standardized tool servers~\citep{wang2025mcpbench}. 
OSWorld-Human complements success metrics with human reference trajectories and efficiency diagnostics~\citep{abhyankar2025osworldhuman}. 
AgentRewardBench further studies the reliability of automatic trajectory evaluation itself, focusing on how web-agent trajectories should be judged rather than only how agents should act~\citep{lu2025agentrewardbench}.

Recent work expands this picture in several directions.
Some papers focus on debugging and recoverability, including failure analysis for agent trajectories and explicit metrics such as ERR~\citep{zhu2025agentdebug,vuddanti2026err}. 
Others study robustness under more realistic interfaces, such as imperfect APIs, context retrieval in coding agents, and memory stress tests in long-horizon interactions~\citep{kim2026wildagteval,li2026contextbench,hu2026evaluatingmemoryagents,jiayang2026amemgym}. 
A broader reliability perspective is developed in recent work calling for a more systematic science of AI agent reliability, alongside practical regression-testing and readiness frameworks such as AgentAssay and LLM Readiness Harness~\citep{rabanser2026reliability,bhardwaj2026agentassay,maiorano2026readiness}. 
Related work also highlights evaluation variance and protocol sensitivity, for example through analyses of randomness in agentic evaluation~\citep{bjarnason2026randomness} and through more rigorous evaluation protocols for coding agents on SWE-bench~\citep{yu2025utboost}. 
Related work also studies when human collaboration is preferable to full automation and how human feedback can improve agent behavior at test time~\citep{natarajan2024humanintheloop,he2025aria}.

\paragraph{Why this is still insufficient for harness-aware evaluation.}
Collectively, these protocols motivate trajectory-aware logging, reproducibility mechanisms, modular judging, and reliability analysis beyond a single scalar success metric.
However, most of this work aims to improve how end-to-end agents are evaluated, rather than to isolate what portion of measured performance should be attributed to the harness itself.
In other words, it strengthens the measurement of agent behavior, but does not by itself resolve the attribution problem created when the evaluated object is a model embedded in a complex runtime.

\subsection{From Model-Only to Model--Harness Evaluation}

\paragraph{Evaluation requirements.}
Existing frameworks support agent evaluation through task, execution, and scoring interfaces~\citep{kapoor2025hal,harbor}. Comparing model--harness systems requires explicit policies for session reuse, memory retention, and resource limits, together with links between task outcomes and execution records.

\paragraph{Position of AlphaDiana.}
AlphaDiana organizes these controls around comparisons of complete agents and interventions on tool, skill, and memory configurations. It preserves harness-native execution, controls file reuse and memory retention separately, and links task outcomes to traces and resource measurements. These interfaces support comparisons such as Qwen3.5-27B with OpenClaw on HLE-Verifiable under explicit task, scoring, and resource policies.